\documentclass{article} 
\usepackage{iclr2026_conference,times}

\usepackage{amsmath,amsfonts,bm}

\def\eqref#1{equation~\ref{#1}}

\def\1{\bm{1}}

\DeclareMathAlphabet{\mathsfit}{\encodingdefault}{\sfdefault}{m}{sl}
\SetMathAlphabet{\mathsfit}{bold}{\encodingdefault}{\sfdefault}{bx}{n}

\usepackage{url}
\usepackage{xurl}
\usepackage{hyperref}
\hypersetup{
    colorlinks=true,
    linkcolor=BrickRed,
    citecolor=NavyBlue,
    urlcolor=NavyBlue
}

\iclrfinalcopy 

\usepackage[utf8]{inputenc} 
\usepackage[T1]{fontenc}    
\usepackage{url}            
\usepackage{booktabs}       
\usepackage{amsfonts}       
\usepackage{nicefrac}       
\usepackage{microtype}      
\usepackage[dvipsnames]{xcolor}

\usepackage{amsmath}
\usepackage{graphicx}
\usepackage{tabularx,colortbl}
\usepackage{multirow, makecell}
\usepackage{pifont}
\usepackage{tabularx}
\usepackage{wrapfig}
\usepackage{tikz}
\usepackage{subcaption}
\usepackage{fnpct}

\usepackage{algorithm}
\usepackage{algorithmicx}
\usepackage{algpseudocode}

\newcommand{\eg}{\textit{e.g.}~}
\newcommand{\ie}{\textit{i.e.}~}

\newcommand{\circledgray}[1]{%
  \tikz[baseline=(char.base)]{%
    \node[shape=circle,fill=gray!40,inner sep=0.5pt,minimum size=6pt] (char) {\textcolor{black}{#1}};
  }%
}

\newcommand{\cmark}{\color{ForestGreen}\ding{51}}%
\newcommand{\xmark}{\color{BrickRed}\ding{55}}%

\definecolor{lightgreen}{RGB}{240, 255, 230}
\definecolor{middlegreen}{RGB}{210, 245, 200}
\definecolor{darkgreen}{RGB}{175, 220, 155}
\newcommand{\lightgreen}{\cellcolor{lightgreen}}
\newcommand{\middlegreen}{\cellcolor{middlegreen}} 
\newcommand{\darkgreen}{\cellcolor{darkgreen}} 
\newcommand{\lightsquare}{\colorbox{lightgreen}{\rule{0pt}{0.45em}\rule{0.9em}{0pt}}~}
\newcommand{\middlesquare}{\colorbox{middlegreen}{\rule{0pt}{0.45em}\rule{0.9em}{0pt}}~}
\newcommand{\darksquare}{\colorbox{darkgreen}{\rule{0pt}{0.45em}\rule{0.9em}{0pt}}~}

\definecolor{lightblue}{RGB}{210,224,255}
\definecolor{middleblue}{RGB}{158,177,220}
\definecolor{darkblue}{RGB}{98,121,175}

\definecolor{lightred}{RGB}{255, 224, 224}

\title{\textsc{SelvaBox}: A high‑resolution dataset for tropical tree crown detection}

\author{
  Hugo Baudchon$^{1,2,}$\textsuperscript{\textdagger},
  Arthur Ouaknine$^{1,3,4}$,
  Martin Weiss$^{1,2}$,
  Mélisande Teng$^{1,2}$,\\
  \textbf{Thomas R. Walla$^{5}$,
  Antoine Caron-Guay$^{2}$,
  Christopher Pal$^{1,6}$,
  Etienne Laliberté$^{2,1,4}$}\\
  $^1$Mila -- Quebec AI Institute \hspace{1em}
  $^2$Université de Montréal \hspace{1em}
  $^3$McGill University\\
  $^4$Rubisco AI \hspace{1em}
  $^5$Colorado Mesa University \hspace{1em}
  $^6$Polytechnique Montreal\\
  \textsuperscript{\textdagger}\texttt{hugo.baudchon@umontreal.ca}
}

\begin{document}

\maketitle

\begin{abstract}
  Detecting individual tree crowns in tropical forests is essential to study these complex and crucial ecosystems impacted by human interventions and climate change. 
  However, tropical crowns vary widely in size, structure, and pattern and are largely overlapping and intertwined, requiring advanced remote sensing methods applied to high-resolution imagery. 
  Despite growing interest in tropical tree crown detection, annotated datasets remain scarce, hindering robust model development.
  We introduce \textsc{SelvaBox}, the largest open‑access dataset for tropical tree crown detection in high-resolution drone imagery. It spans three countries and contains more than $83\,000$ manually labeled crowns -- an order of magnitude larger than all previous tropical forest datasets combined. 
  Extensive benchmarks on \textsc{SelvaBox} reveal two key findings: \circledgray{1} higher-resolution inputs consistently boost detection accuracy; and \circledgray{2} models trained exclusively on \textsc{SelvaBox} achieve competitive zero-shot detection performance on unseen tropical tree crown datasets, matching or exceeding competing methods. Furthermore, jointly training on \textsc{SelvaBox} and three other datasets at resolutions from 3 to 10 cm per pixel within a unified multi-resolution pipeline yields a detector ranking first or second across all evaluated datasets. Our dataset\footnote{\textsc{SelvaBox} dataset: \url{https://huggingface.co/datasets/CanopyRS/SelvaBox}}, code\footnote{Preprocessing library (\textit{geodataset}): \url{https://github.com/hugobaudchon/geodataset}}\footnote{Inference, training \& benchmark (\textit{CanopyRS}): \url{https://github.com/hugobaudchon/CanopyRS}}, and pre-trained weights are made public.
\end{abstract}

\begin{figure}[h!]
    \centering
    \includegraphics[scale=0.1]{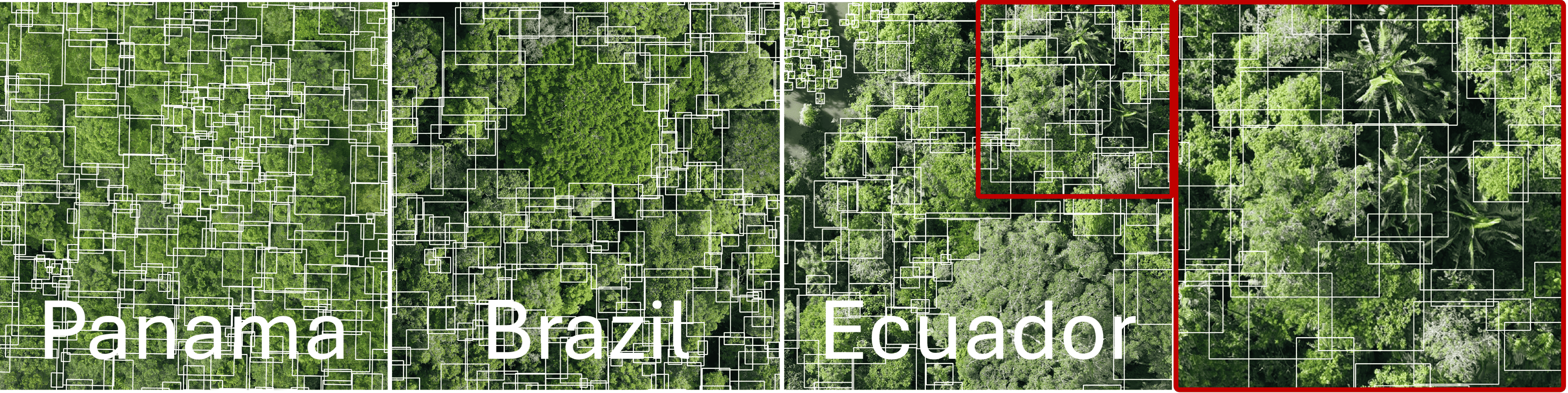}
    \caption{\textbf{The \textsc{SelvaBox} dataset.} The illustrated samples are extracted from rasters recorded in Panama, Brazil and Ecuador with a spatial extent of $80\text{m}\times80\text{m}$ and a resolution of 1.2 to 5.1 cm per pixel. The red square on the right highlights a zoom of the Ecuador sample with a spatial extent of $40\text{m}\times40\text{m}$ at the same resolution.}
    \label{fig:teaser}
\end{figure}

\section{Introduction}

Tropical forests cover 10\% of the land area, but they store most of the biomass and biodiversity of plants on our planet \citep{pan_large_2011, gattiNumberTreeSpecies2022}. Large trees that reach the upper canopy have a disproportionate influence on the functioning of tropical forests, with  the largest 1\% of trees storing half of the carbon of forests worldwide \citep{lutzGlobalImportanceLargediameter2018a}. However, tree demography patterns in tropical forests are being altered, with increasing tree mortality, due to climate change \citep{brienenLongtermDeclineAmazon2015, bonan_forests_2008, esquivel-muelbertCompositionalResponseAmazon2019} and human interventions \citep{harris_global_2021}. As such, monitoring individual trees in tropical forests is essential to understand the current and future potential of these forests to regulate the global climate \citep{daviesForestGEOUnderstandingForest2021}.

Monitoring tropical trees is a difficult task involving slow, costly, and dangerous ground surveys by forest technicians \citep{delimaMakingForestData2022}. 
Forest plots of tens of hectares are the gold standard of tropical tree monitoring to measure and map each individual, but completing a single one can take years of dedicated work by large teams of experts \citep{daviesForestGEOUnderstandingForest2021}.
Remote sensing technologies considerably augment field work, facilitating forest cartography through aerial detection of individual trees across spatial extents vastly exceeding the practical limitations of ground-based inventories \citep{brandtUnexpectedlyLargeCount2020}.
Satellite imagery has been used successfully in forest monitoring tasks \citep{ouaknine_openforest_2025}, including height map estimation \citep{tolan_very_2024, lang_high-resolution_2023} and individual tree crown detection \citep{brandtUnexpectedlyLargeCount2020, tucker_sub-continental-scale_2023, ZHENG2020154, ZHENG2023113485, 10750500}. However, the highest resolution satellite imagery is typically 0.3-0.5 m, which is still too coarse to distinguish trees in dense tropical forest canopies. Moreover, cloudy conditions complicate satellite sensing in the tropics.

By contrast, unoccupied aerial vehicles (UAVs) or drones can achieve cm-resolution ($<5$\,cm), albeit at the expense of spatial coverage \citep{reiersen_reforestree_2022, vasquez_barro_2023, cloutier_influence_2024}.
Recently, datasets of UAV LiDAR \citep{puliti_for-instance_2023, puliti_benchmarking_2025, gaydon_pureforest_2025} and methods for forest structure assessment with LiDAR data have been extensively developed \citep{bai_semantic_2023, ma_forest-pointnet_2023, vermeer_lidar-based_2024, henrich_treelearn_2024}.
However, the cost  of LiDAR sensing limits its adoption in tropical contexts where researchers are financially disadvantaged \citep{de2022making} and calls for the development of RGB-only methods. Most open access high-quality, high-resolution tree detection RGB datasets represent temperate forests of the global North (Tab.~\ref{table:related_datasets}). Tropical forests remain severely underrepresented and have relatively modest annotation counts \citep{ball_accurate_2023, vasquez_barro_2023} despite their critical significance for biodiversity and carbon storage. 

The high tree species diversity \citep{gattiNumberTreeSpecies2022} and heterogeneity in crown sizes, shapes and textures (Fig.~\ref{fig:teaser} and~\ref{fig:boxes_ditrib}) in tropical forests pose unique challenges.
Indeed, solving the problem of detecting numerous objects of highly variable sizes within the same scene is still an open topic in computer vision applied to remote sensing \citep{rabbi_small-object_2020, li_cross-layer_2021, bashir_small_2021}. 
While convolutional neural networks (CNNs) remain the predominant approach for individual tree crown detection \citep{weinstein_individual_2019, zamboni_benchmarking_2021, Onishi2021, yu_comparison_2022, ball_accurate_2023, zhao_systematic_2023, bountos_fomo_2025, HAJJAJI2025102952}, recent studies have explored transformer-based architectures on satellite imagery \citep{JIANG2025103114}, motivated by their effectiveness in multi-scale object recognition tasks (e.g., \cite{liu_swin_2021, zhang_dino_2023}). However, a comprehensive, resolution-aware benchmark comparing these two paradigms on UAV imagery across diverse forest ecosystems and out-of-distribution scenarios remains absent. With the growing number of UAV datasets acquired with different flight parameters, models that can generalize across resolutions and standardized frameworks are needed to bridge the gap between the ecology and computer vision communities.

We address these challenges through our contributions:
\circledgray{1} \textsc{SelvaBox}, a high-resolution drone imagery dataset spanning three neotropical countries (Brazil, Ecuador, Panama) and comprising over $83\,000$ manual bounding box annotations on individual tree crowns;
\circledgray{2} An exhaustive benchmark of detection methods at varying resolutions and input sizes, including a standardized evaluation framework for UAV rasters and a comprehensive assessment of model generalization on out-of-distribution (OOD) samples; 
\circledgray{3} State-of-the-art models trained for tree crown detection out-performing competing methods on both topical and non-tropical forest datasets, in both in-distribution (ID) and OOD settings;
and \circledgray{4} two open-source Python libraries facilitating raster preprocessing, inference, postprocessing and standardized benchmarking.
These contributions aim to simultaneously advance tropical forest monitoring and applications of machine learning to critical environmental challenges.


\section{Related work}
\label{sec:related_work}

\paragraph{Datasets.}

High-resolution drone imagery enables detailed tree characterization at the pixel level (see Figure~\ref{fig:teaser}). 
This capability has catalyzed the development of open access forest monitoring datasets \citep{ouaknine_openforest_2025} specifically designed for tree crown semantic segmentation tasks, including pixel-wise canopy mapping \citep{galuszynski_automated_2022}, woody invasive species identification \citep{kattenborn_uav_2019}, and tree species classification \citep{cloutier_influence_2024, kattenborn_convolutional_2020}.
\begin{wraptable}[17]{r}{0.6\linewidth}
\vspace{-5pt}
\caption{\textbf{Related datasets.} The number of tree crowns manually$^*$ annotated (`\# Trees') are noted in `k' for thousands. The reported resolution or ground sampling distance (`GSD') is in centimeter per pixel. We define the forest `type' as either urban, plantation, natural; `biome' as either temperate, tropical or worldwide (when the dataset spans over several biomes). $^*$except for ReforesTree, see Section~\ref{sec:benchmarking_methods}.}
\centering
\resizebox{\linewidth}{!}{%
\begin{tabular}{c r r c c}
\toprule
Name & \# Trees & GSD & Type & Biome \\
\midrule
NeonTreeEval. \citep{weinstein_benchmark_2021} & 16k & 10 & natural & temperate  \\
ReforesTree \citep{reiersen_reforestree_2022} & 4.6k & 2 & plantation & tropical  \\
Firoze \textit{et al.} \citep{firoze_tree_2023} & 6.5k & 2--5 & natural & temperate  \\
Detectree2 \citep{ball_accurate_2023} & 3.8k & 10 & natural & tropical \\
BCI50ha \citep{vasquez_barro_2023} & 4.7k & 4.5 & natural & tropical \\
BAMFORESTS \citep{troles_bamforests_2024} & 27k & 1.6--1.8 & natural & temperate \\
QuebecTrees \citep{cloutier_influence_2024} & 23k & 1.9 & natural & temperate  \\
Quebec Plantation \citep{lefebvre_uav_2024} & 19.6k & 0.5 & plantation & temperate  \\
OAM-TCD \citep{veitch-michaelis_oam-tcd_2024} & 280k & 10 & mostly urban & worldwide  \\
\midrule
\textsc{SelvaBox} \textbf{(ours)} & \textbf{83k} & \textbf{1.2--5.1} & \textbf{natural} & \textbf{tropical}  \\
\bottomrule
\end{tabular}
}
\label{table:related_datasets}
\end{wraptable}
Tree crown semantic segmentation, a pixel-wise classification task, cannot inherently distinguish individual trees, making it unsuitable for applications such as tree counting or biomass estimation where individual tree crown detection and delineation are essential \citep{fu_automatic_2024}.
Datasets for individual tree crown detection \citep{weinstein_benchmark_2021, reiersen_reforestree_2022} and delineation \citep{ball_accurate_2023, firoze_tree_2023, vasquez_barro_2023, cloutier_influence_2024, lefebvre_uav_2024, veitch-michaelis_oam-tcd_2024}, corresponding to object detection and instance segmentation tasks respectively, have been proposed for both general forest monitoring and specialized applications such as dead tree identification \citep{mosig_deadtreesearth_2024}. 
Table~\ref{table:related_datasets} summarizes open access datasets for general tree crown monitoring.
Despite considerable community efforts to share manually annotated tree crown data, a substantial gap remains in datasets for monitoring tropical trees in natural forests.
\paragraph{Modeling.} 
Deep learning is the dominant paradigm for individual tree crown delineation, superseding earlier computer vision and machine learning methods \citep{kattenborn_review_2021}.
Open access datasets (Tab.~\ref{table:related_datasets}) have facilitated the development of individual tree crown detection models with deep learning architectures, including Faster R-CNN \citep{ren_faster_2015}, Mask R-CNN \citep{he_mask_2017}, and RetinaNet \citep{lin_focal_2017}, as demonstrated with DeepForest \citep{weinstein_individual_2019} and Detectree2 \citep{ball_accurate_2023}.
These CNN-based methods have proven effective in diverse forest scenarios \citep{zhao_systematic_2023}.
%
Tree crown models have also leveraged SAM \citep{kirillov_segment_2023} by providing efficient prompts for zero-shot tree crown delineation \citep{teng_assessing_2025}.
While the FoMo benchmark \citep{bountos_fomo_2025} has explored transformer-based architectures including pretrained DeiT \citep{touvron_training_2021} and DINOv2 \citep{oquab_dinov2_2024} backbones, advanced transformer-based object detection methods \citep{liu_dab-detr_2022, zhang_dino_2023} remain underexplored in this domain. 

\paragraph{Evaluation.}
Previous open access datasets (Tab.~\ref{table:related_datasets}) have evaluated detection methods using classification-based metrics per tree (recall, precision, F1-score) \citep{weinstein_deepforest_2020, weinstein_benchmark_2021, zheng_growing_2021, beloiu_individual_2023}
with detection-based metrics such as intersection over union (IoU) 
and mean average precision (mAP) \citep{hao_automated_2021, yu_comparison_2022, ball_accurate_2023, fu_automatic_2024, firoze_tree_2023, veitch-michaelis_oam-tcd_2024, bountos_fomo_2025}. 
UAV rasters are usually divided in tiles for training and evaluation, but tile-level metrics are susceptible to edge effects (where partial trees appear at tile boundaries) and duplicate detections when scaled to larger areas, complicating accurate tree counting. 
As a consequence, tile-level metrics fail to accurately represent performances at the entire raster level,
 which is what matters to practioners. 
%
For example, tracking the mortality of large tropical trees over time and across vast areas requires aggregating detections from individual tiles into a coherent raster-level map.
A recall metric for keypoint-in-tree prediction tasks at the raster level was proposed to evaluate OAM-TCD \citep{veitch-michaelis_oam-tcd_2024}. In this work, we extend the evaluation of aggregated predictions from individual images to detection-based tasks, including both precision and recall metrics (F1-score) as well as the location of each object.

\paragraph{Multi-resolution.}
Despite growing interest in multi-scale and multi-resolution analysis for deep learning in remote sensing applications \citep{reed_scale-mae_2023, bountos_fomo_2025}, these approaches remain understudied for forest monitoring. 
%
While increased spatial extent per tile improves tree crown classification \citep{nasi_using_2015, liu_impact_2020, kattenborn_convolutional_2020} and higher tile resolution benefits tree crown semantic segmentation more than increased spatial extent \citep{schiefer_mapping_2020}, resolution-induced domain shift remains challenging for individual tree crown detection. Current pre-trained models (e.g., DeepForest, Detectree2) show poor zero-shot performance on OOD samples \citep{gan_tree_2023}, though targeted fine-tuning can mitigate this gap \citep{bountos_fomo_2025}. 
%
Further research is needed to evaluate how tile spatial extent, size, and resolution impact detection performance and to develop fine-tuning methodologies that reduce zero-shot degradation on OOD samples, particularly given the substantial size variation in tropical tree crowns (Fig.~\ref{fig:boxes_ditrib}).


\section{The \textsc{SelvaBox} dataset\label{sec:dataset_description}}

%
\begin{wrapfigure}[14]{r}{0.55\textwidth}
\centering
\vspace{-1.2cm}
\includegraphics[width=0.55\textwidth]{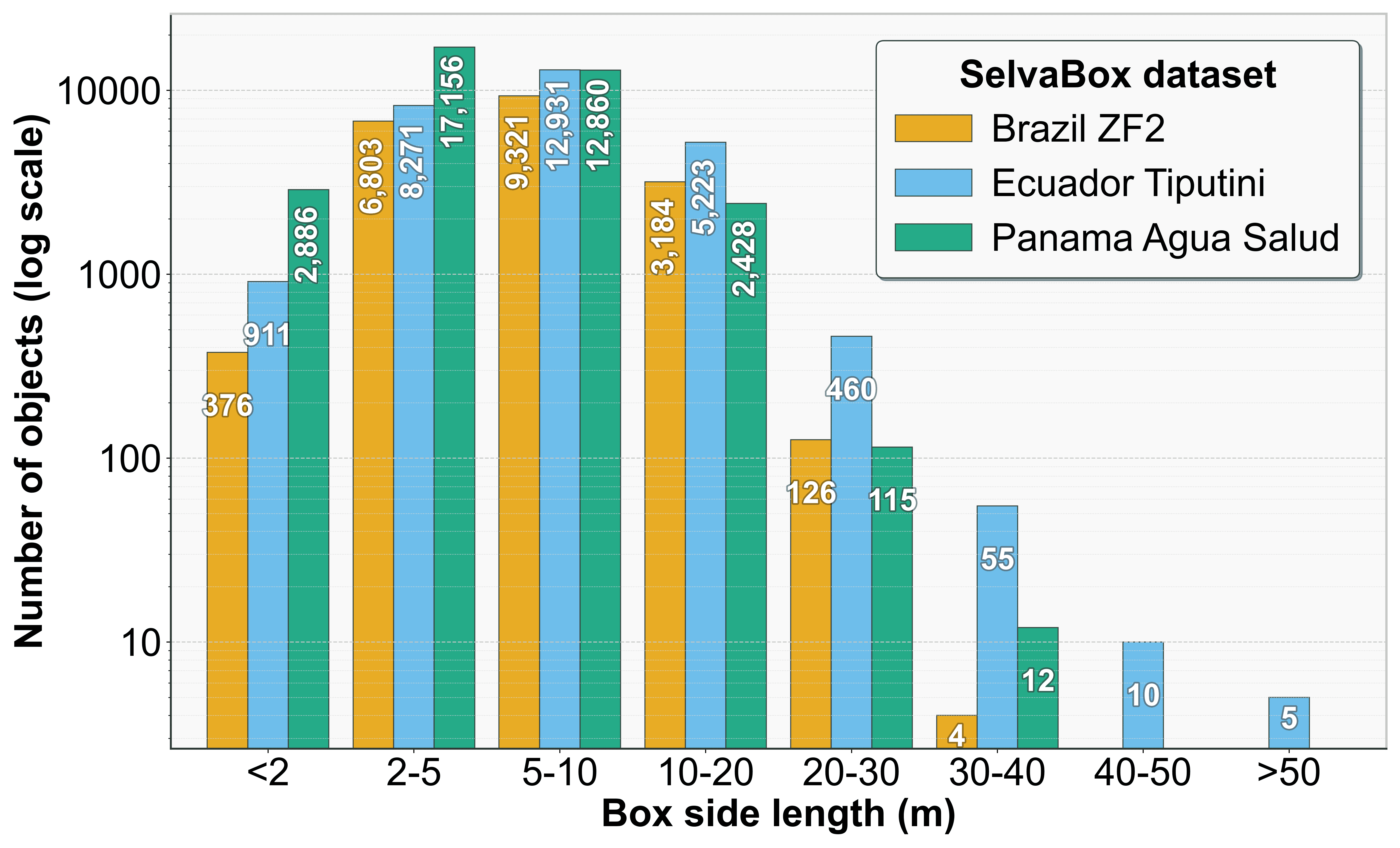}
\caption{\textbf{Distribution of box annotations size in \textsc{SelvaBox} per country.}}
\label{fig:boxes_ditrib}
\end{wrapfigure}
%

We present \textsc{SelvaBox}, a large-scale benchmark dataset addressing the critical open-access annotation scarcity in tropical forest remote sensing (Sec.~\ref{sec:related_work}) while motivating research in individual tree crown detection.
\textsc{SelvaBox} encompasses $83\,137$ individual tree crown bounding boxes on top of $14$ RGB orthomosaics, including $96.6$~ha in Brazil, $96$~ha in Panama and $318.1$~ha in Ecuador, recorded with four different drones (DJI Mavic 3 Entreprise [m3e], DJI Mavic 3 Multispectral [m3m], DJI Mavic Pro [mavicpro], DJI Mavic Mini 2 [mini2]) at ground sampling distance (GSD) between 1.2--5.1 cm per pixel (Tab.~\ref{table:rasters_info} in App.~\ref{sec:appendix_orthomosaics}).
Our drone imagery was acquired over primary and secondary forests, and native tree plantations. It includes diverse sets and shapes of tropical trees as depicted in Figure~\ref{fig:teaser}. More details about the orthomosaics can be found in App.~\ref{sec:appendix_orthomosaics}.


\paragraph{Locations.}
The RGB imagery was acquired in three countries: Brazil, Ecuador, and Panama (Tab.~\ref{table:rasters_info} in App.~\ref{sec:appendix_orthomosaics}). The Brazil data was collected at the ZF-2 station, a forest with high-diversity characteristic of the Central Amazon and growing on nutrient-poor soils. The topography consists of plateaus dissected by valleys \citep{amaralDynamicsTropicalForest2019}. The Ecuador data was recorded at the Tiputini Biodiversity Station (TBS), located within the Yasuní Biosphere Reserve, one of the most biodiverse forests on Earth \citep{valenciaTreeSpeciesDistributions2004}. The climate of this Western Amazonia region is considered to be aseasonal compared to Central Amazonia while the soils tend to be richer in nutrients as they are derived from younger sediments from the Andes \citep{hoornAmazoniaTimeAndean2010}. Finally, the Panama data was acquired from four areas of the  Agua Salud Project \citep{mayoralSurvivalGrowthFive2017}. Two areas are plantations of native tree species \citep{mayoralSurvivalGrowthFive2017}, while the other two are from surrounding secondary forests. The soils of Agua Salud are acidic and nutrient-poor \citep{vanbreugelSoilNutrientsDispersal2019}.
The tree species diversity of Central Panama is considered lower than our other two Amazonian sites.

\paragraph{Annotations.}
The data was manually annotated by five trained biologists. They were asked to draw bounding boxes around every individual tree crown they could reliably detect from the imagery. 
They generated $83\,137$ manual tree annotations during $1\,284$ people-hours with crowns spanning from $<2$ m to $>50$ m in diameter (Fig.~\ref{fig:boxes_ditrib}).
All annotations were produced with ArcGIS Pro version 3.0, stored in hosted feature layers on ArcGIS Online, have georeferenced coordinates, and were exported to geopackages. 
Figure \ref{fig:boxes_ditrib} shows the tree crown annotation bounding-box size distribution, where we notice a long-tail distribution for larger trees, especially in Ecuador. 

Our annotation process used photo-interpretation, the most reliable and feasible approach at this scale.
Field-based validation faces significant technical constraints in tropical forests, including GNSS signal blockage, multipath errors from dense canopy, difficulties linking non-straight tree trunks to canopy imagery, and variable geolocation errors—making the process less efficient and accurate than photo-interpretation \citep{Laliberte2025.09.02.673753}.
Additionally, logistical challenges like intense heat, humidity, heavy rain, and dense vegetation make fieldwork costly and hazardous.
Since LiDAR data requires expensive equipment and specialized annotators compared to photo-interpretation, it is less accessible for tropical forest scientists, who have limited research funding \citep{delimaMakingForestData2022}.
Consequently, we adopted an RGB-only validation approach to ensure scalability and broad applicability.
The annotation process followed a standardized protocol with initial training of domain-expert annotators, multi-pass annotation reviews, and systematic quality control (see App.~\ref{appendix:annotation_protocol} for full details). We supplemented visual interpretation with digital surface models (DSMs) derived from 3D photogrammetric point clouds, using elevation data to distinguish adjacent crowns with similar visual features but different heights.

\paragraph{Spatially separated splits.}
We propose train, validation and test splits, created spatially in the rasters to ensure no pixel overlap between splits and avoid geospatial auto-correlation \citep{KATTENBORN2022100018}, and including 61.4k, 9.6k, and 10.6k boxes respectively. We define our splits by manually creating areas of interest (AOIs) geopackages in the QGIS software (Fig.~\ref{fig:spatial_splits} in App.~\ref{sec:appendix_spatial_splits}). 
Orthomosaic borders with poor visual quality were deliberately excluded during AOI creation to ensure clean, artifact-free splits.
For the test split, we defined the AOIs on rasters with minimal visual reconstruction artifacts while including a maximal diversity and quality in box annotations.

\paragraph{Incomplete annotations.}
Although considerable effort was put into producing a dense tree‑crown mapping during the annotation process, some annotators reported difficulties clearly distinguishing a subset of individual trees on one raster in Brazil and three rasters in Ecuador, resulting in sparser annotations.
Annotation sparsity is a common challenge in tree detection datasets: The Detectree2 dataset contains only tiles covered in area by at least 40\% tree crown annotation polygons \citep{ball_accurate_2023}.
This method introduces noise during the training process as annotations may be missing for up to half of the trees in an image, causing misleading penalization. 
We adopt a different strategy where we mask targeted pixels in our AOIs with missing annotations when dividing the rasters in tiles. 
During training, we expect models to become agnostic to such masked pixels, \ie not predicting boxes in those areas, thus not being penalized due to missing annotations.
Such holes were created for train AOIs,
a sub-set of valid AOIs, while test AOIs were chosen to cover areas where annotations are dense and complete. Figure~\ref{fig:incomplete_annotations} (in App.~\ref{sec:appendix_incomplete_annotations}) shows an example of pixels masked that way.

\paragraph{Tiling and preprocessing.}
When tiling the rasters, \ie dividing rasters into tiles, we use AOI geopackages to mask pixels that are outside of each tile's assigned split. 
Each tree crown annotation is assigned to a single split where it overlaps the most according to the AOIs. For each tile, we keep annotations that overlap at least at 40\% with the tile's extent.
For the ready-to-train dataset, we remove tiles that contain no annotations, more than 80\% black (masked), white or transparent pixels.
A sliding-window tiling approach was used, with 50\% tile overlap for the training and validation splits, and 75\% for the test split to ensure that the largest trees entirely fit in at least one tile (Sec.~\ref{sec:benchmarking_methods}). We release our preprocessing pipeline as a python library called \textit{geodataset}.
The final preprocessed dataset is available on HuggingFace 
under the permissive CC-BY-4.0 license.

\section{Benchmarking models and methods\label{sec:benchmarking_methods}}
We structure our experiments in three sequential phases. First, we identify effective modeling choices by evaluating various object detection models and input image settings on \textsc{SelvaBox}, examining how resolution and spatial extent influence detection accuracy based on in-distribution performance (Sec.~\ref{subsec:model-selection}). Second, we validate the efficacy of multi-resolution domain augmentation by testing whether multi-resolution training improves or degrades performance compared to single-resolution training (Sec.~\ref{subsec:model-selection}). Finally, we assess generalization to other datasets by evaluating three categories: models trained exclusively on \textsc{SelvaBox}, models trained on \textsc{SelvaBox} combined with additional datasets, and models trained without \textsc{SelvaBox} including external methods (Sec.~\ref{subec:selvabox-ood-methodology}).

In addition to \textsc{SelvaBox}, we use the OAM-TCD \citep{veitch-michaelis_oam-tcd_2024}, NeonTreeEvaluation \citep{ben_weinstein_2022_5914554, weinstein_benchmark_2021}, QuebecTrees \citep{cloutier_2023_8148479, cloutier_influence_2024}, BCI50ha \citep{vasquez_barro_2023}, and Detectree2 \citep{ball_2023_8136161} datasets. We excluded the Quebec Plantations dataset \citep{lefebvre_uav_2024}, as it comprises non-tropical, young tree plantations outside the scope of our study. Similarly, we excluded ReforesTree \citep{reiersen_reforestree_2022}, a tropical plantation dataset whose bounding box annotations were generated by inference from a fine-tuned DeepForest model \citep{weinstein_deepforest_2020}, resulting in noisy annotations unsuitable for robust training or evaluation (Fig.~\ref{fig:refores_tree_infer} in App.~\ref{sec:appendix_reforestree_quali}). Additionally, we omitted the dataset published by Firoze \textit{et al.} \citep{firoze_tree_2023}, as it was designed for image sequence-based tree detection, with annotations derived from highly overlapping, video-like image sequences, introducing redundancy and requiring extensive preprocessing. Given that each dataset varies in ground sampling distance (GSD), tree crown size distribution, annotation type, and predefined splits or areas of interest (AOIs), we applied independent preprocessing procedures detailed in Appendix~\ref{sec:appendix_external_dataset_preprocessing}. Our benchmarking, inference, and training pipelines are publicly available in our Python repository \textit{CanopyRS}.

\paragraph{Evaluation metrics.}
To evaluate models at the tile level, we consider the industry‑standard COCO‑style $\text{mAP}_{50:95}$ and $\text{mAR}_{50:95}$ metrics \citep{Lin_2014_coco}. 
Due to the high number of objects per tile in \textsc{SelvaBox} (at 80m ground extent, see Sec.~\ref{subsec:model-selection}), QuebecTrees and BCI50ha, we increase the maxDets parameter of COCOEval from 100 to 400 for those datasets.

As detailed in Section~\ref{sec:related_work}, tile-level metrics do not necessarily reflect raster-level performance, which is the operational target for concrete applications such as large-scale forest inventories.
To address this, we propose RF1$_{75}$, a Raster-level F1 score evaluating final predictions after tile aggregation via Non-Maximum Suppression (NMS).
It uses the same greedy matching as COCO metrics, but requires a single, strict IoU threshold of $\geq75\%$ for a match. This $75\%$ threshold is a balanced criterion for dense canopies, where $50\%$ IoU is too permissive and $90\%$ is overly difficult. 
By integrating the F1 score at the raster level with this IoU restriction, the RF1$_{75}$ metric accounts for precision and recall, both important in forest monitoring applications.
For each dataset, we tune NMS hyperparameters on the validation set, apply the optimal settings to the test set, and report the final RF1$_{75}$ as a weighted average over all rasters (details in App.~\ref{sec:appendix_nms_hyperparameters}). While annotation noise makes a perfect score of 1.0 unlikely, maximizing RF1$_{75}$ is a practical target for reliable ecological monitoring.

\paragraph{Model architectures and training.} 

We compare four object detection approaches for tree crown delineation:  
\circledgray{1} Faster R-CNN with ResNet-50 backbone \citep{ren_2025_faster, he2016residual}, a widely used CNN-based detector;  
\circledgray{2} DeepForest \citep{weinstein_individual_2019, weinstein_deepforest_2020}, a RetinaNet variant trained on NeonTreeEvaluation;  
\circledgray{3} Detectree2 \citep{ball_accurate_2023}, a Mask R-CNN trained on a dataset also called Detectree2, evaluated in two variants: `resize' (multi-resolution tropical) and `flexi' (joint tropical-urban training); and  
\circledgray{4} DINO \citep{zhang_dino_2023}, a DETR-based transformer model that we evaluate with both ResNet-50 and Swin-L backbones \citep{liu_swin_2021}. Note that DINO (the DETR-based detector) and DINO (the self-supervised embedding model) are unrelated despite sharing the same name. While recent DETR-based architectures have reached similar or better performances \citep{10376521}, we chose DINO for 
its adoption by the community through Detectron2 \citep{wu2019detectron2} and Detrex \citep{ren2023detrex}.
DINO, Faster R-CNN, DeepForest, and Detectree2 serve as strong and diverse baselines from both general-purpose and domain-specific tree crown detection literature (Sec.~\ref{sec:related_work}).  All models are initialized from COCO-pretrained checkpoints. 
We implemented our own augmentation pipeline, and use standard crop, resize, flip, rotation and color augmentations (App.~\ref{sec:appendix_basic_augmentations}). Training sessions took between 12 hours and 3 days for both architectures. All hyperparameters used for training and testing are detailed in Appendix~\ref{sec:appendix_training_hyperparameters}.

\subsection{Model, resolution and spatial extent selection on \textsc{SelvaBox}}
\label{subsec:model-selection}

We choose a raster tiling scheme that balances detection accuracy, object coverage, and hardware constraints. Our standard tile is $80 \times 80$\,m at 4.5\,cm/px ($1777 \times 1777$ pixels).
This setting ensures that the largest crowns in \textsc{SelvaBox}, some upwards of 50\,m in diameter (Fig.~\ref{fig:boxes_ditrib}), fit entirely within one tile when using a 75\% overlap between tiles in our test set, while keeping our models (\eg, DINO 5-scale with Swin-L) trainable on 48\,GB GPUs with a batch size of one per GPU.

To assess the trade-offs between spatial resolution and ground extent, we conduct an ablation study across three configurations (Sec.~\ref{sec:experiments} and Tab.~\ref{table:benchmark_80m}). 
We vary the resolution between 4.5, 6, and 10\,cm/px, yielding $1777 \times 1777$, $1333 \times 1333$, and $800 \times 800$ pixel inputs for a fixed $80 \times 80$\,m ground extent.
In parallel, we test $40 \times 40$\,m tiles, which contain fewer crowns per image and still guarantee that over 99.9\% of crowns—those smaller than 30\,m—are fully visible in at least one tile, assuming a 75\% overlap. This ablation isolates the effects of spatial detail, object count, and input size. Each model is trained at a fixed resolution, with only minor cropping augmentation ($\pm 10\%$ of input size) before resizing to a fixed input size. Further experimental details are provided in App.~\ref{sec:appendix_resolution_imgsize}.

We also compare models trained at 6\,cm and 10\,cm GSD while resizing the inputs to assess the impact of both the resolution and input size on models performance.
Tile-level evaluation metrics (mAP$_{50:95}$ and mAR$_{50:95}$) are not comparable \textit{per se} between $40\times40$ and $80\times 80$\,m spatial extent since the tiles differ in object count and spatial boundaries.
But one may compare all results with the RF1$_{75}$ since it is computed at the raster level, after aggregation of individual images predictions.

\begin{table}[t!]
\parbox{\linewidth}{Tables 3 and 4: \textbf{Model, resolution and spatial extent selection on \textsc{SelvaBox}.} Comparison of performances on the proposed test set of \textsc{SelvaBox} with variable tile spatial extent, respectively $40\times40$\,m in Tab.~\ref{table:benchmark_40m} and $80\times80$\,m in Tab.~\ref{table:benchmark_80m}, input size in pixels and ground spatial distance (GSD) in cm. We highlight results per method and backbone as \darksquare the first, \middlesquare the second and \lightsquare the third best scores. We also \textbf{bold} and \underline{underline} the best and second best scores overall. Note that mAP$_{50:95}$ and mAR$_{50:95}$ cannot be compared between $40\times40$\,m and $80\times 80$\,m inputs as images do not match, but we can use RF1$_{75}$ to compare final post-aggregation results at the raster-level.}
\begin{minipage}{0.49\textwidth}
\centering
\vspace{0.3cm}
\caption{\textsc{SelvaBox} at $40\times40$\,m.}
\vspace{-0.2cm}
\resizebox{\textwidth}{!}{
\begin{tabular}{c | c c | c c c c}
\toprule

Method & GSD & I. size & mAP$_{50:95}$ & mAR$_{50:95}$ & RF1$_{75}$ \\
\midrule
\multirow{4}{*}{\multirowcell{3}{Faster \\ R-CNN \\ ResNet50}}
& 10 & 400 & 26.90 \scriptsize ($\pm0.13$) & 40.87 \scriptsize ($\pm0.35$) & 35.78 \scriptsize ($\pm0.44$) \\
& 10 & 666 & 28.40 \scriptsize ($\pm0.13$) & 42.79 \scriptsize ($\pm0.19$) & \lightgreen 37.75 \scriptsize ($\pm0.30$) \\
& 10 & 888 & 28.51 \scriptsize ($\pm0.20$) & 43.36 \scriptsize ($\pm0.19$) & 37.46 \scriptsize ($\pm0.91$) \\
& 6 & 666 & \lightgreen 29.31 \scriptsize ($\pm0.05$) & \lightgreen 43.59 \scriptsize ($\pm0.20$) & \darkgreen 39.97 \scriptsize ($\pm0.33$) \\
& 6 & 888 & \middlegreen 29.40 \scriptsize ($\pm0.34$) & \middlegreen 44.18 \scriptsize ($\pm0.44$) & \middlegreen 38.92 \scriptsize ($\pm0.51$) \\
& 4.5 & 888 & \darkgreen 30.25 \scriptsize ($\pm0.24$) & \darkgreen 45.18 \scriptsize ($\pm0.30$) & \darkgreen 39.97 \scriptsize ($\pm0.67$) \\
\midrule
 
\multirow{4}{*}{\multirowcell{3}{DINO \\ 4-scale \\ ResNet50}}
& 10 & 400 & 30.63 \scriptsize ($\pm0.24$) & 48.06 \scriptsize ($\pm0.33$) & 41.14 \scriptsize ($\pm0.80$) \\
& 10 & 666 & 31.76 \scriptsize ($\pm0.86$) & 50.40 \scriptsize ($\pm0.55$) & 41.57 \scriptsize ($\pm1.94$) \\
& 10 & 888 & 32.19 \scriptsize ($\pm0.33$) & 50.68 \scriptsize ($\pm0.19$) & 42.47 \scriptsize ($\pm0.97$) \\
& 6 & 666 & \lightgreen 33.46 \scriptsize ($\pm0.22$) & \lightgreen 51.80 \scriptsize ($\pm0.31$) & \darkgreen 44.55 \scriptsize ($\pm0.18$) \\
& 6 & 888 & \middlegreen 33.54 \scriptsize ($\pm0.40$) & \middlegreen 52.12 \scriptsize ($\pm0.18$) & \lightgreen 43.34 \scriptsize ($\pm0.79$) \\
& 4.5 & 888 & \darkgreen 34.19 \scriptsize ($\pm0.13$) & \darkgreen 52.53 \scriptsize ($\pm0.40$) & \middlegreen 44.26 \scriptsize ($\pm0.83$) \\
\midrule

\multirow{4}{*}{\multirowcell{3}{DINO \\ 5-scale \\ Swin L-384}}
& 10 & 400 & 33.84 \scriptsize ($\pm0.20$) & 52.02 \scriptsize ($\pm0.25$) & 45.37 \scriptsize ($\pm0.23$) \\
& 10 & 666 & 34.64 \scriptsize ($\pm0.25$) & 52.91 \scriptsize ($\pm0.30$) & 46.39 \scriptsize ($\pm0.52$) \\
& 10 & 888 & 34.92 \scriptsize ($\pm0.34$) & 53.23 \scriptsize ($\pm0.14$) & 45.22 \scriptsize ($\pm0.70$) \\
& 6 & 666 & \middlegreen \underline{37.07 \scriptsize ($\pm0.16$)} & \middlegreen \underline{55.18 \scriptsize ($\pm0.22$)} & \middlegreen \underline{48.50 \scriptsize ($\pm0.60$)} \\
& 6 & 888 & \lightgreen 36.22 \scriptsize ($\pm0.38$) & \lightgreen 54.55 \scriptsize ($\pm0.43$) & \lightgreen 48.13 \scriptsize ($\pm0.60$) \\
& 4.5 & 888 & \darkgreen \textbf{37.78}\scriptsize($\boldsymbol{\pm0.15}$) & \darkgreen \textbf{56.30} \scriptsize ($\boldsymbol{\pm0.21}$) & \darkgreen \textbf{49.76} \scriptsize ($\boldsymbol{\pm0.43}$) \\

\bottomrule
\end{tabular}
}
\label{table:benchmark_40m}
\end{minipage}
\hfill
\begin{minipage}{0.49\textwidth}
\centering
\vspace{0.3cm}
\caption{\textsc{SelvaBox} at $80\times80$\,m.}
\vspace{-0.2cm}
\resizebox{\textwidth}{!}{
\begin{tabular}{c | c c | c c c c}
\toprule

Method & GSD & I. size & mAP$_{50:95}$ & mAR$_{50:95}$ & RF1$_{75}$ \\
\midrule
\multirow{4}{*}{\multirowcell{3}{Faster \\ R-CNN \\ ResNet50}}
& 10 & 800 & 24.94 \scriptsize ($\pm0.34$) & 35.93 \scriptsize ($\pm0.55$) & 34.66 \scriptsize ($\pm0.97$) \\
& 10 & 1333 & 26.25 \scriptsize ($\pm0.14$) & 38.59 \scriptsize ($\pm0.41$) & 36.09 \scriptsize ($\pm0.51$) \\
& 10 & 1777 & \lightgreen 27.58 \scriptsize ($\pm0.24$) &\lightgreen 40.21 \scriptsize ($\pm0.38$) & 35.74 \scriptsize ($\pm1.26$) \\
& 6 & 1333 & 26.52 \scriptsize ($\pm0.80$) & 39.55 \scriptsize ($\pm0.75$) & \middlegreen 36.22 \scriptsize ($\pm1.45$) \\
& 6 & 1777 & \middlegreen 27.89 \scriptsize ($\pm0.35$) & \middlegreen 41.02 \scriptsize ($\pm0.69$) & \lightgreen 35.94 \scriptsize ($\pm0.84$) \\
& 4.5 & 1777 & \darkgreen 28.74 \scriptsize ($\pm0.44$) & \darkgreen 41.27 \scriptsize ($\pm0.59$) & \darkgreen 37.52 \scriptsize ($\pm0.58$) \\
\midrule
 
\multirow{4}{*}{\multirowcell{3}{DINO \\ 4-scale \\ ResNet50}}
& 10 & 800 & 30.90 \scriptsize ($\pm0.51$) & 47.29 \scriptsize ($\pm0.33$) & 41.20 \scriptsize ($\pm0.39$) \\
& 10 & 1333 & 32.39 \scriptsize ($\pm0.02$) & 49.22 \scriptsize ($\pm0.10$) & \lightgreen 43.08 \scriptsize ($\pm0.20$) \\
& 10 & 1777 & 32.51 \scriptsize ($\pm0.89$) & 49.35 \scriptsize ($\pm0.47$) & 42.39 \scriptsize ($\pm1.25$) \\
& 6 & 1333 & \lightgreen 33.06 \scriptsize ($\pm0.29$) & \lightgreen 49.93 \scriptsize ($\pm0.39$) & 42.92 \scriptsize ($\pm0.51$) \\
& 6 & 1777 & \middlegreen 33.62 \scriptsize ($\pm0.10$) & \middlegreen 50.85 \scriptsize ($\pm0.17$) & \darkgreen 44.18 \scriptsize ($\pm0.18$) \\
& 4.5 & 1777 & \darkgreen 33.81 \scriptsize ($\pm0.84$) & \darkgreen 51.00 \scriptsize ($\pm0.77$) &  \middlegreen 43.26 \scriptsize ($\pm0.45$) \\
\midrule

\multirow{4}{*}{\multirowcell{3}{DINO \\ 5-scale \\ Swin L-384}}
& 10 & 800 & 33.90 \scriptsize ($\pm0.09$) & 50.29 \scriptsize ($\pm0.38$) & 44.64 \scriptsize ($\pm0.20$) \\
& 10 & 1333 & 34.22 \scriptsize ($\pm0.34$) & 50.76 \scriptsize ($\pm0.57$) & 45.64 \scriptsize ($\pm1.03$) \\
& 10 & 1777 & 35.30 \scriptsize ($\pm0.26$) & 52.12 \scriptsize ($\pm0.62$) & 45.37 \scriptsize ($\pm0.08$) \\
& 6 & 1333 & \middlegreen \underline{37.12 \scriptsize ($\pm0.38$)} & \middlegreen \underline{53.56 \scriptsize ($\pm0.48$)} & \middlegreen \underline{47.81 \scriptsize ($\pm0.40$)} \\
& 6 & 1777 & \lightgreen 35.77 \scriptsize ($\pm0.84$) & \lightgreen 52.91 \scriptsize ($\pm0.56$) & \lightgreen 45.88 \scriptsize ($\pm1.97$) \\
& 4.5 & 1777 & \darkgreen \textbf{37.79} \scriptsize ($\boldsymbol{\pm0.55}$) & \darkgreen \textbf{54.66} \scriptsize ($\boldsymbol{\pm0.47}$) & \darkgreen \textbf{49.38} \scriptsize ($\boldsymbol{\pm0.76}$) \\

\bottomrule
\end{tabular}
}
\label{table:benchmark_80m}
\end{minipage}
\label{table:benchmark_comparison}
\end{table}

\paragraph{Multi-resolution approach.}

Diversity in camera sensors and recording conditions results in datasets with various resolutions (Tab.~\ref{table:related_datasets} and \ref{table:rasters_info}), complicating or preventing model training across multiple datasets.
We mitigate this through multi-resolution input augmentation that enforces scale-invariance during training, enabling us to combine datasets of various resolutions. 
This simple, yet efficient process randomly crops inputs using a wide range of crop sizes, then randomly resizes the crops. This achieves two effects: \circledgray{1} cropping performs ground extent augmentation, and \circledgray{2} resizing performs GSD augmentation. Details on our multi-resolution augmentation pipeline are in App.~\ref{sec:appendix_multires_example}.

While data augmentation generally improves generalization, extreme transformations may impact convergence and performance. Therefore, we train multi-resolution models on \textsc{SelvaBox} with increasingly large crop ranges (Fig.~\ref{fig:multi-res-rf1}) and the same random resize in the $[1024, 1777]$ pixel range, comparing them at $80\times80$\,m to the best single-resolution, single-input-size models from the previous experiment (\ie DINO Swin-384 at 4.5, 6 and 10\,cm; see Tab.~\ref{table:benchmark_80m}).

\subsection{Methodology to evaluate OOD generalization }
\label{subec:selvabox-ood-methodology}

To evaluate the generalization capabilities of models trained on \textsc{SelvaBox}, we define BCI50ha and Detectree2 (Tab.~\ref{table:related_datasets}) as OOD datasets for test-only evaluation.  
We perform zero-shot evaluations on these datasets, meaning models are tested without any fine-tuning on data completely excluded from training, and characterized by diverse resolutions, image quality, and forest types. 
These two datasets are considered OOD relative to \textsc{SelvaBox} because \circledgray{1} BCI50ha is located on an island in Panama (whereas \textsc{SelvaBox} is on mainland Panama), and Detectree2 is located in Malaysia, on a different continent; and \circledgray{2} both datasets were acquired using different drones, camera sensors, and flight conditions. 
Additionally, we include NeonTreeEvaluation, QuebecTrees, and OAM-TCD as either in-distribution or OOD datasets to assess how varying the number and diversity of datasets used during training affects model generalization.

\begin{wrapfigure}[15]{r}{0.40\textwidth}
\centering
\vspace{-2cm}
\includegraphics[width=0.35\textwidth]{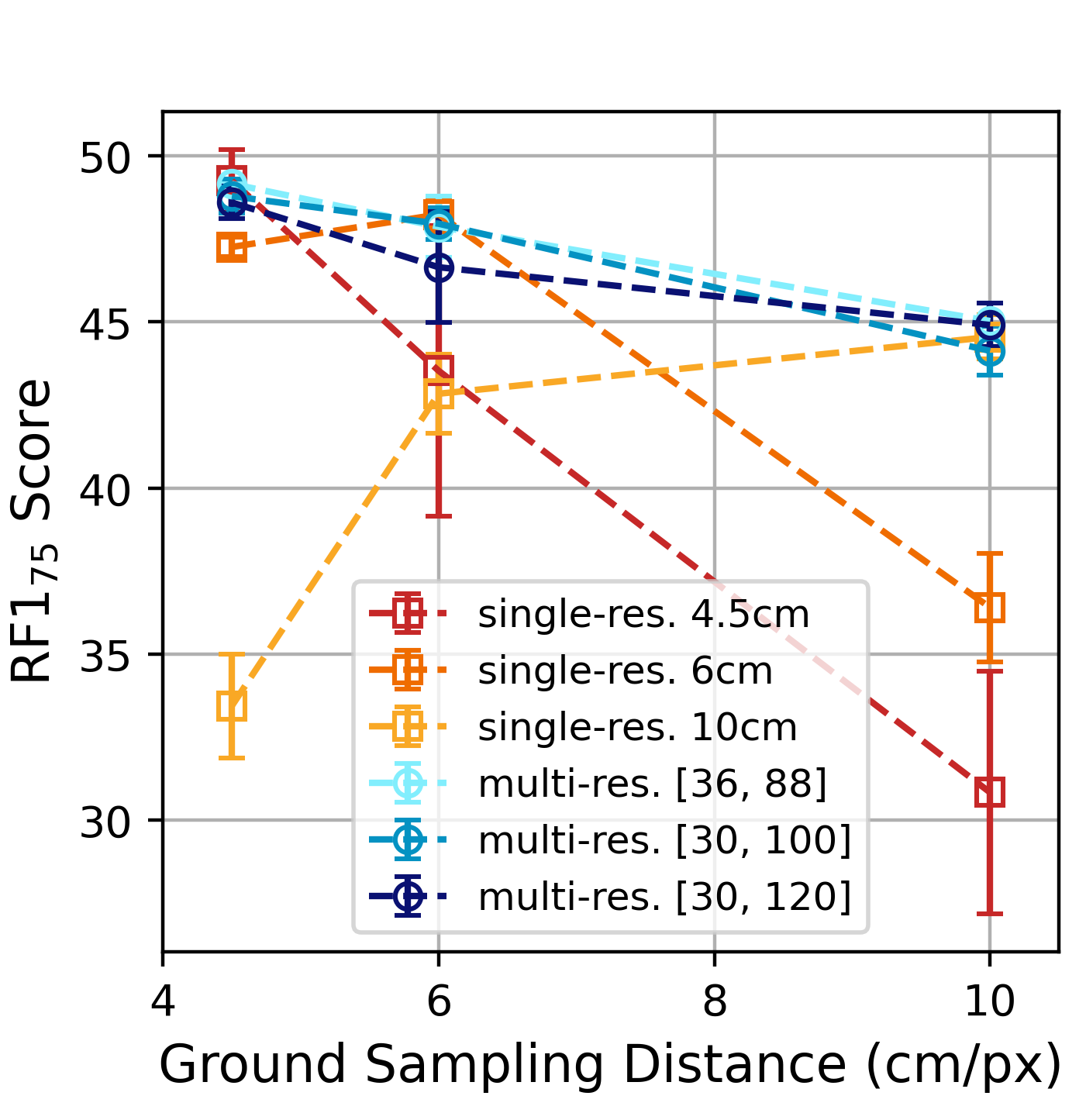}
\caption{\textbf{Multi-resolution vs. single-resolution on \textsc{SelvaBox}.} RF1$_{75}$ for the best single-resolution methods from Tab.~\ref{table:benchmark_80m} trained at fixed $80\times 80$\,m extent vs multi-resolution approaches with varying crop augmentation ranges $[36,88]$, $[30,100]$, $[30,120]$. All methods are `DINO 5-scale Swin L-384'.}
\label{fig:multi-res-rf1}
\end{wrapfigure}

We compare a multi-resolution model trained exclusively on \textsc{SelvaBox}, using a crop augmentation range of $[30, 120]$ meters (equivalent to $[666, 2666]$ pixels), against models trained on different combinations of OAM-TCD, NeonTreeEvaluation, QuebecTrees, and \textsc{SelvaBox} datasets (including DeepForest and Detectree2).
We selected this multi-resolution augmentation range based on our benchmark results (Sec.~\ref{sec:experiments}, Fig.~\ref{fig:multi-res-rf1}), which indicated that this range achieves performance comparable to single-resolution and less aggressive multi-resolution methods on \textsc{SelvaBox}, while also allowing spatial extents of images from different datasets to partially overlap (Tab.~\ref{table:preprocessed_datasets} in App.~\ref{sec:appendix_external_dataset_preprocessing}). Finally, we optimize non-maximum suppression (NMS) hyperparameters using the validation sets of \textsc{SelvaBox} and Detectree2, while keeping BCI50ha strictly zero-shot.
\newpage
\section{Experiments and results\label{sec:experiments}}
First, we evaluate model architectures, resolutions, and spatial extents on SelvaBox (Sec.~\ref{subsec:selvabox-results}).
Then, we validate our multi-resolution training methodology. Finally, we assess generalization on OOD datasets (Sec.~\ref{subsec:ood-results}).

\subsection{\textsc{SelvaBox} results}
\label{subsec:selvabox-results}
Using the methodology in Section~\ref{subsec:model-selection}, we find:

\paragraph{Resolution matters, transformers too.}
In Tables~\ref{table:benchmark_40m} and \ref{table:benchmark_80m}, we find that for all GSD and spatial extents, DINO outperforms Faster R-CNN, and Swin L-384 outperforms ResNet-50. 
We also observe significant improvements in mAP$_{50:95}$, mAR$_{50:95}$ and RF1$_{75}$ when using lower GSD for all architectures. 
While larger input sizes at fixed resolution benefits ResNet-50-based methods, DINO + Swin L-384 models do not see such improvements at 6\,cm per pixel. This suggests diminishing returns from further increases in input size, and only the Swin L-384 backbone fully leverages more detailed inputs.
Finally, we observe that Faster R-CNN reaches best RF1$_{75}$ performance at $40\times40$\,m rather than $80\times 80$\,m, likely due to larger context and higher number of objects making the task more difficult.

\paragraph{Multi-resolution is effective on \textsc{SelvaBox}.}
In Figure \ref{fig:multi-res-rf1}, we observe that all multi-resolution models achieve RF1$_{75}$ results within standard-deviation of the best single-resolution models, for all three resolutions. Additionally, single-resolution models struggle at test-time on unseen resolutions. Results for mAP$_{50:95}$ and mAR$_{50:95}$ are similar and presented in Appendix (Fig.~\ref{fig:appendix_multires_results_map_mar}). This demonstrates that a single multi-resolution model can be trained for transferability across spatial extents and GSDs without performance losses on \textsc{SelvaBox}, instead of training multiple resolution-specific models.

\subsection{OOD results}
\label{subsec:ood-results}

Following the methodology described in Sec.~\ref{subec:selvabox-ood-methodology}, we evaluate zero-shot generalization, we find:
\paragraph{\textsc{SelvaBox} exposes the limitations of current methods and datasets.}
We report results on tropical forests in Table~\ref{tab:results_tropical}.
First, existing methods, namely Detectree2 and DeepForest, perform poorly on \textsc{SelvaBox} in zero-shot evaluation with $6.08$ and $13.14$ RF1$_{75}$ respectively. 
Our method trained with multi-resolutions on NeonTreeEvaluation, QuebecTrees and OAM-TCD reaches $30.81$ RF1$_{75}$ on \textsc{SelvaBox} in zero-shot evaluation, showing great generalization performances on unseen tropical forests.
When \textsc{SelvaBox} is included in-distribution of the training process, our methods achieve state-of-the-art performances with $47.63$ (multi-datasets + \textsc{SelvaBox}) and $48.60$ (\textsc{SelvaBox} only) RF1$_{75}$.
These experimental results show how challenging \textsc{SelvaBox} is for existing methods, filling a gap not covered by existing datasets and methods. 
\paragraph{\textsc{SelvaBox} improves OOD generalization on tropical datasets.} We observe that models trained on \textsc{SelvaBox} achieve state-of-the-art performance in zero-shot evaluation on BCI50ha, at $39.39$ (multi-datasets + \textsc{SelvaBox}) and $41.91$ (\textsc{SelvaBox} only) RF1$_{75}$, followed by Detectree2-resize at $34.97$ RF1$_{75}$. 
On the Detectree2 dataset, the best performing model is Detectree2-resize in RF1$_{75}$ although a potential data leak could have occurred during the evaluation on their dataset, which limits the interpretation of the results, given that we were unable to recover the training-test splits originally used.
Our multi-dataset + \textsc{SelvaBox} method outperforms both Detectree2's models in terms of mAP$_{50:95}$ and mAR$_{50:95}$ on the Detectree2 dataset and beats DeepForest.  It also outperforms our multi-dataset without \textsc{SelvaBox} and \textsc{SelvaBox}-only methods, while being evaluated on a restricted zero-shot regime.
We include corresponding qualitative results in Appendix~\ref{sec:appendix_detectree2_qualitative}. 
To our knowledge, the DINO-Swin-L trained on multi-dataset + \textsc{SelvaBox} including a multi-resolution training process achieves state-of-the-art performance for the tropical tree crown detection task, generalizing well on both \textsc{SelvaBox} and OOD tropical datasets.

\paragraph{State-of-the-art performance on both tropical and non-tropical datasets.}
We present results on temperate and urban forests in Table~\ref{tab:results_temperate}.
We observe that both our multi-dataset methods (with and without \textsc{SelvaBox}) outperforms all the other in-distribution or OOD methods on temperate (NeonTreeEvaluation and QuebecTrees) and urban (OAM-TCD) datasets. 

Furthermore, training on $\textsc{SelvaBox}$ alone allows our methods to outperform competing approaches on both the QuebecTrees and OAM-TCD datasets, achieving better results on QuebecTrees than models trained on NeonTree (a global-scale temperate forest dataset), demonstrating $\textsc{SelvaBox}$'s quality and the generalization capacity of our training process.
We include corresponding qualitative results in Appendix~\ref{sec:appendix_other_datasets_qualitative}.
Our multi-dataset methods reached average performance within standard deviation for non-tropical datasets,
confirming that our multi-dataset approach with \textsc{SelvaBox} reaches state-of-the-art performance on both tropical and non-tropical datasets.

\begin{table}[t!]
  \vspace{0.0cm}
  \centering
    \caption{\textbf{Tropical datasets evaluation}. We respectively denote N for NeonTreeEvaluation, D for Detectree2, D+u for Detectree2 with urban regions, Q for QuebecTrees, O for OAM-TCD, S for \textsc{SelvaBox} and B for BCI50ha. We noted OD to identify out-of-distribution datasets, and RG for relative gain in RF1$_{75}$ of each method compared to the best competing one (Detectree2-rezise in this table). We mark the best and second-best scores in \textbf{bold} and \underline{underline}, respectively. We denote with {\color{BrickRed}\boldmath$\sim$} the Detectree2 competing methods where original train-test splits could not be recovered, preventing controlled evaluation on their dataset and limiting the interpretability of comparative results. Standard deviations (over three seeds) are reported only for models we trained ourselves, whereas Detectree2 and DeepForest (N) rely on single released model without variability estimates. }
  \setlength{\tabcolsep}{1pt} 
  \resizebox{\textwidth}{!}{%
    \begin{tabular}{c c *{3}{|ccc@{\hspace{-1.5pt}}c@{\hspace{-1.5pt}}c}}
      \toprule
        \multirow{2}{*}{\textbf{Method}}
        & \multirow{2}{*}{\makecell{Train\\set(s)}}
        & \multicolumn{5}{c}{\textsc{SelvaBox} (S)}
        & \multicolumn{5}{|c}{Detectree2 (D)}
        & \multicolumn{5}{|c}{BCI50ha (B)} \\
      \cmidrule(lr){3-7} \cmidrule(lr){8-12} \cmidrule(lr){13-17}
        & 
        & mAP$_{50:95}$ & mAR$_{50:95}$ & RF1$_{75}$ & OD & RG
        & mAP$_{50:95}$ & mAR$_{50:95}$ & RF1$_{75}$ & OD & RG
        & mAP$_{50:95}$ & mAR$_{50:95}$ & RF1$_{75}$ & OD & RG \\
      \midrule
      
        Detectree2-resize
        & D
        & 8.62 & 15.47 & 13.14 & \cmark & \footnotesize 0\%
        & \underline{17.67} & \underline{34.11} & \textbf{23.87} & {\color{BrickRed}\boldmath$\sim$} & \footnotesize \textbf{0\%}
        & 32.11 & 48.18 & 34.97 & \cmark & \footnotesize 0\%
      \\
      Detectree2-flexi
        & D+u
        & 6.43 & 13.20 & 9.21 & \cmark & \footnotesize \textcolor{BrickRed}{-30\%}
        & 6.43 & 19.86 & 4.46 & {\color{BrickRed}\boldmath$\sim$} & \footnotesize \textcolor{BrickRed}{-82\%}
        & 12.72 & 29.47 & 4.26 & \cmark & \footnotesize \textcolor{BrickRed}{-88\%}
      \\

      \midrule

        DeepForest 
        & N
        & 4.70
        & 9.08
        & 6.08
        & \cmark & \footnotesize \textcolor{BrickRed}{-54\%}
        & 6.85
        & 19.27
        & 7.83
        & \cmark & \footnotesize \textcolor{BrickRed}{-68\%}
        & 14.48
        & 25.50
        & 10.02
        & \cmark & \footnotesize \textcolor{BrickRed}{-72\%}
        \\

      F. R-CNN-RN50
        & N
        & 1.79\scriptsize($\pm0.21$) & 11.08\scriptsize($\pm0.01$) & 4.54\scriptsize($\pm0.33$) & \cmark & \footnotesize \textcolor{BrickRed}{-66\%}
        & 11.09\scriptsize($\pm1.58$) & 26.28\scriptsize($\pm2.38$) & 14.80\scriptsize($\pm2.57$) & \cmark & \footnotesize \textcolor{BrickRed}{-38\%}
        & 0.72\scriptsize($\pm0.12$) & 4.47\scriptsize($\pm0.95$) & 1.42\scriptsize($\pm0.18$) & \cmark & \footnotesize \textcolor{BrickRed}{-96\%}
        \\

      DINO-Swin-L
        & N
        & 5.67\scriptsize($\pm0.73$) & 17.63\scriptsize($\pm1.13$) & 9.94\scriptsize($\pm2.12$) & \cmark & \footnotesize \textcolor{BrickRed}{-25\%}
        & 14.77\scriptsize($\pm3.58$) & 32.62\scriptsize($\pm4.06$) & \underline{19.87\scriptsize($\pm4.12$)} & \cmark & \footnotesize \textcolor{BrickRed}{\underline{-17\%}}
        & 1.74\scriptsize($\pm0.35$) & 11.89\scriptsize($\pm0.51$) & 3.77\scriptsize($\pm0.59$) & \cmark & \footnotesize \textcolor{BrickRed}{-90\%}
        \\

      \midrule

        DeepForest
        & S
        & 28.84\scriptsize($\pm0.19$) & 44.67\scriptsize($\pm0.09$) & 38.00\scriptsize($\pm0.22$) & \xmark & \footnotesize \textcolor{ForestGreen}{+189\%}
        & 6.34\scriptsize($\pm1.11$) & 18.35\scriptsize($\pm1.78$) & 2.71\scriptsize($\pm0.67$) & \cmark & \footnotesize \textcolor{BrickRed}{-89\%}
        & 25.17\scriptsize($\pm1.09$) & 46.85\scriptsize($\pm0.71$) & 36.46\scriptsize($\pm1.38$) & \cmark & \footnotesize \textcolor{ForestGreen}{+4\%}
        \\

      F. R-CNN-RN50
        & S
        & 28.49\scriptsize($\pm0.05$) & 41.88\scriptsize($\pm0.25$) & 36.37\scriptsize($\pm0.37$) & \xmark & \textcolor{ForestGreen}{+176\%}
        & 3.32\scriptsize($\pm0.76$) & 11.50\scriptsize($\pm1.31$) & 1.04\scriptsize($\pm0.66$) & \cmark & \footnotesize \textcolor{BrickRed}{-96\%}
        & 27.23\scriptsize($\pm1.49$) & 46.70\scriptsize($\pm1.64$) & 31.24\scriptsize($\pm1.39$) & \cmark & \footnotesize \textcolor{BrickRed}{-11\%}
        \\
        
      DINO-Swin-L
        & S
        & \textbf{37.77}\scriptsize($\boldsymbol{\pm0.35}$)
        & \textbf{54.69}\scriptsize($\boldsymbol{\pm0.07}$)
        & \textbf{48.60}\scriptsize($\boldsymbol{\pm0.49}$)
        & \xmark 
        & \footnotesize \textcolor{ForestGreen}{\textbf{+269\%}}
        & 13.27\scriptsize($\pm1.80$) 
        & 28.24\scriptsize($\pm2.75$) 
        & 8.47\scriptsize($\pm3.13$) 
        & \cmark 
        & \footnotesize \textcolor{BrickRed}{-65\%}
        & \textbf{36.87}\scriptsize($\boldsymbol{\pm0.67}$)
        & \textbf{60.30}\scriptsize($\boldsymbol{\pm0.90}$)
        & \textbf{41.91}\scriptsize($\boldsymbol{\pm1.28}$)
        & \cmark 
        & \footnotesize \textcolor{ForestGreen}{\textbf{+19\%}}
        \\

      \midrule

        DeepForest
        & NQO
        & 14.93\scriptsize($\pm1.23$) 
        & 31.76\scriptsize($\pm1.05$) 
        & 21.55\scriptsize($\pm1.57$) 
        & \cmark 
        & \footnotesize \textcolor{ForestGreen}{+64\%}
        & 10.96\scriptsize($\pm1.78$) 
        & 26.14\scriptsize($\pm2.20$) 
        & 8.19\scriptsize($\pm2.46$) 
        & \cmark & \footnotesize \textcolor{BrickRed}{-66\%}
        & 10.84\scriptsize($\pm0.94$) 
        & 31.13\scriptsize($\pm2.08$) 
        & 18.58\scriptsize($\pm1.10$)
        & \cmark & \footnotesize \textcolor{BrickRed}{-47\%}
        \\

      F. R-CNN-RN50
        & NQO
        & 16.39\scriptsize($\pm0.11$) 
        & 29.39\scriptsize($\pm0.11$) 
        & 24.77\scriptsize($\pm0.38$) 
        & \cmark 
        & \footnotesize \textcolor{ForestGreen}{+88\%}
        & 12.50\scriptsize($\pm0.42$) 
        & 28.17\scriptsize($\pm0.64$) 
        & 13.65\scriptsize($\pm0.92$) 
        & \cmark 
        & \footnotesize \textcolor{BrickRed}{-43\%}
        & 11.92\scriptsize($\pm3.43$) 
        & 32.74\scriptsize($\pm4.29$) 
        & 16.16\scriptsize($\pm3.36$)
        & \cmark 
        & \footnotesize \textcolor{BrickRed}{-54\%}
        \\
      
      DINO-Swin-L
        & NQO
        & 20.85\scriptsize($\pm1.46$) 
        & 39.87\scriptsize($\pm1.66$) 
        & 30.81\scriptsize($\pm1.53$) 
        & \cmark 
        & \footnotesize \textcolor{ForestGreen}{+134\%}
        & 15.35\scriptsize($\pm1.88$) 
        & 30.51\scriptsize($\pm2.72$) 
        & 11.31\scriptsize($\pm2.55$) 
        & \cmark 
        & \footnotesize \textcolor{BrickRed}{-53\%}
        & 25.72\scriptsize($\pm1.92$) 
        & 48.78\scriptsize($\pm1.72$) 
        & 25.32\scriptsize($\pm1.87$)
        & \cmark 
        & \footnotesize \textcolor{BrickRed}{-28\%}
        \\
        
      \midrule

      DeepForest
        & NQOS
        & 27.58\scriptsize($\pm0.54$) & 43.69\scriptsize($\pm0.44$) & 35.92\scriptsize($\pm1.20$) & \xmark & \footnotesize \textcolor{ForestGreen}{+173\%}
        & 12.77\scriptsize($\pm0.31$) & 29.39\scriptsize($\pm0.36$) & 9.13\scriptsize($\pm0.45$) & \cmark & \footnotesize \textcolor{BrickRed}{-62\%}
        & 19.53\scriptsize($\pm1.92$) & 43.52\scriptsize($\pm3.19$) & 28.00\scriptsize($\pm3.81$) & \cmark & \footnotesize \textcolor{BrickRed}{-20\%}
        \\

      F. R-CNN-RN50
        & NQOS
        & 24.93\scriptsize($\pm1.10$) & 39.34\scriptsize($\pm0.38$) & 30.56\scriptsize($\pm1.44$) & \xmark & \footnotesize \textcolor{ForestGreen}{+132\%}
        & 13.80\scriptsize($\pm1.91$) & 29.84\scriptsize($\pm2.79$) & 14.42\scriptsize($\pm2.69$) & \cmark & \footnotesize \textcolor{BrickRed}{-40\%}
        & 20.42\scriptsize($\pm1.48$) & 43.25\scriptsize($\pm1.56$) & 23.49\scriptsize($\pm1.17$) & \cmark & \footnotesize \textcolor{BrickRed}{-33\%}
        \\
        
      DINO-Swin-L
        & NQOS
        & \underline{36.95\scriptsize($\pm0.56$)}
        & \underline{53.71\scriptsize($\pm0.32$)}
        & \underline{47.63\scriptsize($\pm0.23$)}
        & \xmark 
        & \footnotesize \textcolor{ForestGreen}{\underline{+262\%}}
        & \textbf{18.20}\scriptsize($\boldsymbol{\pm3.22}$)
        & \textbf{35.20}\scriptsize($\boldsymbol{\pm3.61}$)
        & 19.23\scriptsize($\pm3.33$)
        & \cmark 
        & \footnotesize \textcolor{BrickRed}{-20\%}
        & \underline{33.13\scriptsize($\pm3.06$)}
        & \underline{58.36\scriptsize($\pm2.21$)}
        & \underline{39.39\scriptsize($\pm1.71$)}
        & \cmark 
        & \footnotesize \textcolor{ForestGreen}{\underline{+12\%}}
        \\
      \bottomrule
    \end{tabular}%
  }
  \label{tab:results_tropical}
  \vspace{-0.5cm}
\end{table}
\vspace{-.25cm}
\begin{table}[t!]
  \vspace{0.5cm}
  \centering
    \caption{\textbf{Non-tropical datasets evaluation}. We respectively denote N for NeonTreeEvaluation, D for Detectree2, D+u for Detectree2 with urban regions, Q for QuebecTrees, O for OAM-TCD, S for \textsc{SelvaBox} and B for BCI50ha. We noted OD to identify out-of-distribution datasets, and RG for relative gain in RF1$_{75}$ if available, mAP$_{50:95}$ otherwise, of each method compared to the best competing one (either DeepForest (N) or Detectree2-flexi in this table). We mark the best and second-best scores in \textbf{bold} and \underline{underline}, respectively. We cannot compute RF1$_{75}$ for NeonTreeEvaluation and OAM-TCD as only individual images are available for their test splits. Standard deviations (over three seeds) are reported only for models we trained ourselves, whereas Detectree2 and DeepForest (N) rely on single released model without variability estimates.
    }
  \setlength{\tabcolsep}{1pt} 
  \resizebox{\textwidth}{!}{%
    \begin{tabular}{c c *{3}{|ccc@{\hspace{0pt}}c@{\hspace{-1.5pt}}c}}
      \toprule
        \multirow{2}{*}{\textbf{Method}}
        & \multirow{2}{*}{\makecell{Train\\set(s)}}
        & \multicolumn{5}{c}{NeonTreeEvaluation (N)}
        & \multicolumn{5}{|c}{QuebecTrees (Q)}
        & \multicolumn{5}{|c}{OAM-TCD (O)} \\
      \cmidrule(lr){3-7} \cmidrule(lr){8-12} \cmidrule(lr){13-17}
        & 
        & mAP$_{50:95}$ & mAR$_{50:95}$ & RF1$_{75}$ & OD & RG
        & mAP$_{50:95}$ & mAR$_{50:95}$ & RF1$_{75}$ & OD & RG
        & mAP$_{50:95}$ & mAR$_{50:95}$ & RF1$_{75}$ & OD & RG \\
      \midrule

      Detectree2-resize
        & D
        & 4.09 & 15.67 & \footnotesize N/A & \cmark & \footnotesize \textcolor{BrickRed}{-78\%}
        & 7.62 & 13.85 & 13.98 & \cmark & \footnotesize \textcolor{BrickRed}{-11\%}
        & 2.45 & 12.43 & \footnotesize N/A & \cmark & \footnotesize \textcolor{BrickRed}{-61\%}
        \\

      Detectree2-flexi
        & D+u
        & 1.75 & 9.86 & \footnotesize N/A & \cmark & \footnotesize \textcolor{BrickRed}{-91\%}
        & 9.75 & 16.59 & 15.60 & \cmark & \footnotesize 0\%
        & 5.20 & 13.21 & \footnotesize N/A & \cmark & \footnotesize \textcolor{BrickRed}{-16\%}
        \\

      \midrule

      DeepForest 
        & N
        & 18.06
        & 25.82
        & \footnotesize N/A
        & \xmark
        & \footnotesize 0\%
        & 3.58
        & 7.32
        & 4.82
        & \cmark
        & \footnotesize \textcolor{BrickRed}{-70\%}
        & 6.19
        & 11.42
        & \footnotesize N/A
        & \cmark
        & \footnotesize 0\%
        \\

      F. R-CNN-RN50
        & N
        & 17.08\scriptsize($\pm0.31$) & 27.16\scriptsize($\pm0.09$) & \footnotesize N/A & \xmark & \footnotesize \textcolor{BrickRed}{-6\%}
        & 5.97\scriptsize($\pm0.45$) & 18.39\scriptsize($\pm0.64$) & 10.66\scriptsize($\pm0.19$) & \cmark & \footnotesize \textcolor{BrickRed}{-32\%}
        & 9.75\scriptsize($\pm0.23$) & 18.85\scriptsize($\pm0.82$) & \footnotesize N/A & \cmark & \footnotesize \textcolor{ForestGreen}{+57\%}
        \\

      DINO-Swin-L
        & N
        & \underline{23.68\scriptsize($\pm0.20$)} & \underline{35.18\scriptsize($\pm0.20$)} & \footnotesize N/A & \xmark & \footnotesize \textcolor{ForestGreen}{\underline{+31\%}}
        & 10.46\scriptsize($\pm2.60$) & 23.47\scriptsize($\pm3.34$) & 14.20\scriptsize($\pm4.13$) & \cmark & \footnotesize \textcolor{BrickRed}{-9\%}
        & 18.42\scriptsize($\pm1.66$) & 29.91\scriptsize($\pm1.40$) & \footnotesize N/A & \cmark & \footnotesize \textcolor{ForestGreen}{+197\%}
        \\

      \midrule

      DeepForest
        & S
        & 1.16\scriptsize($\pm0.14$) & 5.52\scriptsize($\pm0.94$) & \footnotesize N/A & \cmark & \footnotesize \textcolor{BrickRed}{-94\%}
        & 21.46\scriptsize($\pm0.47$) & 36.29\scriptsize($\pm0.25$) & 31.09\scriptsize($\pm1.03$) & \cmark & \footnotesize \textcolor{ForestGreen}{+99\%}
        & 9.68\scriptsize($\pm1.12$) & 21.95\scriptsize($\pm1.25$) & \footnotesize N/A & \cmark & \footnotesize \textcolor{ForestGreen}{+56\%}
        \\

      F. R-CNN-RN50
        & S
        & 0.63\scriptsize($\pm0.19$) & 2.98\scriptsize($\pm0.53$) & \footnotesize N/A & \cmark & \footnotesize \textcolor{BrickRed}{-97\%}
        & 17.65\scriptsize($\pm0.27$) & 30.71\scriptsize($\pm0.46$) & 26.10\scriptsize($\pm0.88$) & \cmark & \footnotesize \textcolor{ForestGreen}{+67\%}
        & 8.50\scriptsize($\pm0.50$) & 16.17\scriptsize($\pm1.09$) & \footnotesize N/A & \cmark & \footnotesize \textcolor{ForestGreen}{+37\%}
        \\

      DINO-Swin-L
        & S
        & 5.16\scriptsize($\pm0.57$) & 14.67\scriptsize($\pm1.47$) & \footnotesize N/A & \cmark & \footnotesize \textcolor{BrickRed}{-72\%}
        & 27.34\scriptsize($\pm2.63$) & 44.04\scriptsize($\pm2.69$) & 38.34\scriptsize($\pm2.43$) & \cmark & \footnotesize \textcolor{ForestGreen}{+145\%}
        & 22.58\scriptsize($\pm0.31$) & 35.59\scriptsize($\pm0.52$) & \footnotesize N/A & \cmark & \footnotesize \textcolor{ForestGreen}{+264\%}
        \\

      \midrule

      DeepForest
        & NQO
        & 20.50\scriptsize($\pm0.26$) & 31.13\scriptsize($\pm0.15$) & \footnotesize N/A & \xmark & \footnotesize \textcolor{ForestGreen}{+13\%}
        & 36.75\scriptsize($\pm0.37$) & 49.66\scriptsize($\pm0.58$) & 47.37\scriptsize($\pm0.22$) & \xmark & \footnotesize \textcolor{ForestGreen}{+203\%}
        & 39.00\scriptsize($\pm0.21$) & 49.78\scriptsize($\pm0.18$) & \footnotesize N/A & \xmark & \footnotesize \textcolor{ForestGreen}{+530\%}
        \\

      F. R-CNN-RN50
        & NQO
        & 17.94\scriptsize($\pm0.10$) & 28.04\scriptsize($\pm0.16$) & \footnotesize N/A & \xmark & \footnotesize \textcolor{BrickRed}{-1\%}
        & 33.45\scriptsize($\pm0.84$) & 45.68\scriptsize($\pm1.02$) & 43.65\scriptsize($\pm0.92$) & \xmark & \footnotesize \textcolor{ForestGreen}{+179\%}
        & 38.34\scriptsize($\pm0.26$) & 47.76\scriptsize($\pm0.31$) & \footnotesize N/A & \xmark & \footnotesize \textcolor{ForestGreen}{+519\%}
        \\

      DINO-Swin-L
        & NQO
        & 23.50\scriptsize($\pm0.78$) & 34.85\scriptsize($\pm0.80$) & \footnotesize N/A & \xmark & \footnotesize \textcolor{ForestGreen}{+30\%}
        & \underline{44.53\scriptsize($\pm1.19$)} & \underline{58.48\scriptsize($\pm1.00$)} & \textbf{56.53}\scriptsize($\boldsymbol{\pm0.64}$) & \xmark & \footnotesize \textcolor{ForestGreen}{\textbf{+262\%}}
        & \textbf{44.29}\scriptsize($\boldsymbol{\pm0.33}$) & \textbf{55.57}\scriptsize($\boldsymbol{\pm0.41}$) & \footnotesize N/A & \xmark & \footnotesize \textcolor{ForestGreen}{\textbf{+615\%}}
        \\

      \midrule

      DeepForest
        & NQOS
        & 20.71\scriptsize($\pm0.25$) & 32.14\scriptsize($\pm0.13$) & \footnotesize N/A & \xmark & \footnotesize \textcolor{ForestGreen}{+14\%}
        & 36.53\scriptsize($\pm0.35$) & 49.66\scriptsize($\pm0.55$) & 47.04\scriptsize($\pm0.61$) & \xmark & \footnotesize \textcolor{ForestGreen}{+201\%}
        & 38.37\scriptsize($\pm0.46$) & 49.38\scriptsize($\pm0.24$) & \footnotesize N/A & \xmark & \footnotesize \textcolor{ForestGreen}{+519\%}
        \\

      F. R-CNN-RN50
        & NQOS
        & 18.47\scriptsize($\pm0.16$) & 28.80\scriptsize($\pm0.23$) & \footnotesize N/A & \xmark & \footnotesize \textcolor{ForestGreen}{+2\%}
        & 31.98\scriptsize($\pm0.45$) & 45.10\scriptsize($\pm0.31$) & 42.06\scriptsize($\pm0.73$) & \xmark & \footnotesize \textcolor{ForestGreen}{+169\%}
        & 38.08\scriptsize($\pm0.31$) & 47.87\scriptsize($\pm0.28$) & \footnotesize N/A & \xmark & \footnotesize \textcolor{ForestGreen}{+515\%}
        \\

      DINO-Swin-L
        & NQOS
        & \textbf{23.90}\scriptsize($\boldsymbol{\pm0.49}$) & \textbf{35.53}\scriptsize($\boldsymbol{\pm0.50}$) & \footnotesize N/A & \xmark & \footnotesize \textcolor{ForestGreen}{\textbf{+32\%}}
        & \textbf{45.05}\scriptsize($\boldsymbol{\pm0.59}$) & \textbf{58.74}\scriptsize($\boldsymbol{\pm0.56}$) & \underline{56.41\scriptsize($\pm0.87$)} & \xmark & \footnotesize \textcolor{ForestGreen}{\underline{+261\%}}
        & \underline{44.03\scriptsize($\pm0.53$)} & \underline{55.34\scriptsize($\pm0.67$)} & \footnotesize N/A & \xmark & \footnotesize \textcolor{ForestGreen}{\underline{+611\%}}
        \\

      \bottomrule
    \end{tabular}%
  }
  \label{tab:results_temperate}
  \vspace{-0.5cm}
\end{table}
\clearpage

\subsection{Ablation of RF1 vs IoU}
\begin{wrapfigure}[18]{r}{0.5\textwidth}
\vspace{-1.3cm}
\centering
\includegraphics[width=0.5\textwidth]{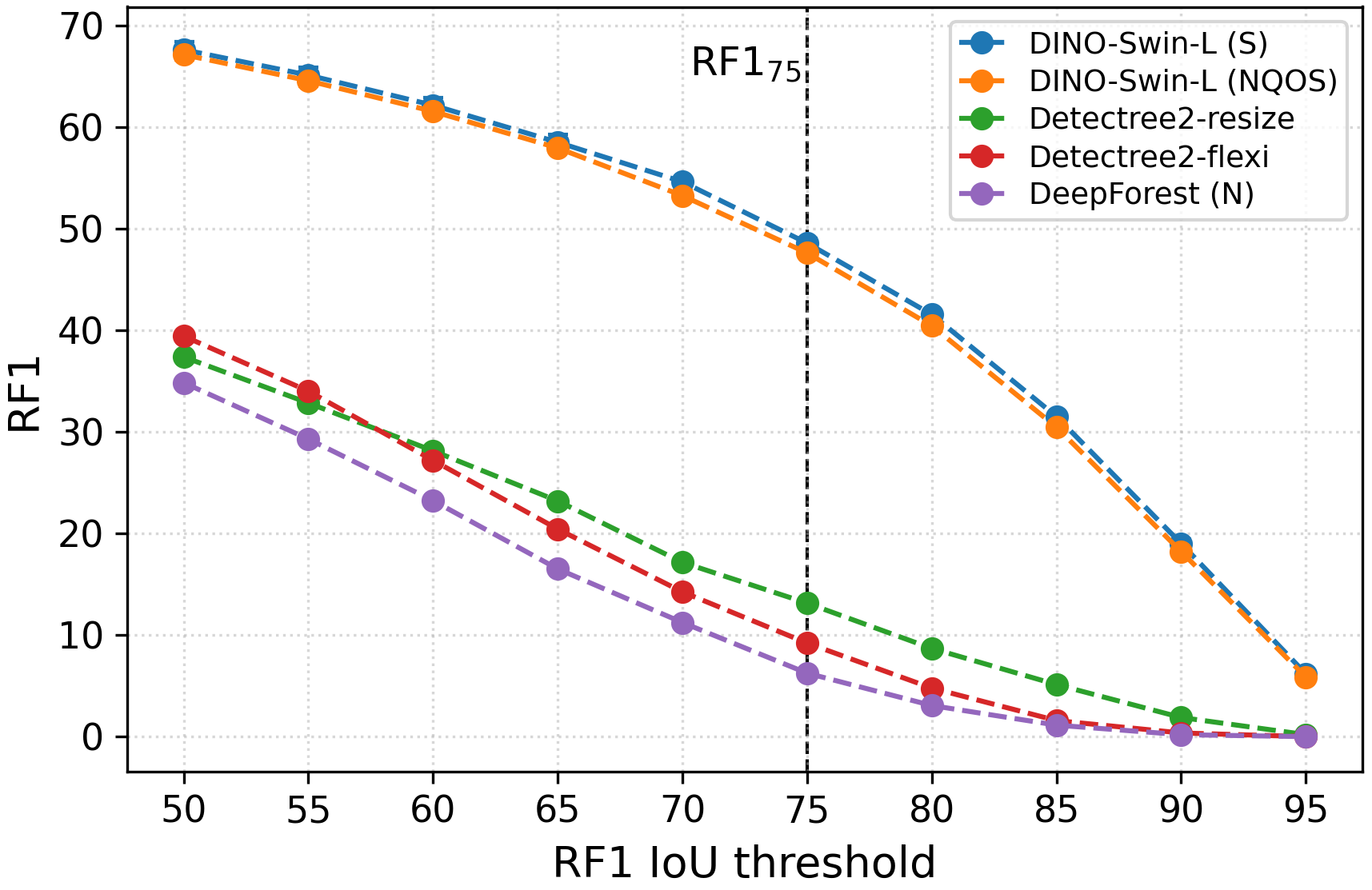}
\caption{\textbf{RF1 vs IoU threshold on $\textsc{SelvaBox}$.} Comparison of two of our DINO-Swin-L variants and competing methods at different IoU thresholds. In this work we focus on RF1$_{75}$ (IoU 75). For each IoU threshold, NMS hyperparameters are independently optimized on the validation set. Results for other datasets are in Appendix~\ref{section:rf1_iou_ablation}.}
\label{fig:rf1_vs_iou_selvabox_mainpaper}
\end{wrapfigure}

To better understand how the RF1 metric varies across IoU thresholds other than 0.75, we plotted RF1 as a function of the IoU threshold (0.50 to 0.95) for $\textsc{SelvaBox}$ (Fig.~\ref{fig:rf1_vs_iou_selvabox_mainpaper}) as well as BCI50ha, Detectree2 and QuebecTrees (App.~\ref{section:rf1_iou_ablation}).
NMS hyperparameters were optimized independently on the validation set independently for each IoU threshold. Results on $\textsc{SelvaBox}$ are consistent with Tables~\ref{tab:results_tropical} and \ref{tab:results_temperate}: we observe a consistent, substantial gap between our in-distribution DINO-Swin-L variants and competing methods (DeepForest and both Detectree2 methods) across all IoU thresholds.  
On BCI50ha and Detectree2 datasets, the Detectree2-resize baseline exhibits a local performance peak around $\text{IoU}=0.70$, sometimes unexpectedly exceeding its scores at lower thresholds (0.50–0.65), which we attribute to per-threshold NMS tuning and the higher variance induced by the substantially smaller size of these datasets. This behavior further underlines the benefits of SelvaBox's scale, where RF1 is less sensitive to annotation noise and crown size distribution. As a natural extension, we leave for future work the design of an RF1$_{50:95}$ metric, analogous to mAP$_{50:95}$, in which NMS hyperparameters would be tuned against the average RF1 over multiple IoU thresholds.

\subsection{Practical advice}
We recommend DINO-Swin-L (NQOS) as the default model for most applications and forest types, as it ranks first or a close second on all datasets. The exception is high-resolution tropical drone imagery without water or human constructions, where DINO-Swin-L (S) is preferable. We also recommend tuning NMS hyperparameters, tile extent and ground resolution on a validation set (if available), especially if the trees of interest are either small or very large.

\section{Ethical considerations and responsible use}
$\textsc{SelvaBox}$ and the released models are intended to support ecological research and operational forest monitoring (e.g., biodiversity assessment, biomass and carbon-stock estimation), not to facilitate activities such as illegal logging, land grabbing, or other forms of environmentally harmful exploitation, nor actions that could undermine the rights and livelihoods of local and Indigenous communities. We release $\textsc{SelvaBox}$ under a CC-BY~4.0 license and code and model checkpoints under an Apache-2.0 license, both permissive licenses. Although annotations were produced and reviewed by expert biologists, the dataset and resulting models inevitably contain noise and biases, and metrics such as RF1$_{75}$ remain sensitive to annotation completeness and evaluation settings. Outputs from $\textsc{SelvaBox}$-trained models should therefore be treated as decision-support tools rather than definitive measurements, and not used in isolation for high-stakes management or policy decisions.

\section{Conclusion}
\label{sec:conclusion_limitations}
We present \textsc{SelvaBox}, the largest tropical tree crown detection dataset to date, with over $83,000$ expert-verified annotations from high-resolution UAV imagery across Central and South American forests.
We achieve state-of-the-art performance across in-distribution and out-of-distribution benchmarks in a zero-shot setting training on \textsc{SelvaBox} and other open-access datasets.
We advocate for the RF1$_{75}$ metric, a raster-level score reflecting forest monitoring needs, and suggest that future work explore an IoU-averaged RF1$_{50:95}$ metric, as well as alternative aggregation methods such as soft-NMS~\citep{8237855} or weighted boxes fusion~\citep{SOLOVYEV2021104117}.
Our dataset, code, and models are fully open to support research in forest monitoring, while acknowledging the potential risks of misuse for illegal exploitation.

\section*{Reproducibility Statement}
All code, data, and experimental details required to reproduce the results of this paper are made available.
The \textsc{SelvaBox} dataset is described in Sections~\ref{sec:dataset_description} and \ref{sec:benchmarking_methods}, with additional details on orthomosaics, splits and annotations in App.~\ref{sec:appendix_selvabox_description}.
The ML-ready \textsc{SelvaBox} dataset is available on HuggingFace and linked on the first page of this manuscript.
The raster-level annotations and AOIs in geopackage format are available on HuggingFace in a separate branch.
Preprocessing steps for external datasets benchmarked in this manuscript are described in App.~\ref{sec:appendix_external_dataset_preprocessing}, and we also release these preprocessed versions on HuggingFace.
Our open-access data preprocessing package, \textit{geodataset}, and our benchmark, inference and training GitHub repository, \textit{CanopyRS}, are described in App.~\ref{sec:python_libraries} and linked on the first page of this manuscript.
The main training hyperparameters and compute setup are described in App.~\ref{sec:appendix_training_hyperparameters}.
The RF1$_{75}$ metric pseudo-code implementation and related inference hyperparameters used in our benchmarks can be found in App.~\ref{sec:appendix_nms_hyperparameters}.
Finally, model weights of our best methods as well as smaller model variants are available on HuggingFace and \textit{CanopyRS} package.

\section*{Acknowledgments}
This project was undertaken thanks to funding from IVADO, including the PRF3 project `AI, biodiversity, and Climate Change', the Canada First Research Excellence Fund, the Canada Research Chair and a Discovery Grant from NSERC to EL, and funding from the Mitacs institute. 
We thank the many people who helped with the acquisition of data (drone imagery and labels), notably: 
Sabrina Demers-Thibeault,
Vincent Le Falher,
Marie-Jeanne Gascon-DeCelles,
Simone Aubé,
Chloé Fiset,
Maxime Têtu-Frégeau,
Frédérik Senez, 
Gonzalo Rivas-Torres, 
the Outreach Robotics team (especially Hugues Lavigne and Julien Rachiele-Tremblay),
Paulo Sérgio,
Adriana Simonetti Peixoto,
Caroline Vasconcelos,
Daniel Magnobosco Marra,
Jefferson Hall,
Guillaume Tougas, and
Isabelle Lefebvre.
We also thank Mila for the compute resources.

\bibliography{iclr2026_conference}
\bibliographystyle{iclr2026_conference}

\clearpage

\appendix

\section*{Appendices \& supplementary material
}
\addcontentsline{toc}{section}{Appendix — Supplementary Material}

\section{The \textsc{SelvaBox} dataset\label{sec:appendix_selvabox_description}}

\subsection{Orthomosaics.\label{sec:appendix_orthomosaics}}
The RGB orthomosaics were generated in Agisoft Metashape version 2.1.
Images were acquired by flying at a constant elevation above the canopy. We kept a forward overlap of $>80\%$ and a side overlap of $>70\%$. Images were acquired around mid-day to minimize shadows. Sky conditions ranged from full sun to overcast.

The main Metashape parameters used for all of our orthomosaic reconstructions were:
\begin{itemize}
    \item Alignment accuracy: High
    \item Point cloud quality: High
    \item Point cloud filtering: Disabled
    \item Orthomosaic blending mode: Mosaic
\end{itemize}

\begin{table}[h]
\caption{\textbf{\textsc{SelvaBox} orthomosaics}. We denote each type of DJI drone as `m3e' for Mavic 3 Enterprise, `m3m' for Mavic 3 Multispectral, `mavicpro' for Mavic Pro, `mini2' for Mavic Mini 2.}
\centering
\resizebox{.99\linewidth}{!}{
\begin{tabular}{l c c c c c c c c c c}
\toprule
Raster name
& Drone
& Country
& Date
& \makecell{Sky\\ conditions}
& \makecell{GSD\\ (cm/px)}
& Forest type
& \#Hectares 
&\#Annotations 
& Proposed split(s) \\
\midrule
zf2quad      &  m3m & Brazil & 2024-01-30 & clear & 2.3 & primary & 15.5 & 1343 & valid \\
zf2tower     &  m3m & Brazil & 2024-01-30 & clear & 2.2 & primary & 9.5 & 1716 & test \\
zf2transectew &  m3m & Brazil & 2024-01-30 & clear & 1.5 & primary & 2.6 & 359 & train \\
zf2campinarana &  m3m & Brazil & 2024-01-31 & clear & 2.3 & primary & 66 & 16396 & train \\
transectotoni & mavicpro & Ecuador & 2017-08-10 & cloudy & 4.3 & primary & 4.3 & 5119 & train \\
tbslake   &  m3m & Ecuador & 2023-05-25 & clear & 5.1 & primary & 19 & 1279 & train, test \\
sanitower  &  mini2 & Ecuador & 2023-09-11 & cloudy & 1.8 & primary & 5.8 & 1721 & train \\
inundated    &  m3e & Ecuador & 2023-10-18 & cloudy & 2.2 & primary & 68 & 9075 & train, valid, test \\
pantano      &  m3e & Ecuador & 2023-10-18 & cloudy & 1.9 & primary & 41 & 4193 & train \\
terrafirme   &  m3e & Ecuador & 2023-10-18 & clear & 2.4 & primary & 110 & 6479 & train \\
asnortheast & m3m & Panama & 2023-12-07 & partial cloud & 1.3 & plantations, secondary & 33 & 12930 & train, valid, test \\
asnorthnorth  & m3m & Panama & 2023-12-07 & cloud & 1.2 & plantations, secondary & 15 & 6020 & train \\
asforestnorthe2 & m3m & Panama & 2023-12-08 & clear & 1.5 & secondary & 20 & 5925 & valid, test \\
asforestsouth2  & m3m & Panama & 2023-12-08 & clear & 1.6 & secondary & 28 & 10582 & train \\
\bottomrule
\end{tabular}}
\label{table:rasters_info}
\end{table}

\begin{table}[h]
\begin{center}
\caption{\textbf{\textsc{SelvaBox} boxes details.} Details of number of boxes for each raster, country and overall as well as their minimum, maximum and median box size expressed in meters.}
\resizebox{.99\linewidth}{!}{
\begin{tabular}{l l l | c c c c}
\toprule

Country & Location & Raster name & \# Boxes & Min box size (m) & Max box size (m) & Median box size (m) \\
\midrule
\multirow{5}{*}{Brazil} & \multirow{5}{*}{ZF2} 
& 20240130\_zf2quad\_m3m & 1343 & 1.02 & 33.00 & 6.34 \\
& & 20240130\_zf2tower\_m3m	& 1716 & 0.97 & 28.71 & 6.16  \\
& & 20240130\_zf2transectew\_m3m & 359 & 0.90 & 26.94 & 5.12  \\
& & 20240131\_zf2campirana\_m3m	& 16396 & 0.93 & 36.72 & 6.01  \\
& & All rasters	& 19814 & 0.90 & 36.72 & 6.03  \\
\midrule

\multirow{7}{*}{Ecuador} & \multirow{7}{*}{Agua Salud} 
& 20231018\_inundated\_m3e & 9075 & 0.52 & 54.27 & 6.41 \\
& & 20231018\_pantano\_m3e & 4193 & 0.92 & 41.60 & 6.66 \\
& & 20231018\_terrafirme\_m3e	& 6479 & 0.81 & 53.19 & 6.26 \\
& & 20170810\_transectotoni\_mavicpro & 5119 & 0.83 & 47.97  & 5.80 \\
& & 20230525\_tbslake\_m3e & 1279 & 1.46 & 41.28 & 8.45 \\
& & 20230911\_sanitower\_mini2 & 1721 & 0.86 & 57.16 & 5.53 \\
& & All rasters	& 27866 & 0.52 & 57.16 & 6.31  \\
\midrule

\multirow{5}{*}{Panama} & \multirow{5}{*}{Agua Salud} 
& 20231208\_asforestnorthe2\_m3m & 5925 & 0.51 & 36.17 & 4.99 \\
& & 20231207\_asnortheast\_amsunclouds\_m3m	& 12930 & 0.50 & 36.42 & 4.17 \\
& & 20231207\_asnorthnorth\_pmclouds\_m3m & 6020 & 0.50 & 29.28 & 4.63 \\
& & 20231208\_asforestsouth2\_m3m	& 10582 & 0.83 & 38.92 & 4.83 \\
& & All rasters	& 35457 & 0.50 & 38.92 & 4.58  \\
\midrule

All & All & All rasters	& 83137 & 0.50 & 57.16 & 5.44  \\

\bottomrule
\end{tabular}
} 
\end{center}
\label{table:dataset_stats}
\end{table}

\clearpage
\subsection{Spatially separated splits.}
\label{sec:appendix_spatial_splits}
\begin{figure}[ht]
    \centering
    \includegraphics[width=0.95\linewidth]{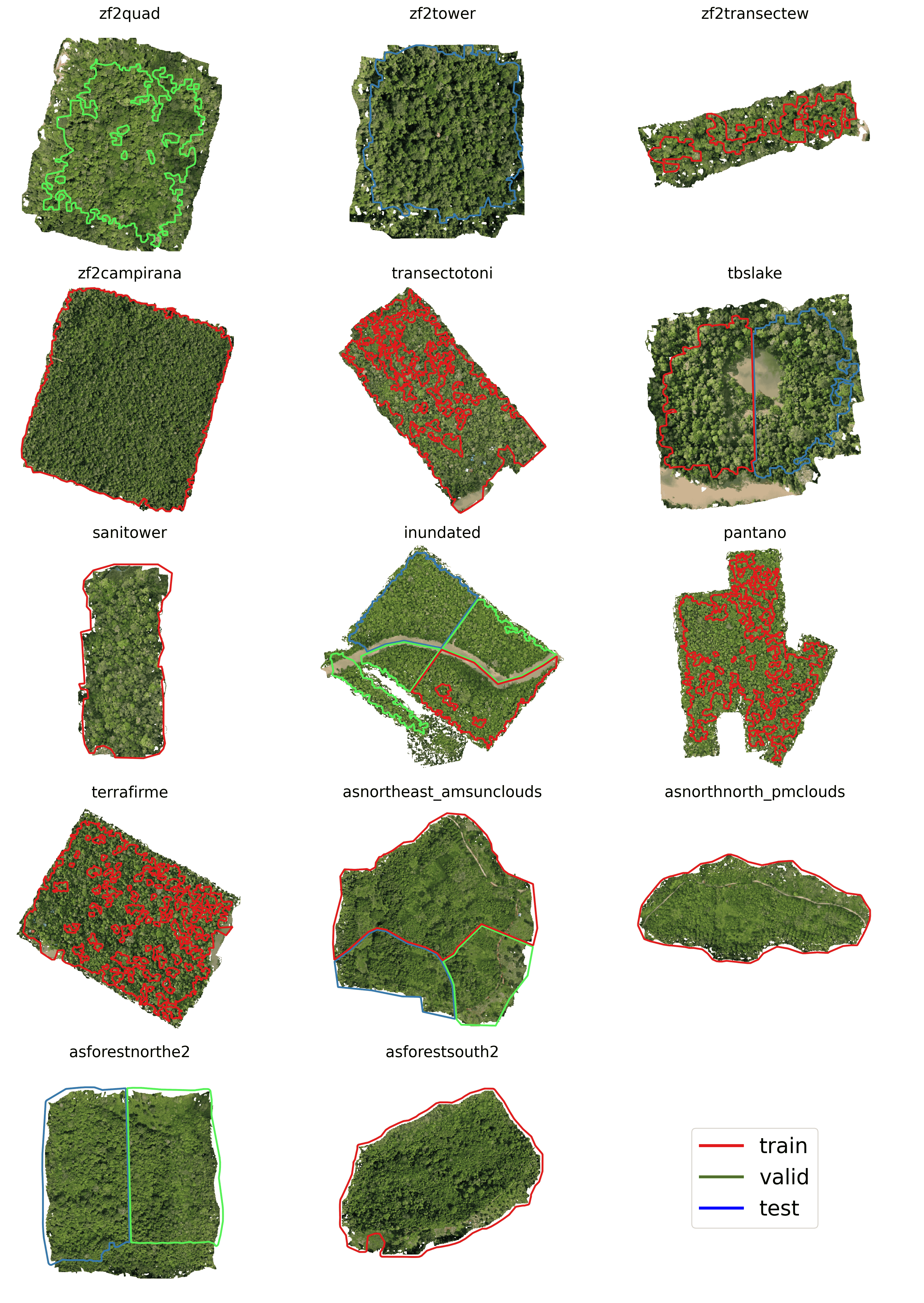}
    \caption{\textbf{Visualization of spatially separated splits.} All 14 rasters of \textsc{SelvaBox} are illustrated with their corresponding train, valid and test AOI-based splits. Images are uniformly sized and not at scale. A few train AOIs (red) have holes to exclude sparse annotations (see Section~\ref{sec:dataset_description}).}
    \label{fig:spatial_splits}
\end{figure}

\clearpage
\subsection{Annotation Protocol}
\label{appendix:annotation_protocol}

The annotations were created by five domain experts with exact same instructions, and all started with a demo and an annotation practice beforehand. All annotations were made in ArcGIS Pro version 3.0 with ArcGIS Online layers to track the online work of two annotators working on the same orthomosaic simultaneously. In large and dense areas, one or several annotators performed an additional pass over the orthomosaic to annotate potential missing trees.

Once annotations were completed by one or several annotators, one or two domain experts performed quality control steps for all annotations of each orthomosaic by following precise guidelines:

\begin{itemize}
  \item[1.] Set up a 60×60\,m grid over the orthomosaic.
  \item[2.] Proceed to the verification by systematically scanning each cell to avoid missing any areas.
  \item[3.] Ensure that there are as many annotated trees as possible in each cell.
  \item[4.] Also annotate dead or leafless trees.
  \item[5.] Check that annotations already completed are correct, adjusting them if necessary.
\end{itemize}

All annotators and reviewers were provided with documentation covering difficult use cases as a reference when they were uncertain about the annotation procedure. For example, they were asked to discuss difficult cases with each other and reach consensus, particularly for ambiguous situations such as intertwined crowns, branches adjacent to large crowns that may correspond to separate understory trees, or vegetation that could be lianas rather than individual trees. Some variability in bounding box tightness (slightly more or less padding around crowns) is also expected. However, none of these concerns were flagged as significant during our systematic quality control.

As a comparison, we point out that annotations in OAM-TCD~\citep{veitch-michaelis_oam-tcd_2024} (NeurIPS 2024) were created by professional annotators who were not domain experts, and only a portion of these annotations were reviewed by ecologists.

\subsection{Incomplete annotations.}
\label{sec:appendix_incomplete_annotations}
\begin{figure}[ht]
    \centering
    \includegraphics[width=1.0\linewidth]{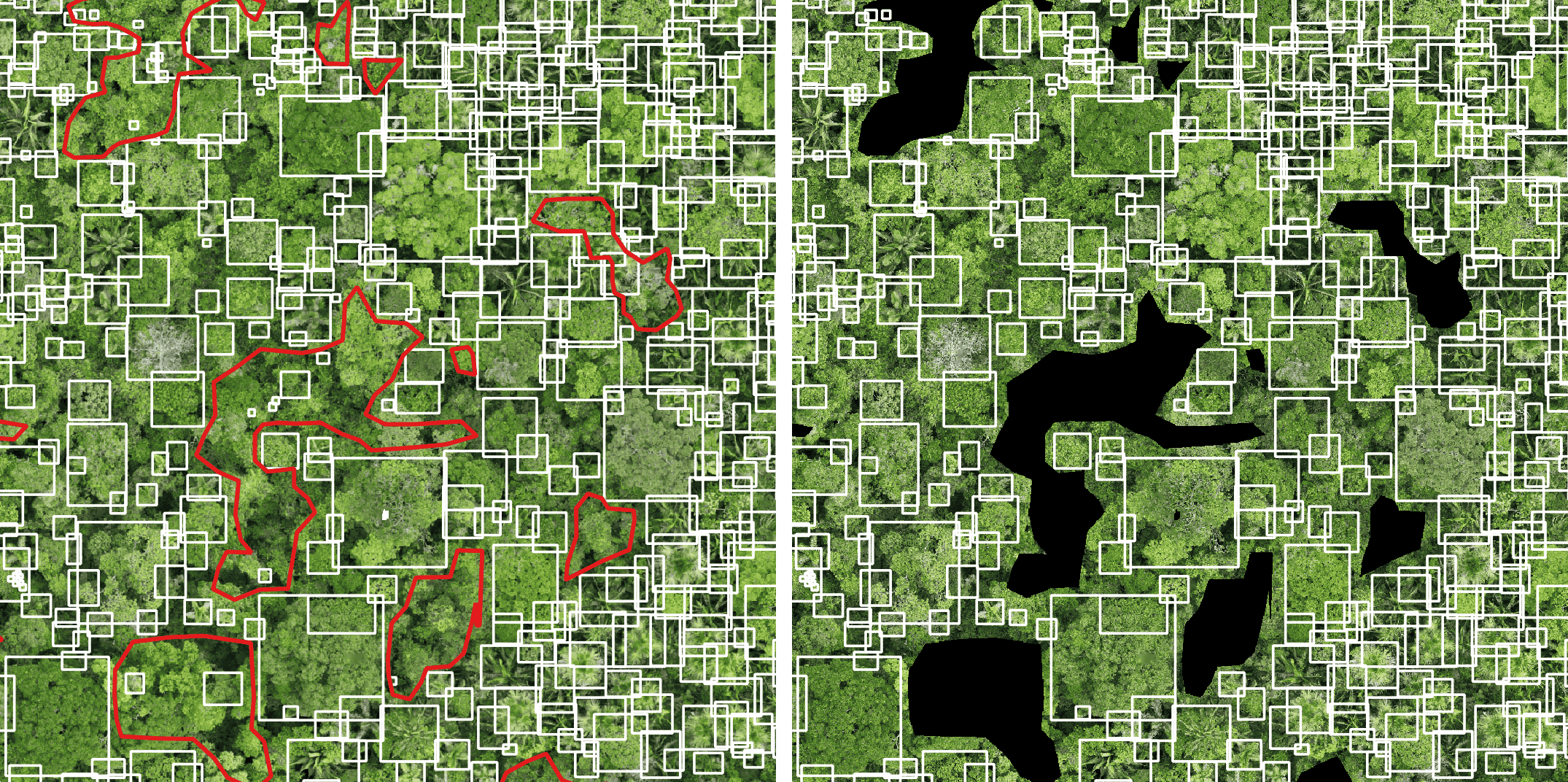}
    \caption{\textbf{Example of masked pixels in sparse annotations zones.} Example on a  $3555\times3555$ pixels training tile ($160\times160$ meters) from the \textit{pantano} raster. On the left is the raw tile, showing holes (red polygons) in the train AOI geopackage where annotations (white boxes) are sparse. On the right is the preprocessed tile, where pixels overlapping the AOI holes have been masked to remove sparse annotations. AOI holes were created mostly where visible trees were not annotated (see Section~\ref{sec:dataset_description}).}
    \label{fig:incomplete_annotations}
\end{figure}

\clearpage
\section{Hyperparameters and augmentations}

\subsection{Augmentations\label{sec:appendix_basic_augmentations}}
For all experiments, we use the same set of basic augmentations:

\begin{table}[h]
\centering
\caption{\textbf{Settings of data augmentations used for all experiments.} Augmentations were applied in the top to bottom order of the table. The Hue augmentation is applied to pixel values in the 0–255 range. The fallback value column describes the behavior of the preprocessing pipeline when an augmentation is not applied. Multi-dataset models use the multi-res. variants of crop and resize augmentations. The `spatial-extent' for our single-res. experiments on \textsc{SelvaBox} is either 40\,m or 80\,m (see Tab.~\ref{table:benchmark_40m} and \ref{table:benchmark_80m}). The crop augmentation for the multi-res. settings is expressed in pixels, where the value is randomly drawn between $x_{\text{min}}$ and $x_{\text{max}}$ that will correspond to different spatial extents depending on the dataset (see Fig.~\ref{fig:appendix_multires_example} and Sec.~\ref{sec:appendix_external_dataset_preprocessing}). The resize augmentation will either be applied with a fixed value $y$, expressed in pixel, for the single-res. applications on \textsc{SelvaBox}, or randomly drawn between $y_{\text{min}}$ and $y_{\text{max}}$ for the multi-resolution and multi-dataset training approaches.}
\begin{tabular}{c c c c}
\toprule
Augmentation & Probability & Augmentation Range & Fallback value\\
\midrule
Flip Horizontal & 0.5 & --- & --- \\
Flip Vertical & 0.5 & --- & --- \\
Rotation & 0.5 & [-$30^\circ$, +$30^\circ$] & --- \\
Brightness & 0.5 & [-20\%, +20\%] & --- \\ 
Contrast & 0.5 & [-20\%, +20\%] & --- \\
Saturation & 0.5 & [-20\%, +20\%] & --- \\
Hue & 0.3 & [-10, +10] & --- \\
Crop (single-res.) & 0.5 & spatial extent $\times$ [-10\%, +10\%] & spatial extent \\
Crop (multi-res.) & 0.5 & [$x_{\text{min}}$, $x_{\text{max}}$] & max. image size \\
Resize (single-res.) & 1.0 & $y$ & --- \\
Resize (multi-res.) & 1.0 & [$y_{\text{min}}$, $y_{\text{max}}$] & ---\\
\bottomrule
\end{tabular}
\label{table:hyperparams}
\end{table}

\clearpage
\subsection{Training hyperparameters\label{sec:appendix_training_hyperparameters}}

This section lists the hyperparameters found for each of our settings. We performed grid search ($\approx 10$ hyperparameter combinations) for every setting on four hyperparameters -- the learning-rate, its scheduler, the total number of epochs and the batch size. We left all other hyperparameters at their default values as specified in Detectron2 and Detrex configuration files. CosineLR refers to a cosine learning-rate schedule without restart. We applied a 5 000-step warmup at the start of each training session. Training was performed on either 48 GB NVIDIA RTX 8000 or L40S GPUs, depending on compute-cluster availability. Most sessions used one or two GPUs; however, DINO + Swin L-384 with large input sizes, multi-resolution, or multi-dataset settings required four GPUs (one image per GPU per batch) due to their high memory footprint.

\begin{table}[h]
\centering
\caption{\textbf{Hyperparameters selected for the input size and GSD experimental analyses on \textsc{SelvaBox}.} Hyperparameters selected for each method and spatial extent in Tables~\ref{table:benchmark_40m} and \ref{table:benchmark_80m}. An initial search shown that, for each architecture and spatial extent, the optimal hyperparameters were nearly identical across GSDs; accordingly, we applied the same settings to all GSDs within each spatial extent.}
\resizebox{.99\linewidth}{!}{
\begin{tabular}{c | c | c c c c c}
\toprule
Method & Extent (m) & Optimizer & LR & Scheduler & Max Epochs & Batch Size \\
\midrule
Faster R-CNN (ResNet50) & $40\times40$ & SGD  & $5\times10^{-3}$ & CosineLR & 500 & 8 \\
DINO 4-scale (ResNet50) & $40\times40$ & AdamW  & $1\times10^{-4}$ & CosineLR & 200 & 4 \\
DINO 5-scale (Swin L-384) & $40\times40$ & AdamW & $5\times10^{-5}$ & CosineLR & 500 & 8 \\
\midrule
Faster R-CNN (ResNet50) & $80\times80$ & SGD & $5\times10^{-3}$ & CosineLR & 500 & 4 \\
DINO 4-scale (ResNet50) & $80\times80$ & AdamW & $1\times10^{-4}$ & CosineLR & 500 & 4 \\
DINO 5-scale (Swin L-384) & $80\times80$ & AdamW & $1\times10^{-4}$ & CosineLR & 500 & 4 \\
\bottomrule
\end{tabular}
}
\label{table:hyperparams_res_extent}
\end{table}

\begin{table}[h]
\centering
\caption{\textbf{Hyperparameters selected for the multi-resolution experimental analysis on \textsc{SelvaBox}.} These hyperparameters were optimal as being the same ones as used for DINO 5-scale (Swin L-384) at $80\times80$ m spatial extent. The associated models performance are in Figures~\ref{fig:multi-res-rf1}, \ref{fig:appendix_multires_results_map_mar} and Table~\ref{table:multi_res}.}
\resizebox{.99\linewidth}{!}{
\begin{tabular}{c | c | c c c c c}
\toprule
Method & Train Crop Range (m) & Optimizer & LR & Scheduler & Max Epochs & Batch Size \\
\midrule
DINO 5-scale (Swin L-384) & $[36, 88]$ & AdamW & $1\times10^{-4}$ & CosineLR & 500 & 4 \\
DINO 5-scale (Swin L-384) & $[30, 100]$ & AdamW & $1\times10^{-4}$ & CosineLR & 500 & 4 \\
DINO 5-scale (Swin L-384) & $[30, 120]$ & AdamW & $1\times10^{-4}$ & CosineLR & 500 & 4 \\
\bottomrule
\end{tabular}
}
\label{table:hyperparams_multires}
\end{table}

\begin{table}[h]
\centering
\caption{\textbf{Hyperparameters selected for the OOD experimental analyses with multi-dataset trainings.} For the MultiStepLR scheduler, we reduced the learning rate by a factor of 10 at $80\%$ and again at $90\%$ of the total training epochs. The associated models performance are in Tables~\ref{tab:results_tropical} and \ref{tab:results_temperate}.}
\resizebox{.99\linewidth}{!}{
\begin{tabular}{c | c | c c c c c c c}
\toprule
Method & Train Datasets & Optimizer & LR & Scheduler & Max Epochs & Batch Size \\
\midrule
DeepForest & N & N/A & N/A & N/A & N/A & N/A \\
Faster R-CNN (ResNet50) & N & SGD & $5\times10^{-3}$ & CosineLR & 500 & 8 \\
DINO 5-scale (Swin L-384) & N & AdamW & $1\times10^{-4}$ & CosineLR & 80 & 4 \\
\midrule
DeepForest  & S & SGD & $5\times10^{-3}$ & CosineLR & 500 & 8\\
Faster R-CNN (ResNet50) & S & SGD & $5\times10^{-3}$ & CosineLR & 500 & 8\\
DINO 5-scale (Swin L-384) & S & AdamW & $1\times10^{-4}$ & CosineLR & 500 & 4 \\
\midrule
DeepForest  & N+Q+O & SGD & $2\times10^{-3}$ & CosineLR & 200 & 4\\
Faster R-CNN (ResNet50) & N+Q+O & SGD & $5\times10^{-3}$ & CosineLR & 200 & 4\\
DINO 5-scale (Swin L-384) & N+Q+O & AdamW & $1\times10^{-4}$ & CosineLR & 80 & 4 \\
\midrule
DeepForest  & N+Q+O+S & SGD & $5\times10^{-3}$ & CosineLR & 120 & 8\\
Faster R-CNN (ResNet50) & N+Q+O+S & SGD & $5\times10^{-3}$ & CosineLR & 120 & 8\\
DINO 5-scale (Swin L-384) & N+Q+O+S & AdamW & $1\times10^{-4}$ & MultiStepLR & 80 & 4 \\
\bottomrule
\end{tabular}
}
\label{table:hyperparams_ood}
\end{table}

\clearpage
\subsection{Inference hyperparameters\label{sec:appendix_nms_hyperparameters}}

We detail the pseudocode for the RF1$_{75}$ metric in Algorithm~\ref{alg:dataset_eval} (see Section~\ref{sec:benchmarking_methods}). Setting $\tau_\text{iou}=0.75$ corresponds to RF1$_{75}$. Before applying the NMS, we discard predictions whose bounding box lies within a 5\%–wide band along the tiles borders. We perform a grid search on the valid set over the non‐maximum suppression IoU threshold $\tau_{\mathrm{nms}}$ and the minimum detection confidence score $s_{\min}$, each taking values in the discrete set $\{0.00,0.05,0.10,\dots,1.00\}$. We multiprocess the grid search on 12 CPU cores to speed up the process. After finding the optimal $\tau_{\mathrm{nms}}$ and $s_{\min}$ on the best model seed, we apply it on the test set to all model seeds to compute the final RF1$_{75}$ score with standard deviation.

\begin{algorithm}[h]
\caption{Per‐dataset evaluation with weighted RF1}
\label{alg:dataset_eval}
\begin{algorithmic}[1]
\Require Dataset $\mathcal{D}$ of rasters, detector $\mathcal{M}$,  
$\tau_{\mathrm{nms}}$, $s_{\min}$, $\tau_\text{iou}$
\State $\mathcal{R}\gets \emptyset$              \Comment{list of per‐raster F1 scores}
\State $\mathcal{W}\gets \emptyset$              \Comment{list of per‐raster truth counts}
\For{each raster $r \in \mathcal{D}$}
  \State $P \gets \emptyset$                     \Comment{accumulate tile preds}
  \State $G \gets \mathrm{LoadGroundTruth}(r)$   \Comment{load geo‐truth}
  \For{each tile $t$ in $r$}
    \State $p \gets \mathcal{M}.\mathrm{predict}(t)$
    \State $P \gets P \cup p$
  \EndFor
  \State $P_{\mathrm{conf}}\gets \{p\in P : p.\mathrm{score}\ge s_{\min}\}$
  \State $P' \gets \mathrm{NonMaxSuppression}(P_{\mathrm{conf}},\tau_{\mathrm{nms}})$
  \State $(tp,fp,fn) \gets \mathrm{GreedyMatch}(P',G,\tau_\text{iou})$
  \State $\mathrm{precision}\gets tp/(tp+fp)$
  \State $\mathrm{recall}\gets tp/(tp+fn)$
  \State $\mathrm{f1}\gets 2\,\frac{\mathrm{precision}\,\mathrm{recall}}%
                                 {\mathrm{precision}+\mathrm{recall}}$
  \State $n \gets |G|$                          \Comment{truth count}
  \State $\mathcal{R} \gets \mathcal{R} \cup \mathrm{f1}$
  \State $\mathcal{W} \gets \mathcal{W} \cup n$
\EndFor
\State $W \gets \sum_{n\in\mathcal{W}} n$
\State $\mathrm{RF1} \gets
        \tfrac{1}{W}\sum_{i=1}^{|\mathcal{R}|}\mathcal{R}_i\cdot\mathcal{W}_i$
\State \textbf{store} weighted‐average RF1
\end{algorithmic}
\end{algorithm}

\begin{algorithm}[h]
\caption{Greedy matching for RF1}
\label{alg:greedy_match}
\begin{algorithmic}[1]
\Procedure{GreedyMatch}{$P',G,\tau_\text{iou}$}
  \State sort $P'$ by descending score
  \State mark all $g\in G$ as \texttt{unmatched}
  \State $tp \gets 0,\quad fp \gets 0$
  \For{each prediction $p\in P'$}
    \State $g^* \gets \arg\max_{g\in G\,:\,g.\text{unmatched}=\text{true}}\mathrm{IoU}(p,g)$
    \If{$\mathrm{IoU}(p,g^*) \ge \tau_\text{iou}$}
      \State $tp \gets tp + 1$
      \State mark $g^*$ as \texttt{matched}
    \Else
      \State $fp \gets fp + 1$
    \EndIf
  \EndFor
  \State $fn \gets \bigl|\{g\in G : g.\text{unmatched}=\text{true}\}\bigr|$
  \State \Return $(tp,fp,fn)$
\EndProcedure
\end{algorithmic}
\end{algorithm}

\begin{table}[h]
\centering
\caption{\textbf{Optimal inference hyperparameters for the input size and GSD experimental analysis at $40\times40$ meters on \textsc{SelvaBox}.} Both optimal NMS and score thresholds are selected by maximizing the $\text{RF1}_{75}$ metric as described in Algorithm~\ref{alg:dataset_eval}. The associated models performance are in Table~\ref{table:benchmark_40m}.}
\resizebox{.65\linewidth}{!}{
\begin{tabular}{c | c c | c c}
\toprule
Method & GSD & I.\ size & NMS IoU ($\tau_{\mathrm{nms}}$) & Score thr. ($s_{\min}$) \\
\midrule
\multirow{6}{*}{\multirowcell{2}{Faster RCNN \\ ResNet50}}
& 10 & 400 & 0.50 & 0.85\\
& 10 & 666 & 0.60 & 0.70\\
& 10 & 888 & 0.50 & 0.80\\
& 6 & 666 & 0.55 & 0.90\\
& 6 & 888 & 0.70 & 0.90\\
& 4.5 & 888 & 0.65 & 0.85\\
\midrule
\multirow{6}{*}{\multirowcell{2}{DINO 4-scale \\ ResNet50}} 
& 10 & 400 & 0.70 & 0.45\\
& 10 & 666 & 0.50 & 0.35\\
& 10 & 888 & 0.75 & 0.35\\
& 6 & 666 & 0.65 & 0.45\\
& 6 & 888 & 0.35 & 0.35\\
& 4.5 & 888 & 0.65 & 0.40\\
\midrule
\multirow{6}{*}{\multirowcell{2}{DINO 5-scale \\ Swin L-384}} 
& 10 & 400 & 0.75 & 0.35\\
& 10 & 666 & 0.80 & 0.45\\
& 10 & 888 & 0.35 & 0.35\\
& 6 & 666 & 0.55 & 0.35\\
& 6 & 888 & 0.45 & 0.40\\
& 4.5 & 888 & 0.50 & 0.35\\
\bottomrule
\end{tabular}
}
\label{table:hyperparams_nms_40m}
\end{table}

\begin{table}[h]
\centering
\caption{\textbf{Optimal inference hyperparameters for the input size and GSD experimental analysis at $80\times80$ meters on \textsc{SelvaBox}.} Both optimal NMS and score thresholds are selected by maximizing the $\text{RF1}_{75}$ metric on the validation set of \textsc{SelvaBox} as described in Algorithm~\ref{alg:dataset_eval}. The associated models performance are in Table~\ref{table:benchmark_80m}.}
\resizebox{.65\linewidth}{!}{
\begin{tabular}{c | c c | c c}
\toprule
Method & GSD & I.\ size & NMS IoU ($\tau_{\mathrm{nms}}$) & Score thr. ($s_{\min}$) \\
\midrule
\multirow{6}{*}{\multirowcell{2}{Faster RCNN \\ ResNet50}}
& 10 & 800 & 0.70 & 0.75\\
& 10 & 1333 & 0.40 & 0.70\\
& 10 & 1777 & 0.35 & 0.60\\
& 6 & 1333 & 0.40 & 0.70\\
& 6 & 1777 & 0.45 & 0.75\\
& 4.5 & 1777 & 0.25 & 0.35\\
\midrule
\multirow{6}{*}{\multirowcell{2}{DINO 4-scale \\ ResNet50}} 
& 10 & 800 & 0.35 & 0.45\\
& 10 & 1333 & 0.75 & 0.45\\
& 10 & 1777 & 0.70 & 0.40\\
& 6 & 1333 & 0.35 & 0.40\\
& 6 & 1777 & 0.75 & 0.35\\
& 4.5 & 1777 & 0.40 & 0.35\\
\midrule
\multirow{6}{*}{\multirowcell{2}{DINO 5-scale \\ Swin L-384}} 
& 10 & 800 & 0.75 & 0.35\\
& 10 & 1333 & 0.80 & 0.40\\
& 10 & 1777 & 0.70 & 0.35\\
& 6 & 1333 & 0.75 & 0.45\\
& 6 & 1777 & 0.65 & 0.35\\
& 4.5 & 1777 & 0.75 & 0.45\\
\bottomrule
\end{tabular}
}
\label{table:hyperparams_nms_80m}
\end{table}

\begin{table}[t!]
\centering
\caption{\textbf{Optimal inference hyperparameters for the multi-resolution experimental analysis on \textsc{SelvaBox}.} Both optimal NMS and score thresholds are selected by maximizing the $\text{RF1}_{75}$ metric on the validation set of \textsc{SelvaBox} as described in Algorithm~\ref{alg:dataset_eval}. The associated models performance are in Figures~\ref{fig:multi-res-rf1}, \ref{fig:appendix_multires_results_map_mar} and Table~\ref{table:multi_res}.}
\resizebox{.80\linewidth}{!}{
\begin{tabular}{c c | c | c c}
\toprule
Method & Train Crop Range (m) & Test GSD (cm) & NMS IoU ($\tau_{\mathrm{nms}}$) & Score thr. ($s_{\min}$) \\
\midrule
\multirow{3}{*}[1ex]{\multirowcell{1}{DINO 5-scale \\ Swin L-384}} 
& \multirow{3}{*}{\multirowcell{1}{$[36, 88]$}}  
& 10 & 0.70 & 0.45\\
& & 6 & 0.60 & 0.45\\
& & 4.5 & 0.70 & 0.45\\
\midrule
\multirow{3}{*}[1ex]{\multirowcell{1}{DINO 5-scale \\ Swin L-384}}   
& \multirow{3}{*}{\multirowcell{1}{$[30, 100]$}} 
& 10 & 0.70 & 0.40\\
& & 6 & 0.70 & 0.40\\
& & 4.5 & 0.60 & 0.40\\
\midrule
\multirow{3}{*}[1ex]{\multirowcell{1}{DINO 5-scale \\ Swin L-384}}  
& \multirow{3}{*}{\multirowcell{1}{$[30, 120]$}} 
& 10 & 0.70 & 0.40\\
& & 6 & 0.50 & 0.35\\
& & 4.5 & 0.80 & 0.40\\
\bottomrule
\end{tabular}
}
\label{table:hyperparams_nms_multires}
\end{table}

\begin{table}[t!]
\centering
\caption{\textbf{Optimal inference hyperparameters for the experimental analyses with multi-dataset trainings.} Both optimal NMS and score thresholds are selected by maximizing the $\text{RF1}_{75}$ metric on the validation sets of both \textsc{SelvaBox} and Detectree2 as described in Algorithm~\ref{alg:dataset_eval}. The associated models performance are in Tables~\ref{tab:results_tropical} and \ref{tab:results_temperate}.}
\resizebox{.65\linewidth}{!}{
\begin{tabular}{c c | c c}
\toprule
Method & Train dataset(s) & NMS IoU ($\tau_{\mathrm{nms}}$) & Score thr. ($s_{\min}$) \\
\midrule
    Detectree2-resize
    & D
    & 0.30
    & 0.25
    \\
    Detectree2-flexi
    & D+urban
    & 0.80
    & 0.20
    \\
    
    \midrule
    DeepForest 
    & N
    & 0.80
    & 0.05
    \\
    F. R-CNN-ResNet50
    & N
    & 0.10
    & 0.50
    \\
    DINO-Swin-L
    & N
    & 0.80
    & 0.55
    \\
    
    \midrule
    DeepForest 
    & S
    & 0.30
    & 0.40
    \\
    F. R-CNN-ResNet50
    & S
    & 0.20
    & 0.45
    \\
    DINO-Swin-L
    & S
    & 0.80
    & 0.40
    \\
    \midrule
    DeepForest 
    & N+Q+O
    & 0.70
    & 0.40
    \\
    F. R-CNN-ResNet50
    & N+Q+O
    & 0.50
    & 0.45
    \\
    DINO-Swin-L
    & N+Q+O
    & 0.70
    & 0.40
    \\
    \midrule
    DeepForest 
    & N+Q+O+S
    & 0.30
    & 0.40
    \\
    F. R-CNN-ResNet50
    & N+Q+O+S
    & 0.20
    & 0.50
    \\
    DINO-Swin-L
    & N+Q+O+S
    & 0.70
    & 0.50
    \\
\bottomrule
\end{tabular}
}
\label{table:hyperparams_nms_ood}
\end{table}

\clearpage
\section{Benchmarking resolutions and image sizes\label{sec:appendix_resolution_imgsize}}

\begin{table}[ht]
\centering
\caption{\textbf{Model, resolution and spatial extent selection on \textsc{SelvaBox} at $40\times40$\,m.} Comparison of performances on the proposed test set of \textsc{SelvaBox} with variable tile spatial extent. Tile size and ground spatial distance (GSD) are in cm. We highlight results per method and backbone as \darksquare the first, \middlesquare the second and \lightsquare the third best scores. We also \textbf{bold} and \underline{underline} the best and second best scores overall. Note that mAP$_{50}$, mAP$_{50:95}$, mAR$_{50}$ and mAR$_{50:95}$ cannot be compared between $40\times40$\,m and $80\times 80$\,m inputs as images do not match, but we can use RF1$_{75}$ to compare final post-aggregation results at the raster-level.}
\resizebox{.9\linewidth}{!}{
\begin{tabular}{c | c c | c c c c c}
\toprule

Method & GSD & I. size & mAP$_{50}$ & mAP$_{50:95}$ & mAR$_{50}$ & mAR$_{50:95}$ & RF1$_{75}$ \\
\midrule
\multirow{6}{*}{\multirowcell{2}{Faster RCNN \\ ResNet50}}
& 10 & 400 & 54.92 \scriptsize ($\pm0.08$) & 26.90 \scriptsize ($\pm0.13$) & 74.48 \scriptsize ($\pm0.42$) & 40.87 \scriptsize ($\pm0.35$) & 35.78 \scriptsize ($\pm0.44$) \\
& 10 & 666 & 57.03 \scriptsize ($\pm0.08$) & 28.40 \scriptsize ($\pm0.13$) & \lightgreen 76.53 \scriptsize ($\pm0.49$) & 42.79 \scriptsize ($\pm0.19$) & \lightgreen 37.75 \scriptsize ($\pm0.30$) \\
& 10 & 888 & 56.42 \scriptsize ($\pm0.30$) & 28.51 \scriptsize ($\pm0.20$) & 76.21 \scriptsize ($\pm0.14$) & 43.36 \scriptsize ($\pm0.19$) & 37.46 \scriptsize ($\pm0.91$) \\
& 6 & 666 & \lightgreen 57.13 \scriptsize ($\pm0.17$) & \lightgreen 29.31 \scriptsize ($\pm0.05$) & 76.25 \scriptsize ($\pm0.66$) & \lightgreen 43.59 \scriptsize ($\pm0.20$) & \darkgreen 39.97 \scriptsize ($\pm0.33$) \\
& 6 & 888 & \middlegreen 57.27 \scriptsize ($\pm0.54$) & \middlegreen 29.40 \scriptsize ($\pm0.34$) & \middlegreen 77.26 \scriptsize ($\pm0.77$) & \middlegreen 44.18 \scriptsize ($\pm0.44$) & \middlegreen 38.92 \scriptsize ($\pm0.51$) \\
& 4.5 & 888 & \darkgreen 58.33 \scriptsize ($\pm0.21$) & \darkgreen 30.25 \scriptsize ($\pm0.24$) & \darkgreen 78.41 \scriptsize ($\pm0.15$) & \darkgreen 45.18 \scriptsize ($\pm0.30$) & \darkgreen 39.97 \scriptsize ($\pm0.67$) \\
\midrule
 
\multirow{6}{*}{\multirowcell{2}{DINO 4-scale \\ ResNet50}}
& 10 & 400 & 56.98 \scriptsize ($\pm0.25$) & 30.63 \scriptsize ($\pm0.24$) & 76.92 \scriptsize ($\pm0.74$) & 48.06 \scriptsize ($\pm0.33$) & 41.14 \scriptsize ($\pm0.80$) \\
& 10 & 666 & 57.62 \scriptsize ($\pm0.64$) & 31.76 \scriptsize ($\pm0.86$) & 78.56 \scriptsize ($\pm0.16$) & 50.40 \scriptsize ($\pm0.55$) & 41.57 \scriptsize ($\pm1.94$) \\
& 10 & 888 & 58.11 \scriptsize ($\pm0.64$) & 32.19 \scriptsize ($\pm0.33$) & 78.55 \scriptsize ($\pm0.34$) & 50.68 \scriptsize ($\pm0.19$) & 42.47 \scriptsize ($\pm0.97$) \\
& 6 & 666 & \lightgreen 58.71 \scriptsize ($\pm0.34$) & \lightgreen 33.46 \scriptsize ($\pm0.22$) & \lightgreen 78.95 \scriptsize ($\pm0.26$) & \lightgreen 51.80 \scriptsize ($\pm0.31$) & \darkgreen 44.55 \scriptsize ($\pm0.18$) \\
& 6 & 888 & \middlegreen 58.78 \scriptsize ($\pm0.51$) & \middlegreen 33.54 \scriptsize ($\pm0.40$) & \middlegreen 79.16 \scriptsize ($\pm0.02$) & \middlegreen 52.12 \scriptsize ($\pm0.18$) & \lightgreen 43.34 \scriptsize ($\pm0.79$) \\
& 4.5 & 888 & \darkgreen 60.11 \scriptsize ($\pm0.36$) & \darkgreen 34.19 \scriptsize ($\pm0.13$) & \darkgreen 79.87 \scriptsize ($\pm0.15$) & \darkgreen 52.53 \scriptsize ($\pm0.40$) & \middlegreen 44.26 \scriptsize ($\pm0.83$) \\
\midrule

\multirow{6}{*}{\multirowcell{2}{DINO 5-scale \\ Swin L-384}}
& 10 & 400 & 60.44 \scriptsize ($\pm0.32$) & 33.84 \scriptsize ($\pm0.20$) & 79.84 \scriptsize ($\pm0.29$) & 52.02 \scriptsize ($\pm0.25$) & 45.37 \scriptsize ($\pm0.23$) \\
& 10 & 666 & 61.26 \scriptsize ($\pm0.30$) & 34.64 \scriptsize ($\pm0.25$) & 80.77 \scriptsize ($\pm0.17$) & 52.91 \scriptsize ($\pm0.30$) & 46.39 \scriptsize ($\pm0.52$) \\
& 10 & 888 & 61.06 \scriptsize ($\pm0.55$) & 34.92 \scriptsize ($\pm0.34$) & 80.70 \scriptsize ($\pm0.13$) & 53.23 \scriptsize ($\pm0.14$) & 45.22 \scriptsize ($\pm0.70$) \\
& 6 & 666 & \middlegreen \underline{62.91 \scriptsize ($\pm0.46$)} & \middlegreen \underline{37.07 \scriptsize ($\pm0.16$)} & \middlegreen \underline{81.58 \scriptsize ($\pm0.12$)} & \middlegreen \underline{55.18 \scriptsize ($\pm0.22$)} & \middlegreen \underline{48.50 \scriptsize ($\pm0.60$)} \\
& 6 & 888 & \lightgreen 62.45 \scriptsize ($\pm0.17$) & \lightgreen 36.22 \scriptsize ($\pm0.38$) & \lightgreen 81.47 \scriptsize ($\pm0.18$) & \lightgreen 54.55 \scriptsize ($\pm0.43$) & \lightgreen 48.13 \scriptsize ($\pm0.60$) \\
& 4.5 & 888 & \darkgreen \textbf{63.41 \scriptsize ($\pm0.29$)} & \darkgreen \textbf{37.78 \scriptsize ($\pm0.15$)} & \darkgreen \textbf{82.33 \scriptsize ($\pm0.35$)} & \darkgreen \textbf{56.30 \scriptsize ($\pm0.21$)} & \darkgreen \textbf{49.76 \scriptsize ($\pm0.43$)} \\

\bottomrule
\end{tabular}
} 
\label{table:benchmark_resolution_40m}

\end{table}

\begin{table}[h]
\centering
\caption{\textbf{Model, resolution and spatial extent selection on \textsc{SelvaBox} at $80\times80$\,m.} Comparison of performances on the proposed test set of \textsc{SelvaBox} with variable tile spatial extent. Tile size and ground spatial distance (GSD) are in cm. We highlight results per method and backbone as \darksquare the first, \middlesquare the second and \lightsquare the third best scores. We also \textbf{bold} and \underline{underline} the best and second best scores overall. Note that mAP$_{50}$, mAP$_{50:95}$, mAR$_{50}$ and mAR$_{50:95}$ cannot be compared between $40\times40$\,m and $80\times 80$\,m inputs as images do not match, but we can use RF1$_{75}$ to compare final post-aggregation results at the raster-level.}
\resizebox{.9\linewidth}{!}{
\begin{tabular}{c | c c | c c c c c}
\toprule

Method & GSD & I. size & mAP$_{50}$ & mAP$_{50:95}$ & mAR$_{50}$ & mAR$_{50:95}$ & RF1$_{75}$ \\
\midrule
\multirow{6}{*}{\multirowcell{2}{Faster RCNN \\ ResNet50}}
& 10 & 800 & 50.50 \scriptsize ($\pm0.44$) & 24.94 \scriptsize ($\pm0.34$) & 64.72 \scriptsize ($\pm1.25$) & 35.93 \scriptsize ($\pm0.55$) & 34.66 \scriptsize ($\pm0.97$) \\
& 10 & 1333 & 51.37 \scriptsize ($\pm0.11$) & 26.25 \scriptsize ($\pm0.14$) & 67.57 \scriptsize ($\pm0.63$) & 38.59 \scriptsize ($\pm0.41$) & 36.09 \scriptsize ($\pm0.51$) \\
& 10 & 1777 & \lightgreen 54.20 \scriptsize ($\pm0.55$) & \lightgreen 27.58 \scriptsize ($\pm0.24$) & \lightgreen 70.65 \scriptsize ($\pm1.84$) & \lightgreen 40.21 \scriptsize ($\pm0.38$) & 35.74 \scriptsize ($\pm1.26$) \\
& 6 & 1333 & 51.96 \scriptsize ($\pm0.64$) & 26.52 \scriptsize ($\pm0.80$) & 69.77 \scriptsize ($\pm1.53$) & 39.55 \scriptsize ($\pm0.75$) & \middlegreen 36.22 \scriptsize ($\pm1.45$) \\
& 6 & 1777 & \middlegreen 54.68 \scriptsize ($\pm0.26$) & \middlegreen 27.89 \scriptsize ($\pm0.35$) & \darkgreen 72.32 \scriptsize ($\pm1.35$) & \middlegreen 41.02 \scriptsize ($\pm0.69$) & \lightgreen 35.94 \scriptsize ($\pm0.84$) \\
& 4.5 & 1777 & \darkgreen 56.21 \scriptsize ($\pm0.76$) & \darkgreen 28.74 \scriptsize ($\pm0.44$) & \middlegreen 72.12 \scriptsize ($\pm0.76$) & \darkgreen 41.27 \scriptsize ($\pm0.59$) & \darkgreen 37.52 \scriptsize ($\pm0.58$) \\
\midrule
 
\multirow{6}{*}{\multirowcell{2}{DINO 4-scale \\ ResNet50}}
& 10 & 800 & 58.32 \scriptsize ($\pm0.44$) & 30.90 \scriptsize ($\pm0.51$) & 76.33 \scriptsize ($\pm0.28$) & 47.29 \scriptsize ($\pm0.33$) & 41.20 \scriptsize ($\pm0.39$) \\
& 10 & 1333 & 59.65 \scriptsize ($\pm0.20$) & 32.39 \scriptsize ($\pm0.02$) & 77.61 \scriptsize ($\pm0.07$) & 49.22 \scriptsize ($\pm0.10$) & \lightgreen 43.08 \scriptsize ($\pm0.20$) \\
& 10 & 1777 & 59.31 \scriptsize ($\pm1.29$) & 32.51 \scriptsize ($\pm0.89$) & 77.23 \scriptsize ($\pm0.34$) & 49.35 \scriptsize ($\pm0.47$) & 42.39 \scriptsize ($\pm1.25$) \\
& 6 & 1333 & \lightgreen 59.84 \scriptsize ($\pm0.42$) & \lightgreen 33.06 \scriptsize ($\pm0.29$) & \lightgreen 77.91 \scriptsize ($\pm0.17$) & \lightgreen 49.93 \scriptsize ($\pm0.39$) & 42.92 \scriptsize ($\pm0.51$) \\
& 6 & 1777 & \middlegreen 60.48 \scriptsize ($\pm0.26$) & \middlegreen 33.62 \scriptsize ($\pm0.10$) & \middlegreen 78.32 \scriptsize ($\pm0.21$) & \middlegreen 50.85 \scriptsize ($\pm0.17$) & \darkgreen 44.18 \scriptsize ($\pm0.18$) \\
& 4.5 & 1777 & \darkgreen 61.09 \scriptsize ($\pm0.45$) & \darkgreen 33.81 \scriptsize ($\pm0.84$) & \darkgreen 78.93 \scriptsize ($\pm0.32$) & \darkgreen 51.00 \scriptsize ($\pm0.77$) &  \middlegreen 43.26 \scriptsize ($\pm0.45$) \\
\midrule

\multirow{6}{*}{\multirowcell{2}{DINO 5-scale \\ Swin L-384}}
& 10 & 800 & 62.02 \scriptsize ($\pm0.08$) & 33.90 \scriptsize ($\pm0.09$) & 78.89 \scriptsize ($\pm0.22$) & 50.29 \scriptsize ($\pm0.38$) & 44.64 \scriptsize ($\pm0.20$) \\
& 10 & 1333 & 61.73 \scriptsize ($\pm0.72$) & 34.22 \scriptsize ($\pm0.34$) & 79.03 \scriptsize ($\pm0.87$) & 50.76 \scriptsize ($\pm0.57$) & 45.64 \scriptsize ($\pm1.03$) \\
& 10 & 1777 & 62.86 \scriptsize ($\pm0.78$) & 35.30 \scriptsize ($\pm0.26$) & 79.94 \scriptsize ($\pm0.68$) & 52.12 \scriptsize ($\pm0.62$) & 45.37 \scriptsize ($\pm0.08$) \\
& 6 & 1333 & \darkgreen \textbf{64.91 \scriptsize ($\pm0.30$)} & \middlegreen \underline{37.12 \scriptsize ($\pm0.38$)} & \middlegreen \underline{81.01 \scriptsize ($\pm0.09$)} & \middlegreen \underline{53.56 \scriptsize ($\pm0.48$)} & \middlegreen \underline{47.81 \scriptsize ($\pm0.40$)} \\
& 6 & 1777 & \lightgreen 63.34 \scriptsize ($\pm0.58$) & \lightgreen 35.77 \scriptsize ($\pm0.84$) & \lightgreen 80.59 \scriptsize ($\pm0.16$) & \lightgreen 52.91 \scriptsize ($\pm0.56$) & \lightgreen 45.88 \scriptsize ($\pm1.97$) \\
& 4.5 & 1777 & \middlegreen \underline{64.59 \scriptsize ($\pm1.03$)} & \darkgreen \textbf{37.79 \scriptsize ($\pm0.55$)} & \darkgreen \textbf{81.35 \scriptsize ($\pm0.71$)} & \darkgreen \textbf{54.66 \scriptsize ($\pm0.47$)} & \darkgreen \textbf{49.38 \scriptsize ($\pm0.76$)} \\

\bottomrule
\end{tabular}
} 
\label{table:benchmark_resolution_80m}

\end{table}

\clearpage
\section{Multi-resolution approach}

\subsection{Multi-resolution example \label{sec:appendix_multires_example}}

\begin{figure}[ht]
    \centering
    \includegraphics[width=1.0\linewidth]{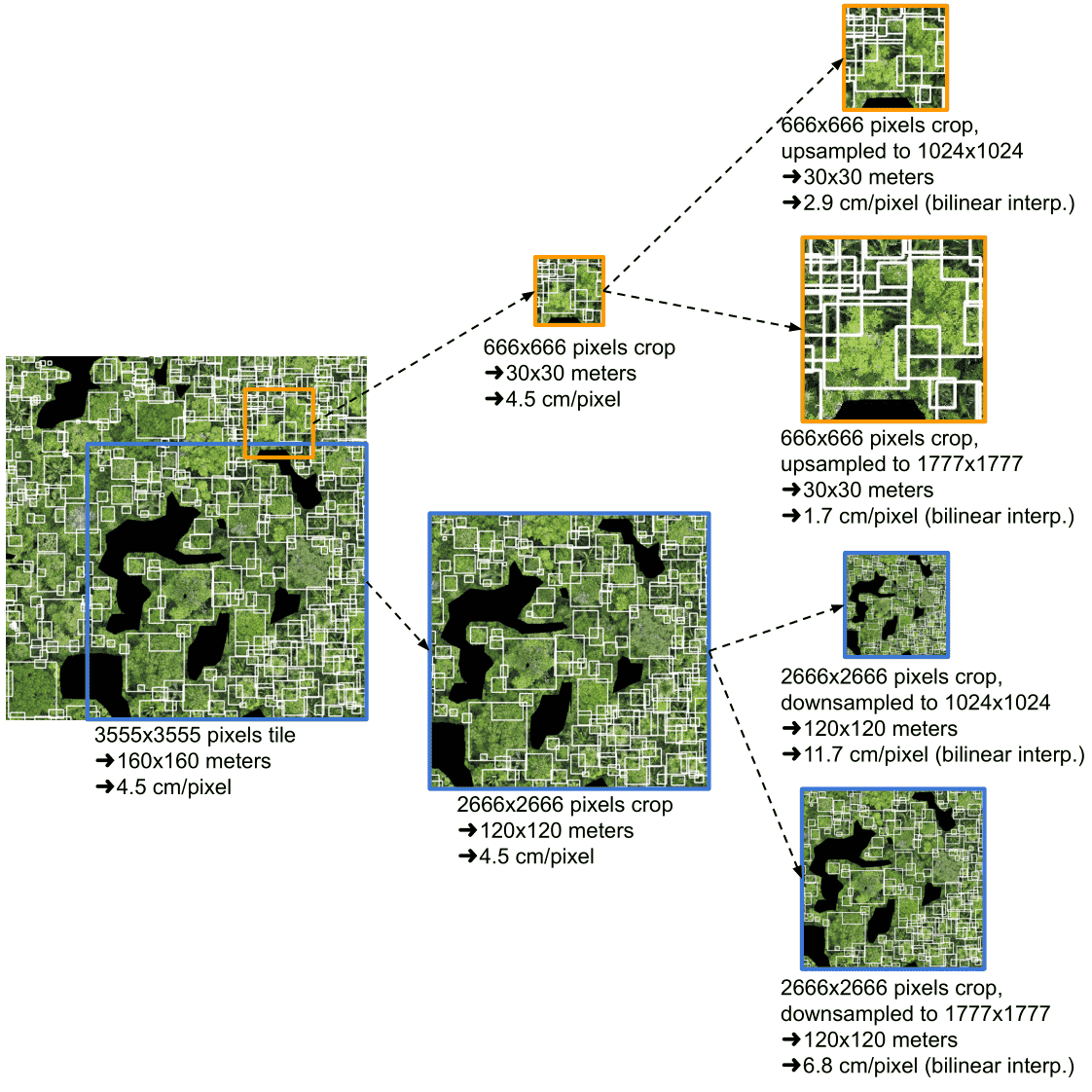}
    \caption{\textbf{Example of cropping and resizing augmentations for the multi-resolution approach.} We showcase the $[30, 120]$\,m configuration used in our benchmark: a $3555 \times 3555$ tile at $4.5\text{cm}=0.045$\,m GSD, equivalent to a $160 \times 160$\,m spatial extent, will be cropped with a random crop size value in $[666, 2666]$ pixels, and then resized to a random value in $[1024, 1777]$ pixels. 
    This process has two effects: \protect\circledgray{1} cropping performs augmentation for spatial extent -- in our example, the original input has the potential to be cropped in a ground extent range of $[30, 120]$\,m; \protect\circledgray{2} resizing performs the GSD augmentation -- in our example, the largest possible crop (in blue) of 2666 pixels (or 120\,m) can be downsampled to $1024 \times 1024$, which yields a maximum effective GSD of $0.045\;\text{m} \times \frac{2666}{1024}=0.117\;\text{m}=11.7\;\text{cm}$ per pixel, far from the original 4.5 cm per pixel. Similarly, the smallest possible crop (in orange) of 666 pixels (or 30\,m) can be upsampled to $1777\times1777$ pixels, yielding a minimum effective GSD of $0.045\;\text{m} \times \frac{666}{1777}=0.017\;\text{m}=1.7\;\text{cm}$ per pixel. Note that for small crops, the effective GSD after upsampling (via bilinear interpolation) can fall below the original 4.5 cm/pixel, even though no new image detail is added.}
    \label{fig:appendix_multires_example}
\end{figure}

\clearpage
\subsection{Multi-resolution additional results\label{sec:appendix_multires_results}}

\begin{figure}[ht]
    \centering
    \includegraphics[width=0.75\linewidth]{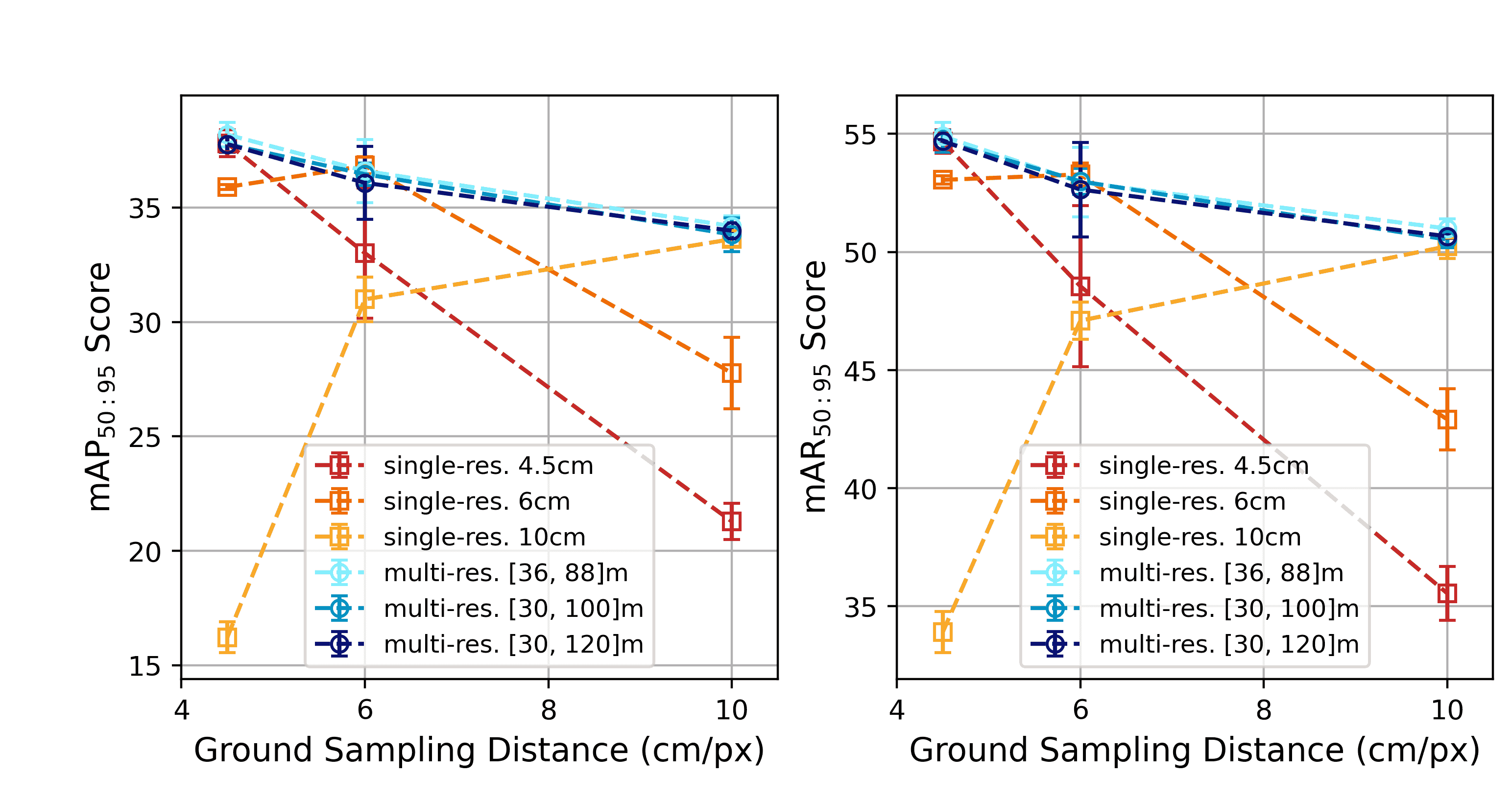}
    \caption{\textbf{Multi-resolution vs. single-resolution on \textsc{SelvaBox}.} Comparison of mAP$_{50:95}$ and mAR$_{50:95}$ between best performing single-resolution methods from Table~\ref{table:benchmark_80m} trained with a fixed spatial extent of $80\times 80$ m, against multi-resolution approaches with increasingly large crop augmentation ranges ($[36, 88]$, $[30, 100]$ and $[30, 120]$). All methods are `DINO 5-scale Swin L-384'. It supports results illustrated in Figure~\ref{fig:multi-res-rf1}.}
    \label{fig:appendix_multires_results_map_mar}
\end{figure}

\begin{table}[ht]
\centering
\caption{\textbf{Multi-resolution vs. single-resolution on \textsc{SelvaBox}.} Comparison of best performing methods from  Table~\ref{table:benchmark_80m} trained with a fixed spatial extent against multi-resolution approaches. All methods are `DINO 5-scale Swin L-384', have been trained at 4.5cm. We mark the best and second-best scores in \textbf{bold} and \underline{underline}, respectively. These results are also illustrated in Figures~\ref{fig:multi-res-rf1} and \ref{fig:appendix_multires_results_map_mar}.}
\resizebox{.99\linewidth}{!}{
\begin{tabular}{c | c c | c c c c c}
\toprule

\makecell{Train\\ extent\\(m)} & \makecell{Test\\ extent\\(m)} & \makecell{Test\\res.\\(cm/px)} & mAP$_{50}$ & mAP$_{50:95}$ & mAR$_{50}$ & mAR$_{50:95}$ & RF1$_{75}$ \\
\midrule
80 & 80 & 10 & 62.02 \scriptsize ($\pm0.08$) & 33.90 \scriptsize ($\pm0.09$) & 78.89 \scriptsize ($\pm0.22$) & 50.29 \scriptsize ($\pm0.38$) & 44.64 \scriptsize ($\pm0.20$) \\
80 & 80 & 6 & 64.91 \scriptsize ($\pm0.30$) & 37.12 \scriptsize ($\pm0.38$) & 81.01 \scriptsize ($\pm0.09$) & 53.56 \scriptsize ($\pm0.48$) & 47.81 \scriptsize ($\pm0.40$) \\
80 & 80 & 4.5 & 64.59 \scriptsize ($\pm1.03$) & \underline{37.79 \scriptsize ($\pm0.55$)} & 81.35 \scriptsize ($\pm0.71$) & 54.66 \scriptsize ($\pm0.47$) & \textbf{49.38 \scriptsize ($\pm0.76$)} \\
\midrule
\multirow{3}{*}{$[36, 88]\cup\{160\}$} & 80 & 10 & 63.33 \scriptsize ($\pm0.48$) & 34.19 \scriptsize ($\pm0.44$) & 79.98 \scriptsize ($\pm0.21$) & 50.99 \scriptsize ($\pm0.41$) & 45.03 \scriptsize ($\pm0.53$) \\
& 80 & 6 & \underline{65.38 \scriptsize($\pm0.41$)} & 36.60 \scriptsize ($\pm1.38$) & 81.29 \scriptsize ($\pm0.20$) & 52.95 \scriptsize ($\pm1.47$) & 47.87 \scriptsize ($\pm0.92$) \\
& 80 & 4.5 & \textbf{65.68 \scriptsize ($\pm0.09$)} & \textbf{38.19 \scriptsize ($\pm0.54$)} &  \underline{81.85 \scriptsize ($\pm0.05$)} & \textbf{54.90 \scriptsize ($\pm0.59$)} & \underline{49.16 \scriptsize ($\pm0.06$)} \\
\midrule
\multirow{3}{*}{$[30, 100]\cup\{160\}$} & 80 & 10 & 62.52 \scriptsize ($\pm1.30$) & 33.82 \scriptsize ($\pm0.74$) & 79.42 \scriptsize ($\pm0.35$) & 50.52 \scriptsize ($\pm0.35$) & 44.13 \scriptsize ($\pm0.73$) \\
& 80 & 6 & 64.70 \scriptsize ($\pm0.48$) & 36.46 \scriptsize ($\pm0.49$) & 80.99 \scriptsize ($\pm0.12$) & 52.99 \scriptsize ($\pm0.55$) & 47.96 \scriptsize ($\pm0.48$) \\
& 80 & 4.5 & 65.11 \scriptsize ($\pm0.28$) & 37.77 \scriptsize ($\pm0.36$) & 81.47 \scriptsize ($\pm0.15$) & 54.68 \scriptsize ($\pm0.47$) & 48.79 \scriptsize ($\pm0.51$) \\
\midrule
\multirow{3}{*}{$[30, 120]\cup\{160\}$} & 80 & 10 & 62.76 \scriptsize ($\pm0.49$) & 33.99 \scriptsize ($\pm0.35$) &  79.51 \scriptsize ($\pm0.09$) & 50.66 \scriptsize ($\pm0.08$) & 44.91 \scriptsize ($\pm0.65$)\\
& 80 & 6 & 64.44 \scriptsize ($\pm0.26$) & 36.08 \scriptsize ($\pm1.59$) & 80.68 \scriptsize ($\pm0.42$) & 52.64 \scriptsize ($\pm2.00$) & 46.65 \scriptsize ($\pm1.67$)\\
& 80 & 4.5 & 64.92 \scriptsize ($\pm0.53$) & 37.77 \scriptsize ($\pm0.35$) & 81.19 \scriptsize ($\pm0.08$) & \underline{54.69 \scriptsize ($\pm0.07$)} & 48.60 \scriptsize ($\pm0.49$)\\
\bottomrule

\end{tabular}
} 
\label{table:multi_res}
\end{table}

\clearpage
\section{Ablation Study on RF1 IoU threshold\label{section:rf1_iou_ablation}}

\begin{figure}[ht]
    \centering
    \includegraphics[width=0.5\linewidth]{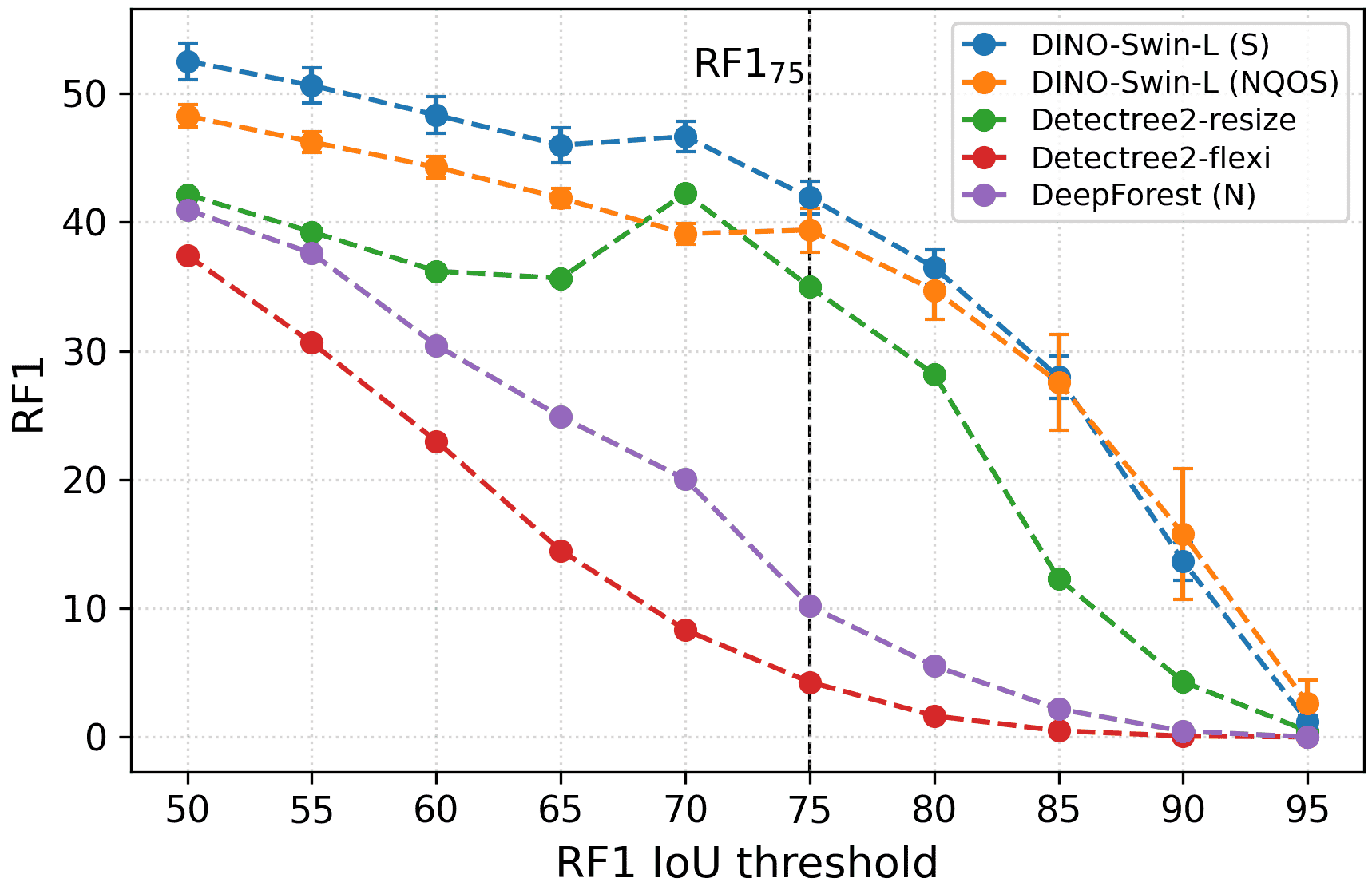}
    \caption{\textbf{RF1 vs IoU threshold on BCI50ha.} Comparison of two of our DINO-Swin-L variants and competing methods at different IoU thresholds. In this work we focus on RF1$_{75}$ (IoU=0.75). For each IoU threshold, NMS hyperparameters are re-optimized on the validation set.}
    \label{fig:rf1_vs_iou_bci50ha_mainpaper}
\end{figure}

\begin{figure}[ht]
    \centering
    \includegraphics[width=0.5\linewidth]{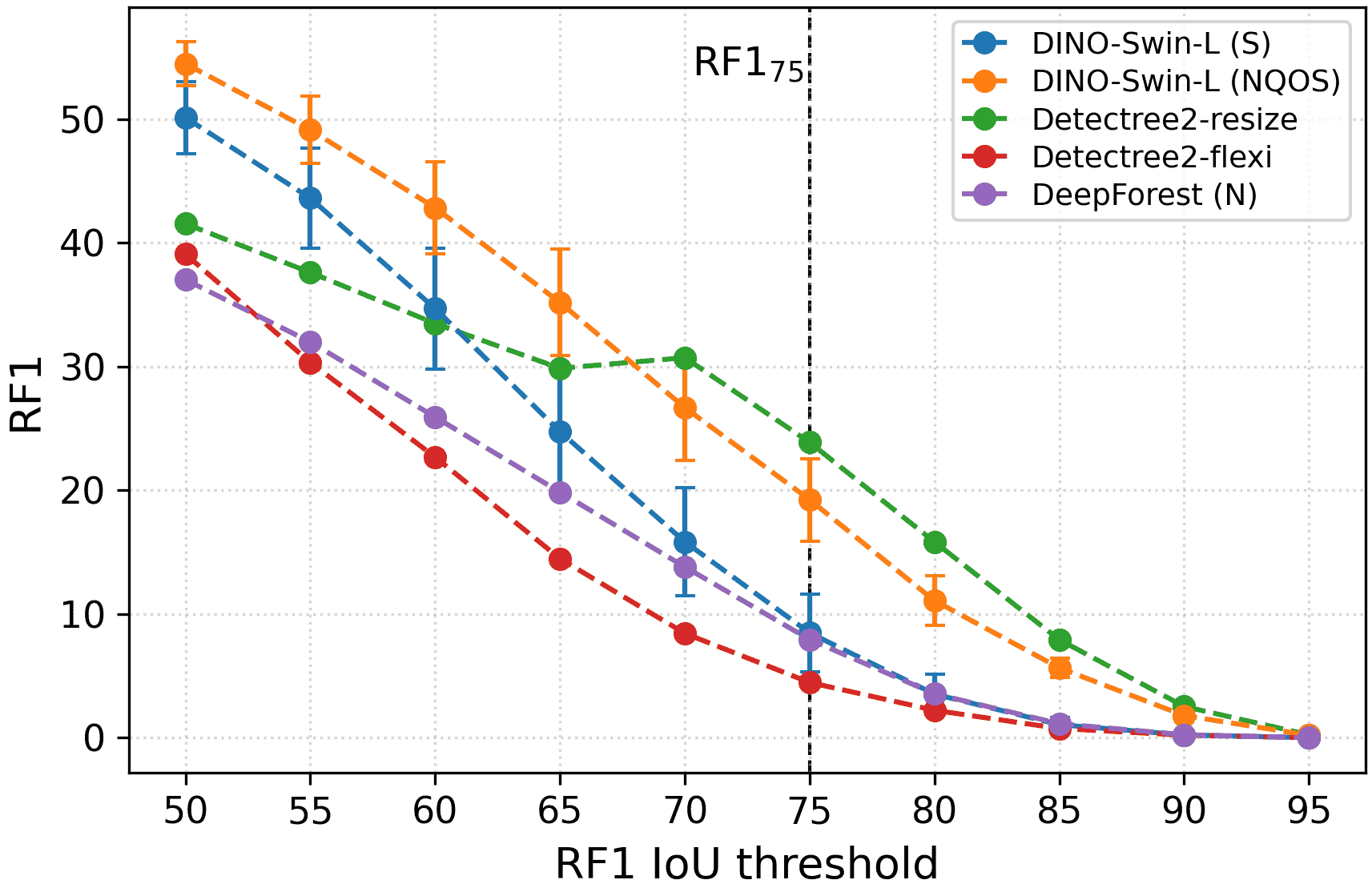}
    \caption{\textbf{RF1 vs IoU threshold on Detectree2.} Comparison of two of our DINO-Swin-L variants and competing methods at different IoU thresholds. In this work we focus on RF1$_{75}$ (IoU=0.75). For each IoU threshold, NMS hyperparameters are re-optimized on the validation set.}
    \label{fig:rf1_vs_iou_detectree2_mainpaper}
\end{figure}

\begin{figure}[ht]
    \centering
    \includegraphics[width=0.5\linewidth]{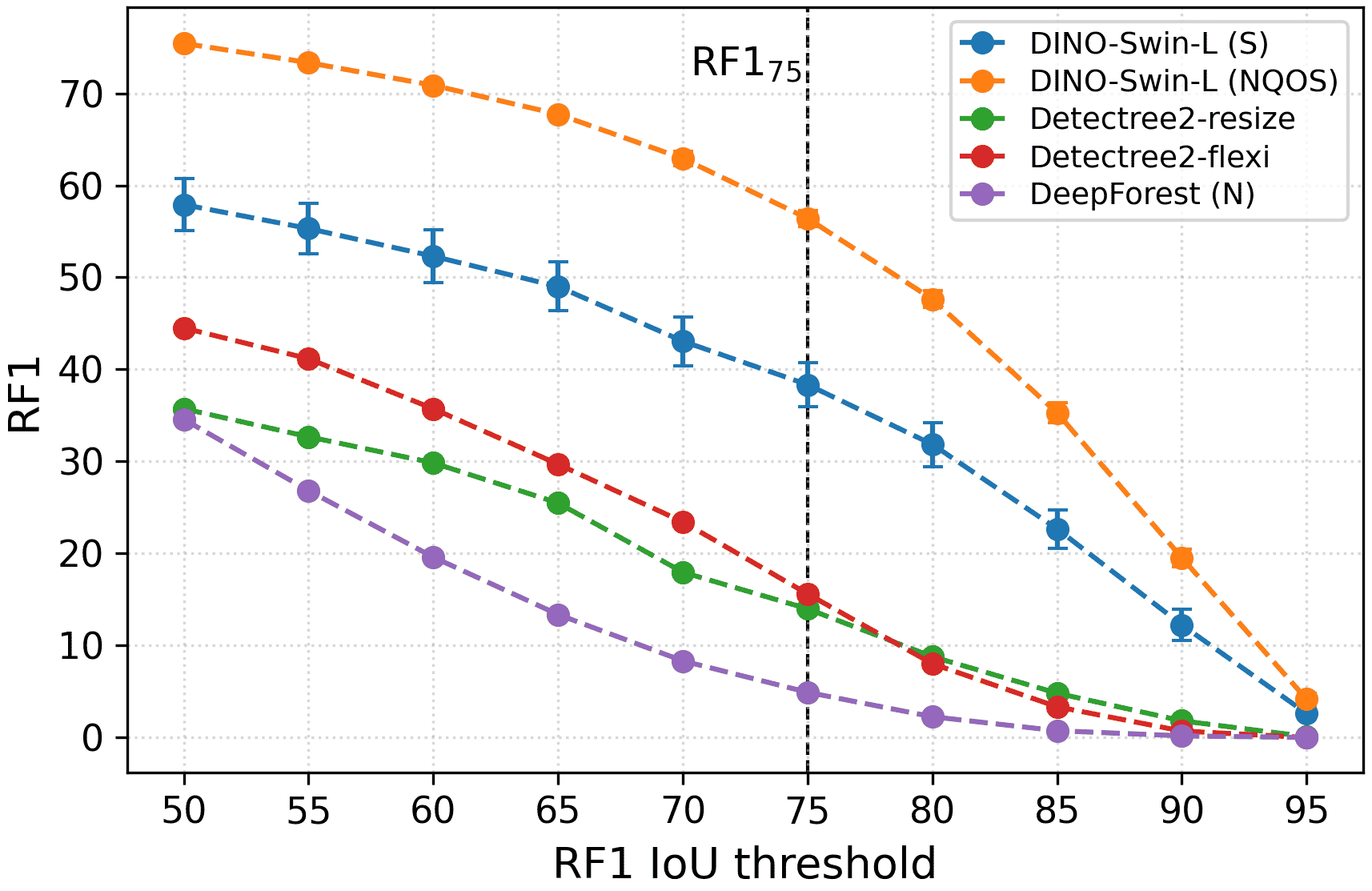}
    \caption{\textbf{RF1 vs IoU threshold on QuebecTrees.} Comparison of two of our DINO-Swin-L variants and competing methods at different IoU thresholds. In this work we focus on RF1$_{75}$ (IoU=0.75). For each IoU threshold, NMS hyperparameters are re-optimized on the validation set.}
    \label{fig:rf1_vs_iou_quebectrees_mainpaper}
\end{figure}

\clearpage
\section{Out-of-distribution analysis}

\subsection{External datasets preprocessing}
\label{sec:appendix_external_dataset_preprocessing}

For NeonTreeEvaluation, we keep the proposed $400 \times 400$ pixels test inputs at 10\,cm GSD and define train and validation AOIs on their rasters. 
Similarly, for QuebecTrees, we keep the proposed test split AOI while defining our own train and validation AOIs. 
As Detectree2's train, validation, and test splits are not shared publicly, we defined our own validation and test AOIs, while keeping the input size as $1000 \times 1000$ to follow their guidelines. 
BCI50ha is only used for OOD evaluation (see OOD experiments in Sections \ref{sec:benchmarking_methods} and \ref{sec:experiments}), so we define test AOIs spanning both rasters.

OAM-TCD contains two types of annotations: individual trees and tree groups.
Unfortunately, tree groups would introduce noise during the training process as all other datasets focus on individual tree detection.
Therefore, we only consider individual trees annotations and we mask the pixels associated to tree groups from the training data to ensure consistency. This process is similar to how we mask specific low quality pixels and sparse annotations in \textsc{SelvaBox} as detailed in Section~\ref{sec:dataset_description}.
OAM-TCD provides five predefined cross-validation folds; we train on folds 0–3 and use fold 4 exclusively for validation.
We further divide the $2048 \times 2048$ validation and test tiles of OAM-TCD into $1024 \times 1024$ tiles with 50\% overlap, as $204.8 \times 204.8$\,m GSD would be significantly larger than other datasets. We refer to Table~\ref{table:preprocessed_datasets} for more details on final preprocessed datasets statistics and information.

For each dataset divided into tiles, we apply the same AOI-based pixel masking, black/white/transparent pixel cover threshold, and 0-annotation tile removal, as described in Section~\ref{sec:dataset_description}. We use 50\% overlap between tiles for all datasets for which we divided rasters into tiles, except BCI50ha where we use 75\% to maximize cover for 50+ meters tree crowns (same as \textsc{SelvaBox} test split).
%
We also release these preprocessed external datasets on HuggingFace, including the proposed AOIs and raster-level annotation geopackages for all datasets, in a standardized ML-ready format and with their original CC-BY 4.0 license to ensure reproducibility of our benchmark and facilitate experiments of researchers and practitioners for tree‑crown detection. We used version 1.0.0 of OAM-TCD~\footnote{OAM-TCD: \url{https://zenodo.org/records/11617167}}, version v1 of QuebecTrees~\footnote{QuebecTrees: \url{https://zenodo.org/records/8148479}}, version v2 of Detectree2~\footnote{Detectree2: \url{https://zenodo.org/records/8136161}}, version 0.2.2 of NeonTreeEvaluation~\footnote{NeonTreeEvaluation: \url{https://zenodo.org/records/5914554}}, and version 2 of BCI50ha~\footnote{BCI50ha: \href{https://smithsonian.figshare.com/articles/dataset/Barro_Colorado_Island_50-ha_plot_crown_maps_manually_segmented_and_instance_segmented_/24784053}{Smithsonian Barro Colorado Island 50-ha plot crown maps}}.

\begin{table}[h]
\centering
\caption{
\textbf{Preprocessing and training parameters for all datasets used.}
The \textsc{SelvaBox} parameters correspond to the $[30, 120]$\,m multi-resolution setting.
Although test tiles outnumber training tiles numerically, training tiles are deliberately larger in spatial extent to facilitate augmentation strategies, resulting in greater total geographic coverage within the train split.
The minimum effective train resolution range is reached by using bilinear interpolation from the smallest possible crop size to the largest possible input resize value. *At training time, we resize NeonTreeEvaluation training tiles to 2000 pixels before
cropping to ensure that the effective train extent range reaches the 40\,m used in the test split.
}
\resizebox{.99\linewidth}{!}{
\begin{tabular}{l | c | c c c c c c | c c c}
\toprule
Dataset
  & \makecell{GSD\\(cm/px)}
  & \makecell{\#\,Train\\Images}
  & \makecell{Train size\\ (px)}
  & \makecell{Augm. Crop\\ range (px)}
  & \makecell{Augm. Resize\\ range (px)}
  & \makecell{Effective train\\ extent range (m)}
  & \makecell{Effective train\\ res. range (cm/px)}
  & \makecell{\#\,Test\\Images}
  & \makecell{Test size\\(px)}
  & \makecell{Test extent\\(m)} \\
\midrule
NeonTreeEvaluation      & 10 & 912 & 1200 & [666, 2666] & [1024, 1777] & [40, 120]$^*$ & [2.3, 11.7] & 194 & 400  & 40 \\
OAM-TCD                 & 10 & 3024 & 2048 & [666, 2666] & [1024, 1777] & [66.6, 204.8] & [3.8, 20] & 2527 & 1024 & 102.4 \\
QuebecTrees             & 3 & 148 & 3333 & [666, 2666] & [1024, 1777] & $[20, 80]\cup\{100\}$ & [1.1, 9.8] & 168 & 1666 & 50 \\
\textsc{SelvaBox}               & 4.5 & 585 & 3555 & [666, 2666] & [1024, 1777] & $[30, 120]\cup\{160\}$ & [1.7, 15.6] & 1477 & 1777 & 80 \\
Detectree2              & 10 & N/A & N/A & N/A & N/A & N/A & N/A & 311 &  1000 & 100 \\
BCI50ha                 & 4.5 & N/A & N/A & N/A & N/A & N/A & N/A & 2706 & 1777 & 80 \\
\bottomrule
\end{tabular}}
\label{table:preprocessed_datasets}
\vspace{-0.5cm}
\end{table}

\begin{figure}[H]
    \centering
    \includegraphics[width=1.0\linewidth]{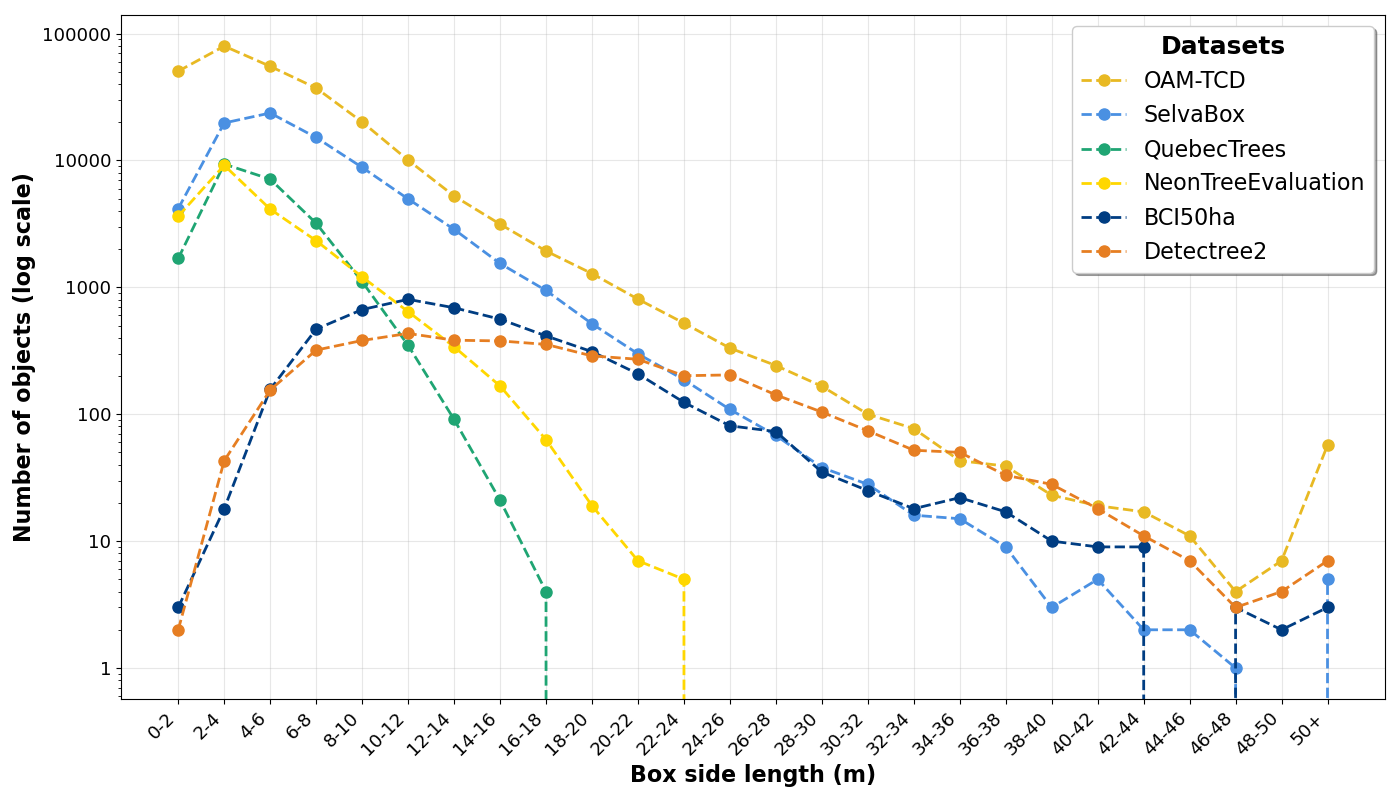}
    \vspace{-0.5cm}
    \caption{\textbf{Distribution of box annotations size across datasets.}}
    \label{fig:appendix_box_size}
\end{figure}

\clearpage

\subsection{Statistical comparison between datasets}

To rigorously compare SelvaBox with existing datasets, we quantified crown-size distribution differences using three complementary statistical approaches:

\paragraph{Methodology} We computed the Jensen-Shannon (JS) distance and Kullback-Leibler (KL) divergence across datasets. Since KL divergence is asymmetric and unbounded, making interpretation difficult, we prioritize JS distance, which is symmetric, bounded between 0 and 1, and well-suited for discrete distributions. We also performed two-sample Kolmogorov-Smirnov (KS) tests comparing SelvaBox against existing datasets. The KS test evaluates the maximum vertical distance between empirical cumulative distribution functions (ECDFs), providing a non-parametric, distribution-free measure robust to differences in dataset size. Given that KS test p-values follow $p \approx 2e^{-2 n \cdot D^2}$ \cite{JSSv008i18}, where $n$ is dataset size and $D$ the KS statistic, and our dataset sizes range from $n = 3,947$ (Detectree2) to $266,663$ (OAM-TCD), we expect p-values approaching $1.04\cdot 10^{-35}$ when $D \ge 0.1$ and  $n \ge 3,947$.

\paragraph{Results} SelvaBox exhibits substantially different crown-size distributions from tropical datasets (JS Distance: BCI50ha = 0.6248, Detectree2 = 0.6476) and moderately different distributions from temperate datasets (NeonTreeEvaluation = 0.2789, QuebecTrees = 0.2622). Pairwise KS tests reveal highly significant differences across all comparisons (p < 0.001, KS statistics ranging from 0.2117 to 0.6338), confirming that SelvaBox's crown-size distribution is statistically distinct from all existing datasets. The large annotation counts (3,947 to 266,663 samples) ensure these distributional differences are meaningful and robust to dataset scale variations.

\paragraph{Interpretation} OAM-TCD shows the greatest similarity to SelvaBox (JS Distance = 0.2152), likely because its large geographic scale and multi-biome coverage encompass diverse crown morphologies, unlike datasets restricted to single regions or biomes. The asymmetric KL divergence values further support these conclusions, demonstrating how SelvaBox uniquely captures the crown-size distributions and structural diversity of tropical forests.

\begin{table}[h]
  \centering
  \caption{\textbf{Statistical comparison of crown-size distributions between \textsc{SelvaBox} and existing datasets.} Higher JS distance, KL divergence, and KS statistics indicate greater distributional differences, while very small KS $p$-values indicate that the null hypothesis of identical distributions can be rejected.}
  \setlength{\tabcolsep}{4pt}
  \resizebox{.99\linewidth}{!}{
  \begin{tabular}{l|l|ccccc}
    \toprule
    && BCI50ha & Detectree2 & NeonTreeEval. & QuebecTrees & OAM-TCD \\
    \midrule
    \multirow{4}{*}{SelvaBox}
      & JS Distance        & 0.6248 & 0.6476 & 0.2789 & 0.2622 & 0.2152 \\
      & KL Divergence      & 2.4293 & 2.1779 & 0.3826 & 0.7028 & 0.1687 \\
      & KS Test            & 0.6231 & 0.6338 & 0.3257 & 0.2270 & 0.2117 \\
      & KS Test $p$-value  & $<1 \cdot 10^{-35}$ & $<1 \cdot 10^{-35}$ & $<1 \cdot 10^{-35}$ & $<1 \cdot 10^{-35}$ & $<1 \cdot 10^{-35}$ \\
    \bottomrule
  \end{tabular}}
\end{table}

\clearpage
\subsection{External methods evaluation}
We keep the default Detectree2 inference parameters provided in their python library. For DeepForest, we use their python library directly to benchmark their method but limit input size to $1000 \times 1000$ pixels maximum following their documentation guidelines and examples.

\subsection{ReforesTree dataset qualitative results.\label{sec:appendix_reforestree_qualitative}}
\label{sec:appendix_reforestree_quali}
\begin{figure}[ht]
    \centering
    \includegraphics[width=1.0\linewidth]{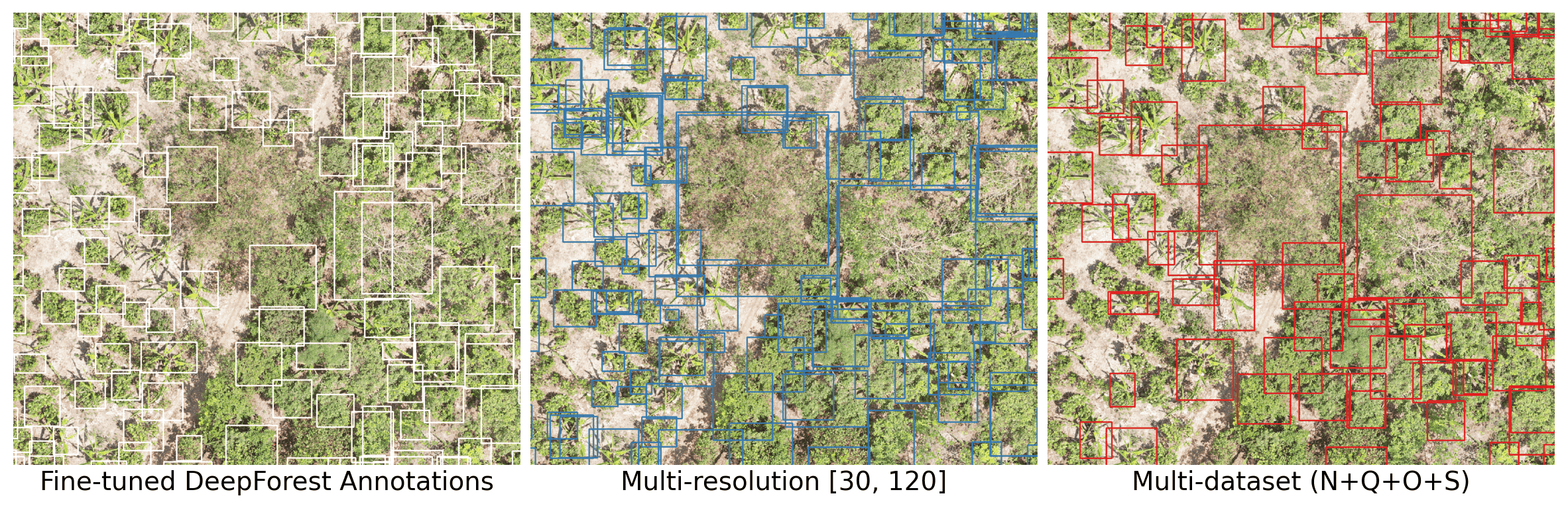}
    \vspace{-0.5cm}
    \caption{\textbf{Qualitative results on ReforesTree.} In white the ReforesTree annotations generated from an in-distribution and fine-tuned DeepForest model, in blue our best multi-resolution [30, 120] model and in red our best model trained on multi-dataset + \textsc{SelvaBox} (both our methods are OOD). Results are shown post-NMS, using the optimal NMS IoU ($\tau_{\mathrm{nms}}$) and score ($s_{\min}$) thresholds for RF1$_{75}$ from Algorithm~\ref{alg:dataset_eval} (see Section~\ref{sec:appendix_nms_hyperparameters} for exact values). These examples illustrate the superior detection performance of our DINO-Swin models compared to ReforesTree annotations, especially for larger trees.
    }
    \label{fig:refores_tree_infer}
\end{figure}

\clearpage
\subsection{Tropical datasets qualitative results.\label{sec:appendix_detectree2_qualitative}}

\begin{figure}[ht]
    \centering
    \includegraphics[width=1.0\linewidth]{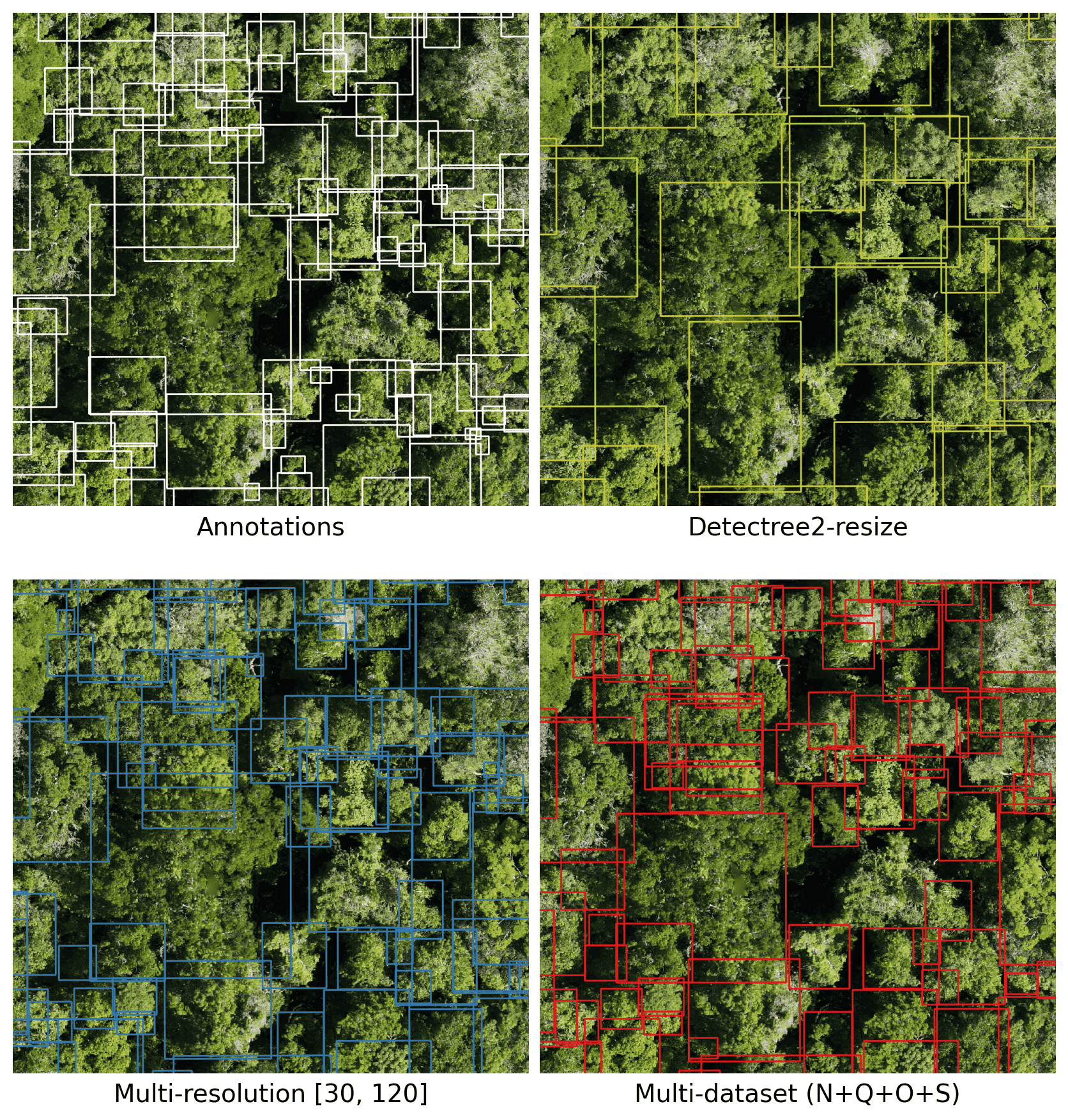}
    \vspace{-0.5cm}
    \caption{\textbf{Qualitative results on \textsc{SelvaBox} (Brazil)}. 
    We compare the annotations in white, the best competing method Detectree2-resize (OOD) in yellow, our best multi-resolution [30, 120] model (ID) in blue and our best model trained on multi-dataset + \textsc{SelvaBox} (ID) in red. Results are shown post-NMS, using the optimal NMS IoU ($\tau_{\mathrm{nms}}$) and score ($s_{\min}$) thresholds for RF1$_{75}$ from Algorithm~\ref{alg:dataset_eval} (see Section~\ref{sec:appendix_nms_hyperparameters} for exact values).}
    \label{fig:selvabox_tree_infer_brazil}
\end{figure}

\begin{figure}[ht]
    \centering
    \includegraphics[width=1.0\linewidth]{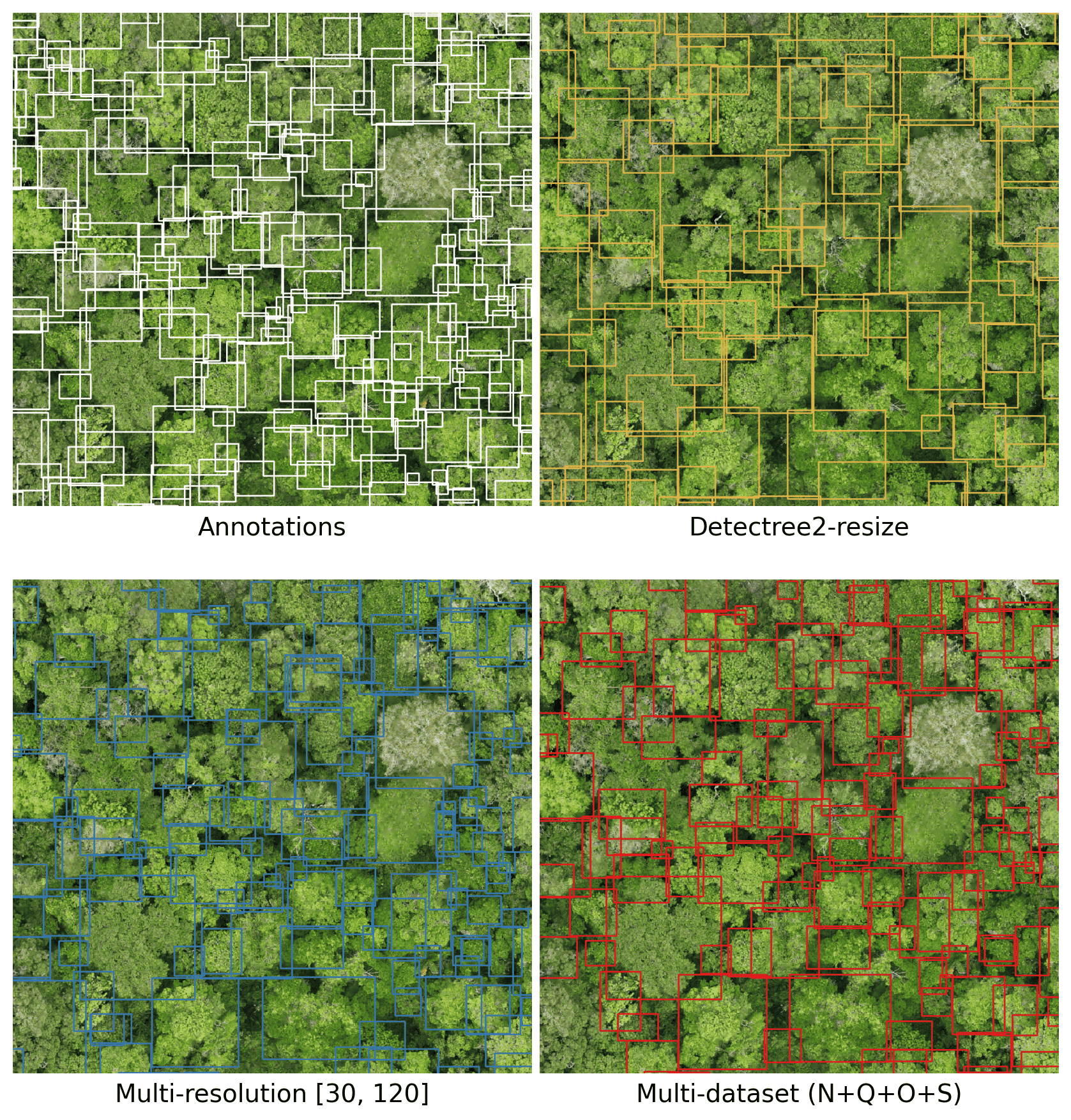}
    \vspace{-0.5cm}
    \caption{\textbf{Qualitative results on \textsc{SelvaBox} (Ecuador)}. 
    We compare the annotations in white, the best competing method Detectree2-resize (OOD) in yellow, our best multi-resolution [30, 120] model (ID) in blue and our best model trained on multi-dataset + \textsc{SelvaBox} (ID) in red. Results are shown post-NMS, using the optimal NMS IoU ($\tau_{\mathrm{nms}}$) and score ($s_{\min}$) thresholds for RF1$_{75}$ from Algorithm~\ref{alg:dataset_eval} (see Section~\ref{sec:appendix_nms_hyperparameters} for exact values).}
    \label{fig:selvabox_tree_infer_ecuador}
\end{figure}

\begin{figure}[ht]
    \centering
    \includegraphics[width=1.0\linewidth]{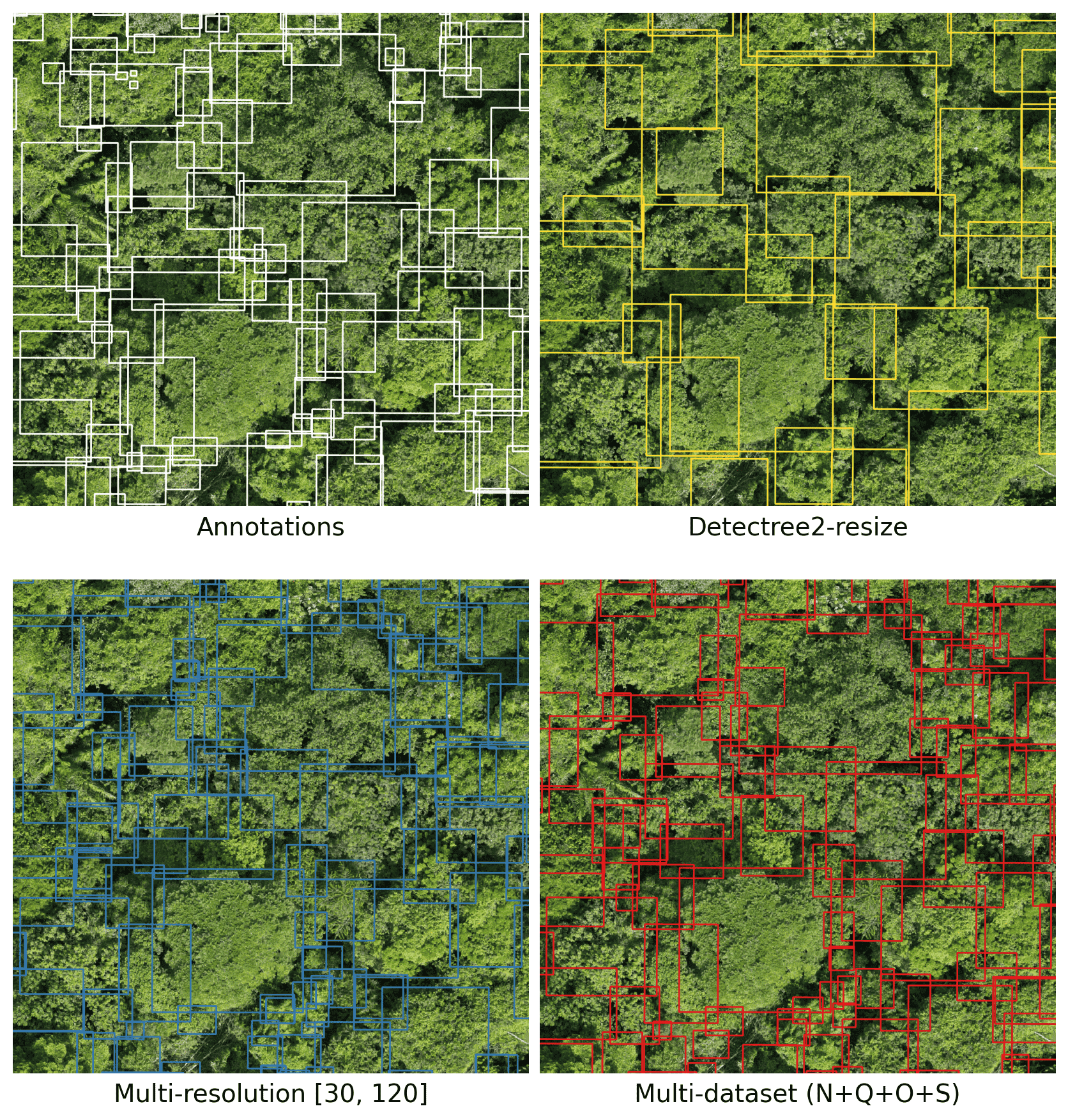}
    \vspace{-0.5cm}
    \caption{\textbf{Qualitative results on \textsc{SelvaBox} (Panama)}. 
    We compare the annotations in white, the best competing method Detectree2-resize (OOD) in yellow, our best multi-resolution [30, 120] model (ID) in blue and our best model trained on multi-dataset + \textsc{SelvaBox} (ID) in red. Results are shown post-NMS, using the optimal NMS IoU ($\tau_{\mathrm{nms}}$) and score ($s_{\min}$) thresholds for RF1$_{75}$ from Algorithm~\ref{alg:dataset_eval} (see Section~\ref{sec:appendix_nms_hyperparameters} for exact values).}
    \label{fig:selvabox_tree_infer_panama}
\end{figure}

\begin{figure}[ht]
    \centering
    \includegraphics[width=1.0\linewidth]{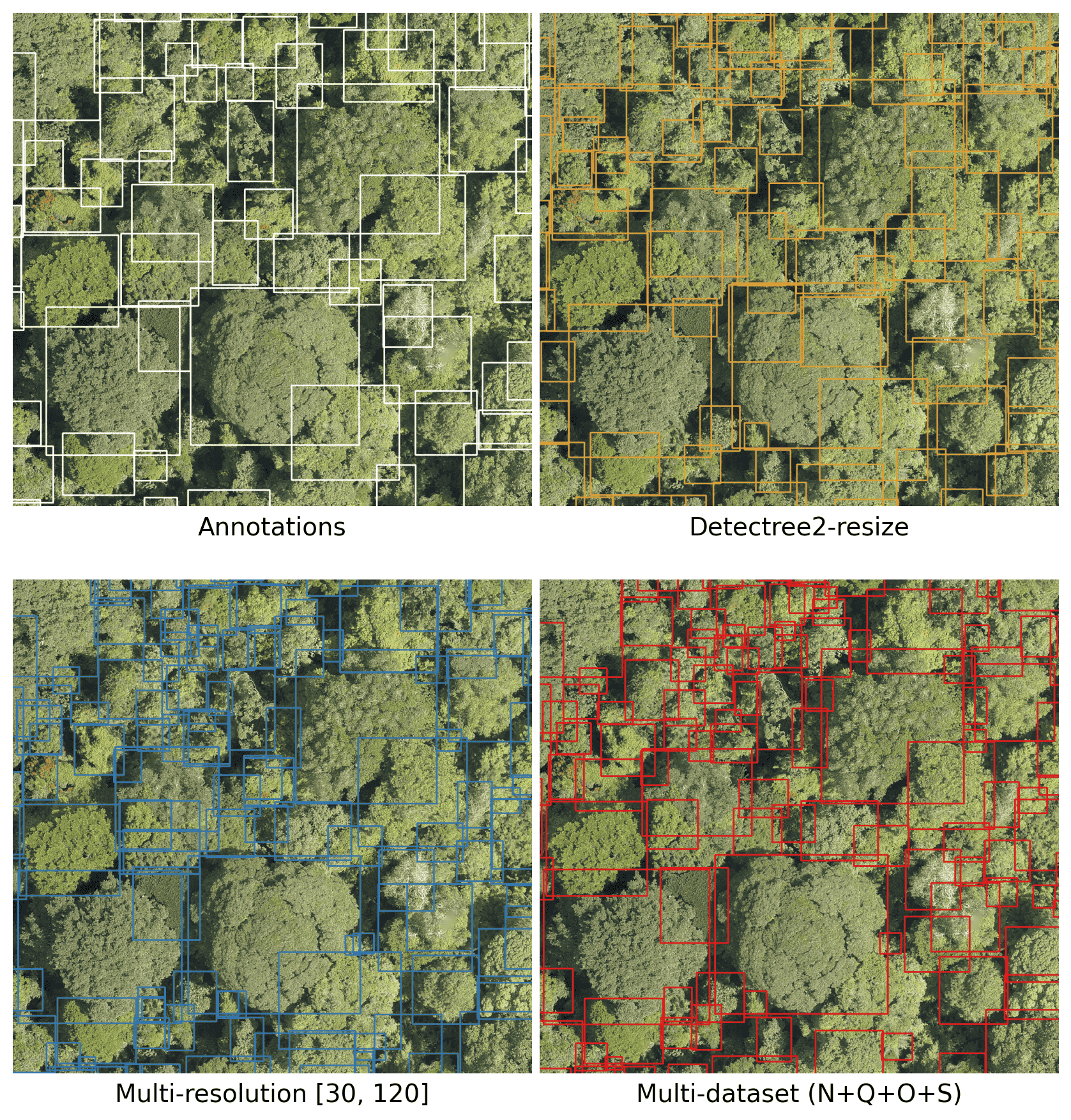}
    \vspace{-0.5cm}
    \caption{\textbf{Qualitative results on BCI50ha}. 
    We compare the annotations in white, the best competing method Detectree2-resize (OOD) in yellow, our best multi-resolution [30, 120] model (OOD) in blue and our best model trained on multi-dataset + \textsc{SelvaBox} (OOD) in red. Results are shown post-NMS, using the optimal NMS IoU ($\tau_{\mathrm{nms}}$) and score ($s_{\min}$) thresholds for RF1$_{75}$ from Algorithm~\ref{alg:dataset_eval} (see Section~\ref{sec:appendix_nms_hyperparameters} for exact values).}
    \label{fig:bci50ha_tree_infer}
\end{figure}

\begin{figure}[ht]
    \centering
    \includegraphics[width=1.0\linewidth]{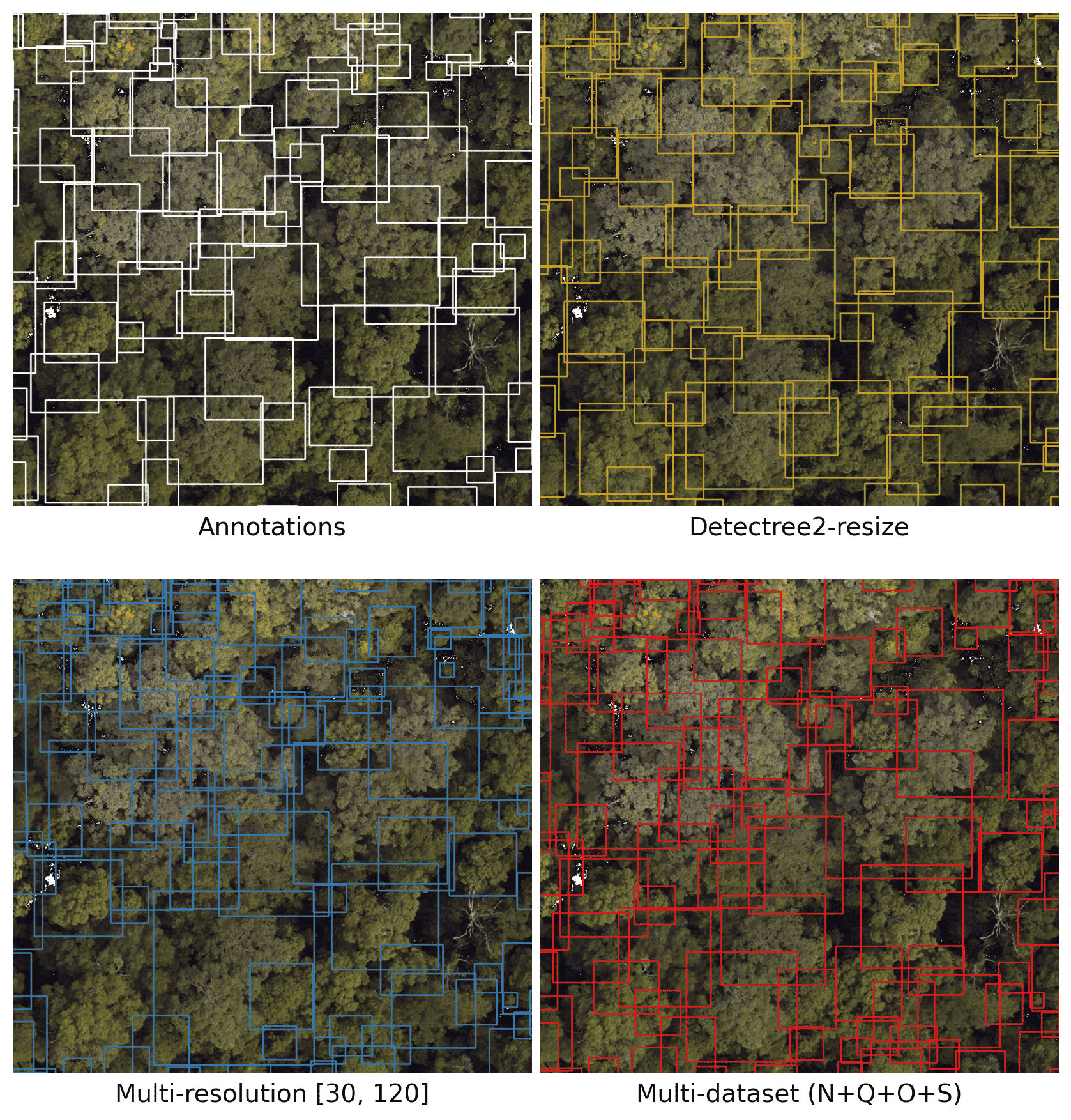}
    \vspace{-0.5cm}
    \caption{\textbf{Qualitative results on Detectree2 dataset}. 
    We compare the annotations in white, the best competing method Detectree2-resize (ID; possibly affected by train–test leakage, since we couldn’t recover their data splits) in yellow, our best multi-resolution [30, 120] model (OOD) in blue and our best model trained on multi-dataset + \textsc{SelvaBox} (OOD) in red. Results are shown post-NMS, using the optimal NMS IoU ($\tau_{\mathrm{nms}}$) and score ($s_{\min}$) thresholds for RF1$_{75}$ from Algorithm~\ref{alg:dataset_eval} (see Section~\ref{sec:appendix_nms_hyperparameters} for exact values).}
    \label{fig:detectree2_tree_infer}
\end{figure}

\clearpage
\subsection{Non-tropical datasets qualitative results.\label{sec:appendix_other_datasets_qualitative}}

\begin{figure}[ht]
    \centering
    \includegraphics[width=1.0\linewidth]{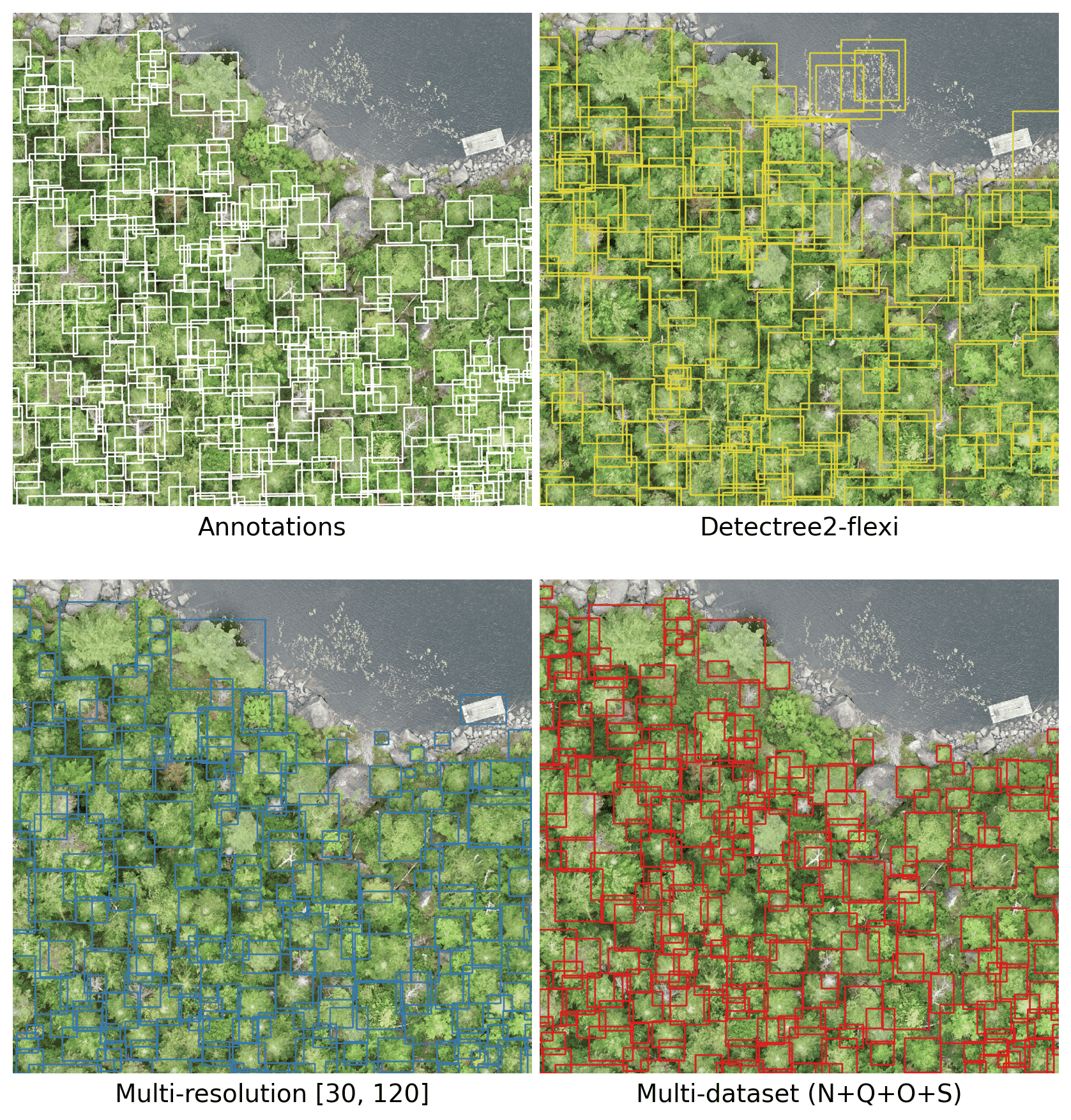}
    \vspace{-0.5cm}
    \caption{\textbf{Qualitative results on QuebecTrees}. 
    We compare the annotations in white, the best competing method Detectree2-flexi (OOD) in yellow, our best multi-resolution [30, 120] model (OOD) in blue and our best model trained on multi-dataset + \textsc{SelvaBox} (ID) in red. Results are shown post-NMS, using the optimal NMS IoU ($\tau_{\mathrm{nms}}$) and score ($s_{\min}$) thresholds for RF1$_{75}$ from Algorithm~\ref{alg:dataset_eval} (see Section~\ref{sec:appendix_nms_hyperparameters} for exact values).}
    \label{fig:quebectrees_tree_infer}
\end{figure}




\clearpage

\section{Python Libraries\label{sec:python_libraries}} 

\subsection{geodataset}

We’ve released our pip-installable Python library \textit{geodataset} on GitHub under the permissive Apache~2.0 license. The library serves four main purposes: \circledgray{1} Tilerizers for cutting rasters into tiles—with resampling, AOI, and pixel-masking support—for training/evaluation (as COCO-style JSON) or inference; \circledgray{2} an Aggregator tool that converts predicted object coordinates back into the original CRS and efficiently performs NMS on large sets of detections (at the raster-level); \circledgray{3} base dataset classes for training and inference that integrate easily with PyTorch’s DataLoader; and \circledgray{4} standardized conventions for naming tiles and COCO JSON files. See the repository documentation for more details.

\subsection{CanopyRS}

We’ve released a Python GitHub repository called \textit{CanopyRS} to replicate our results, benchmark models, and infer on new forest imagery. It’s distributed under the permissive Apache~2.0 license and leverages \textit{geodataset} for pre- and post-processing, with Detectron2 and Detrex handling model training. Its modular design makes it easy to extend in future work—for example, supporting instance segmentation, clustering, or classification of individual trees. See the repository documentation for more details.

\section{Use of Large Language Models (LLMs)}

We used LLMs as general-purpose assistive tools to improve the clarity and conciseness of the text, as well as for occasional coding assistance. These tools were not used for research ideation or to generate novel scientific content. All conceptual and experimental contributions were made by the authors.

\end{document}